\documentclass[10pt,twocolumn,letterpaper]{article}

\usepackage[pagenumbers]{cvpr}  

\usepackage{cuted}      
\usepackage{caption}    
\usepackage{tikz}       

\usepackage{pifont,multirow,longtable,array,placeins}
\usepackage{pgfplots}\pgfplotsset{compat=1.17}
\usepackage{algorithm}\usepackage{algpseudocode}
\usetikzlibrary{calc}
\definecolor{ourteal}{RGB}{14,138,140}\definecolor{altlav}{RGB}{155,134,198}\definecolor{inkgray}{RGB}{107,113,120}

\definecolor{cvprblue}{rgb}{0.21,0.49,0.74}
\usepackage[breaklinks,colorlinks,allcolors=cvprblue]{hyperref}

\def\confName{CVPR}
\def\confYear{2027}
\usepackage{xcolor}

\newcommand{\ours}{OPIE}

\newcommand{\pairacc}{PairAcc}

\newcommand{\ci}[2]{\,{\scriptsize[#1,\,#2]}}

\newcommand{\br}[1]{\texttt{#1}}

\newcommand{\devPairAccGain}{+0.106}
\newcommand{\devPairAccGainCI}{\ci{+0.029}{+0.183}}
\newcommand{\devMarginGain}{+0.083}
\newcommand{\devMarginGainCI}{\ci{+0.055}{+0.113}}

\newcommand{\devGateR}{7/7}

\newcommand{\devConsThreePairAccGain}{+0.072}
\newcommand{\devConsThreePairAccGainCI}{\ci{-0.011}{+0.147}}
\newcommand{\devConsThreeMarginGain}{+0.073}
\newcommand{\devConsThreeMarginGainCI}{\ci{+0.046}{+0.100}}

\newcommand{\devConsThreeGateR}{6/7}

\newcommand{\devBGateR}{7/7}

\newcommand{\devBConsThreeGateR}{6/7}

\newcommand{\freshMarginNative}{0.012}

\newcommand{\sealedMarginNative}{0.020}

\newcommand{\freshNativeMaxDiffMargin}{0.00612}

\newcommand{\sealedNativeMaxDiffMargin}{0.00734}

\newcommand{\dgASnr}{1.85}

\newcommand{\dgACos}{0.59}

\newcommand{\dgAAmp}{1.23}

\newcommand{\dgAPairAcc}{0.862}

\newcommand{\dgAMargin}{0.234}

\newcommand{\dgAZero}{1.06}

\newcommand{\dgPSnr}{1.35}

\newcommand{\dgPCos}{0.17}

\newcommand{\dgPAmp}{0.95}

\newcommand{\dgPPairAcc}{0.607}

\newcommand{\dgPMargin}{0.034}

\newcommand{\dgPZero}{1.38}

\newcommand{\dgUSnr}{0.24}

\newcommand{\dgUCos}{0.16}

\newcommand{\dgUAmp}{0.25}

\newcommand{\dgUPairAcc}{0.739}

\newcommand{\dgUMargin}{0.054}

\newcommand{\dgUZero}{1.03}

\newcommand{\abGtNPairAcc}{+0.021}
\newcommand{\abGtNPairAccCI}{\ci{\ensuremath{-}0.054}{+0.097}}
\newcommand{\abGtNMargin}{+0.036}
\newcommand{\abGtNMarginCI}{\ci{+0.009}{+0.064}}

\newcommand{\abGtNDPairAcc}{\ensuremath{-}0.092}
\newcommand{\abGtNDPairAccCI}{\ci{\ensuremath{-}0.138}{\ensuremath{-}0.049}}
\newcommand{\abGtNDMargin}{\ensuremath{-}0.050}
\newcommand{\abGtNDMarginCI}{\ci{\ensuremath{-}0.064}{\ensuremath{-}0.037}}

\newcommand{\abNoDirPairAcc}{\ensuremath{-}0.004}
\newcommand{\abNoDirPairAccCI}{\ci{\ensuremath{-}0.078}{+0.069}}
\newcommand{\abNoDirMargin}{+0.033}
\newcommand{\abNoDirMarginCI}{\ci{+0.007}{+0.059}}

\newcommand{\abNoDirDPairAcc}{\ensuremath{-}0.117}
\newcommand{\abNoDirDPairAccCI}{\ci{\ensuremath{-}0.151}{\ensuremath{-}0.080}}
\newcommand{\abNoDirDMargin}{\ensuremath{-}0.053}
\newcommand{\abNoDirDMarginCI}{\ci{\ensuremath{-}0.067}{\ensuremath{-}0.038}}
\newcommand{\abNoDirDObj}{0.0000}

\newcommand{\abNoDirDCont}{0.0000}

\newcommand{\abRankEightPairAcc}{+0.115}
\newcommand{\abRankEightPairAccCI}{\ci{+0.041}{+0.195}}
\newcommand{\abRankEightMargin}{+0.078}
\newcommand{\abRankEightMarginCI}{\ci{+0.053}{+0.106}}

\newcommand{\abRankEightDPairAcc}{+0.002}
\newcommand{\abRankEightDPairAccCI}{\ci{\ensuremath{-}0.028}{+0.034}}
\newcommand{\abRankEightDMargin}{\ensuremath{-}0.008}
\newcommand{\abRankEightDMarginCI}{\ci{\ensuremath{-}0.015}{0.000}}

\newcommand{\abRankFourPairAcc}{+0.075}
\newcommand{\abRankFourPairAccCI}{\ci{+0.003}{+0.147}}
\newcommand{\abRankFourMargin}{+0.065}
\newcommand{\abRankFourMarginCI}{\ci{+0.041}{+0.089}}

\newcommand{\abRankFourDPairAcc}{\ensuremath{-}0.038}
\newcommand{\abRankFourDPairAccCI}{\ci{\ensuremath{-}0.078}{+0.002}}
\newcommand{\abRankFourDMargin}{\ensuremath{-}0.022}
\newcommand{\abRankFourDMarginCI}{\ci{\ensuremath{-}0.032}{\ensuremath{-}0.012}}

\newcommand{\abDataSmallPairAcc}{+0.104}
\newcommand{\abDataSmallPairAccCI}{\ci{+0.021}{+0.187}}
\newcommand{\abDataSmallMargin}{+0.085}
\newcommand{\abDataSmallMarginCI}{\ci{+0.057}{+0.114}}

\newcommand{\abDataSmallDPairAcc}{\ensuremath{-}0.009}
\newcommand{\abDataSmallDPairAccCI}{\ci{\ensuremath{-}0.038}{+0.022}}
\newcommand{\abDataSmallDMargin}{\ensuremath{-}0.002}
\newcommand{\abDataSmallDMarginCI}{\ci{\ensuremath{-}0.009}{+0.005}}

\newcommand{\abDataMidPairAcc}{+0.116}
\newcommand{\abDataMidPairAccCI}{\ci{+0.039}{+0.192}}
\newcommand{\abDataMidMargin}{+0.086}
\newcommand{\abDataMidMarginCI}{\ci{+0.061}{+0.113}}

\newcommand{\abDataMidDPairAcc}{+0.003}
\newcommand{\abDataMidDPairAccCI}{\ci{\ensuremath{-}0.024}{+0.029}}
\newcommand{\abDataMidDMargin}{0.000}
\newcommand{\abDataMidDMarginCI}{\ci{\ensuremath{-}0.009}{+0.007}}

\newcommand{\ftOursLr}{$10^{-6}$}
\newcommand{\wsPairAcc}{\ensuremath{-}0.024}
\newcommand{\wsPairAccCI}{\ci{\ensuremath{-}0.056}{+0.010}}
\newcommand{\wsMargin}{\ensuremath{-}0.007}
\newcommand{\wsMarginCI}{\ci{\ensuremath{-}0.015}{0.000}}

\newcommand{\wsxScoreMax}{\ensuremath{-}0.0003}

\newcommand{\cpFreshN}{117}
\newcommand{\cpFreshNat}{+0.0407}
\newcommand{\cpFreshNatCI}{\ci{+0.0351}{+0.0468}}
\newcommand{\cpFreshCtrl}{+0.0339}
\newcommand{\cpFreshCtrlCI}{\ci{+0.0294}{+0.0387}}
\newcommand{\cpFreshPairAcc}{+0.106}
\newcommand{\cpFreshPairAccCI}{\ci{+0.079}{+0.133}}

\newcommand{\cpSealedN}{118}
\newcommand{\cpSealedNat}{+0.0386}
\newcommand{\cpSealedNatCI}{\ci{+0.0331}{+0.0443}}
\newcommand{\cpSealedCtrl}{+0.0300}
\newcommand{\cpSealedCtrlCI}{\ci{+0.0256}{+0.0345}}
\newcommand{\cpSealedPairAcc}{+0.087}
\newcommand{\cpSealedPairAccCI}{\ci{+0.058}{+0.115}}

\newcommand{\awFreshPairAcc}{\ensuremath{-}0.004}
\newcommand{\awFreshPairAccCI}{\ci{\ensuremath{-}0.025}{+0.017}}

\newcommand{\awFreshMargin}{+0.0071}
\newcommand{\awFreshMarginCI}{\ci{+0.0003}{+0.0137}}
\newcommand{\awFreshMarginCtrl}{+0.0044}
\newcommand{\awFreshMarginCtrlCI}{\ci{\ensuremath{-}0.0013}{+0.0101}}

\newcommand{\awFreshDecision}{CONFIRMED NONINFERIOR}
\newcommand{\awFreshCtrlDecision}{CONFIRMED NONINFERIOR}
\newcommand{\awSealedPairAcc}{+0.007}
\newcommand{\awSealedPairAccCI}{\ci{\ensuremath{-}0.009}{+0.023}}

\newcommand{\awSealedMargin}{+0.0035}
\newcommand{\awSealedMarginCI}{\ci{\ensuremath{-}0.0033}{+0.0100}}
\newcommand{\awSealedMarginCtrl}{+0.0037}
\newcommand{\awSealedMarginCtrlCI}{\ci{\ensuremath{-}0.0019}{+0.0097}}

\newcommand{\awSealedDecision}{CONFIRMED NONINFERIOR}
\newcommand{\awSealedCtrlDecision}{CONFIRMED NONINFERIOR}

\newcommand{\tsRateNative}{0.583}
\newcommand{\tsRateRepaired}{0.657}

\newcommand{\tsRepNat}{+0.073}
\newcommand{\tsRepNatCI}{\ci{+0.007}{+0.133}}
\newcommand{\tsRepNatP}{0.026}

\newcommand{\apAurocReal}{0.975}

\newcommand{\tsXNativeRate}{0.880}
\newcommand{\tsXLoraRate}{0.930}
\newcommand{\tsXLoraNativeDiff}{+0.050}
\newcommand{\tsXLoraNativeCI}{\ci{+0.020}{+0.080}}
\newcommand{\tsXLoraNativeP}{0.0012}

\newcommand{\scXMargin}{+0.096}
\newcommand{\scXMarginLo}{+0.082}
\newcommand{\scXMarginHi}{+0.109}
\newcommand{\scXMarginCI}{[\scXMarginLo{},\scXMarginHi{}]}
\newcommand{\scXFullMargin}{+0.096}
\newcommand{\scXFullMarginLo}{+0.083}
\newcommand{\scXFullMarginHi}{+0.110}
\newcommand{\scXFullMarginCI}{[\scXFullMarginLo{},\scXFullMarginHi{}]}

\newcommand{\scXPair}{+0.163}

\newcommand{\scXFVD}{1.02}

\newcommand{\scXFullFVD}{1.62}

\newcommand{\awXMargin}{+0.0050}

\newcommand{\awXPair}{+0.0045}

\newcommand{\awXFVD}{1.06}

\newcommand{\htXGain}{+0.107}

\newcommand{\bbXBTwoFVD}{1.69}

\newcommand{\scXAuditIValue}{1.4825}
\newcommand{\scXAuditIValueLo}{1.2680}
\newcommand{\scXAuditIValueHi}{1.5572}
\newcommand{\scXAuditIValueCI}{[\scXAuditIValueLo{},\scXAuditIValueHi{}]}
\newcommand{\scXAuditJValue}{1.0003}
\newcommand{\scXAuditJValueLo}{0.9260}
\newcommand{\scXAuditJValueHi}{1.1026}
\newcommand{\scXAuditJValueCI}{[\scXAuditJValueLo{},\scXAuditJValueHi{}]}

\newcommand{\scXAuditLValue}{0.9599}
\newcommand{\scXAuditLValueLo}{0.8439}
\newcommand{\scXAuditLValueHi}{1.0368}
\newcommand{\scXAuditLValueCI}{[\scXAuditLValueLo{},\scXAuditLValueHi{}]}

\newcommand{\scXAuditOValue}{-0.0002}
\newcommand{\scXAuditOValueLo}{-0.0009}
\newcommand{\scXAuditOValueHi}{0.0007}
\newcommand{\scXAuditOValueCI}{[\scXAuditOValueLo{},\scXAuditOValueHi{}]}

\newcommand{\scXAuditRValue}{0.0085}
\newcommand{\scXAuditRValueLo}{0.0050}
\newcommand{\scXAuditRValueHi}{0.0130}
\newcommand{\scXAuditRValueCI}{[\scXAuditRValueLo{},\scXAuditRValueHi{}]}

\newcommand{\scXAuditTValue}{0.1074}
\newcommand{\scXAuditTValueLo}{0.0894}
\newcommand{\scXAuditTValueHi}{0.1268}
\newcommand{\scXAuditTValueCI}{[\scXAuditTValueLo{},\scXAuditTValueHi{}]}

\newcommand{\scXAuditVValue}{0.0076}
\newcommand{\scXAuditVValueLo}{0.0051}
\newcommand{\scXAuditVValueHi}{0.0107}
\newcommand{\scXAuditVValueCI}{[\scXAuditVValueLo{},\scXAuditVValueHi{}]}
\newcommand{\scXAuditWValue}{0.1875}
\newcommand{\scXAuditWValueLo}{0.1413}
\newcommand{\scXAuditWValueHi}{0.2348}
\newcommand{\scXAuditWValueCI}{[\scXAuditWValueLo{},\scXAuditWValueHi{}]}

\newcommand{\scXAuditAHValue}{1.6219}
\newcommand{\scXAuditAHValueLo}{1.3846}
\newcommand{\scXAuditAHValueHi}{1.6674}
\newcommand{\scXAuditAHValueCI}{[\scXAuditAHValueLo{},\scXAuditAHValueHi{}]}
\newcommand{\scXAuditAIValue}{1.0192}
\newcommand{\scXAuditAIValueLo}{0.9383}
\newcommand{\scXAuditAIValueHi}{1.0852}
\newcommand{\scXAuditAIValueCI}{[\scXAuditAIValueLo{},\scXAuditAIValueHi{}]}

\newcommand{\scXAuditAKValue}{0.9567}
\newcommand{\scXAuditAKValueLo}{0.8734}
\newcommand{\scXAuditAKValueHi}{1.0299}
\newcommand{\scXAuditAKValueCI}{[\scXAuditAKValueLo{},\scXAuditAKValueHi{}]}

\newcommand{\scXAuditANValue}{-0.0004}
\newcommand{\scXAuditANValueLo}{-0.0009}
\newcommand{\scXAuditANValueHi}{0.0002}
\newcommand{\scXAuditANValueCI}{[\scXAuditANValueLo{},\scXAuditANValueHi{}]}

\newcommand{\scXAuditAQValue}{0.0074}
\newcommand{\scXAuditAQValueLo}{0.0049}
\newcommand{\scXAuditAQValueHi}{0.0105}
\newcommand{\scXAuditAQValueCI}{[\scXAuditAQValueLo{},\scXAuditAQValueHi{}]}

\newcommand{\scXAuditASValue}{0.0955}
\newcommand{\scXAuditASValueLo}{0.0824}
\newcommand{\scXAuditASValueHi}{0.1091}
\newcommand{\scXAuditASValueCI}{[\scXAuditASValueLo{},\scXAuditASValueHi{}]}

\newcommand{\scXAuditAUValue}{0.0066}
\newcommand{\scXAuditAUValueLo}{0.0049}
\newcommand{\scXAuditAUValueHi}{0.0086}
\newcommand{\scXAuditAUValueCI}{[\scXAuditAUValueLo{},\scXAuditAUValueHi{}]}
\newcommand{\scXAuditAVValue}{0.1631}
\newcommand{\scXAuditAVValueLo}{0.1294}
\newcommand{\scXAuditAVValueHi}{0.1958}
\newcommand{\scXAuditAVValueCI}{[\scXAuditAVValueLo{},\scXAuditAVValueHi{}]}

\newcommand{\scXAuditBGValue}{1.5291}
\newcommand{\scXAuditBGValueLo}{1.2671}
\newcommand{\scXAuditBGValueHi}{1.5808}
\newcommand{\scXAuditBGValueCI}{[\scXAuditBGValueLo{},\scXAuditBGValueHi{}]}
\newcommand{\scXAuditBHValue}{1.0186}
\newcommand{\scXAuditBHValueLo}{0.9249}
\newcommand{\scXAuditBHValueHi}{1.0774}
\newcommand{\scXAuditBHValueCI}{[\scXAuditBHValueLo{},\scXAuditBHValueHi{}]}

\newcommand{\scXAuditBJValue}{0.9646}
\newcommand{\scXAuditBJValueLo}{0.8582}
\newcommand{\scXAuditBJValueHi}{1.0500}
\newcommand{\scXAuditBJValueCI}{[\scXAuditBJValueLo{},\scXAuditBJValueHi{}]}

\newcommand{\scXAuditBMValue}{-0.0006}
\newcommand{\scXAuditBMValueLo}{-0.0014}
\newcommand{\scXAuditBMValueHi}{0.0002}
\newcommand{\scXAuditBMValueCI}{[\scXAuditBMValueLo{},\scXAuditBMValueHi{}]}

\newcommand{\scXAuditBPValue}{0.0063}
\newcommand{\scXAuditBPValueLo}{0.0032}
\newcommand{\scXAuditBPValueHi}{0.0102}
\newcommand{\scXAuditBPValueCI}{[\scXAuditBPValueLo{},\scXAuditBPValueHi{}]}

\newcommand{\scXAuditBRValue}{0.0838}
\newcommand{\scXAuditBRValueLo}{0.0659}
\newcommand{\scXAuditBRValueHi}{0.1029}
\newcommand{\scXAuditBRValueCI}{[\scXAuditBRValueLo{},\scXAuditBRValueHi{}]}

\newcommand{\scXAuditBTValue}{0.0056}
\newcommand{\scXAuditBTValueLo}{0.0037}
\newcommand{\scXAuditBTValueHi}{0.0082}
\newcommand{\scXAuditBTValueCI}{[\scXAuditBTValueLo{},\scXAuditBTValueHi{}]}
\newcommand{\scXAuditBUValue}{0.1392}
\newcommand{\scXAuditBUValueLo}{0.0943}
\newcommand{\scXAuditBUValueHi}{0.1825}
\newcommand{\scXAuditBUValueCI}{[\scXAuditBUValueLo{},\scXAuditBUValueHi{}]}

\newcommand{\clXAuditAWValue}{0.0900}
\newcommand{\clXAuditAWValueLo}{0.0317}
\newcommand{\clXAuditAWValueHi}{0.1492}
\newcommand{\clXAuditAWValueCI}{[\clXAuditAWValueLo{},\clXAuditAWValueHi{}]}

\newcommand{\awXAuditAQValue}{-0.0009}
\newcommand{\awXAuditAQValueLo}{-0.0014}
\newcommand{\awXAuditAQValueHi}{-0.0004}
\newcommand{\awXAuditAQValueCI}{[\awXAuditAQValueLo{},\awXAuditAQValueHi{}]}

\newcommand{\awXAuditASValue}{0.0050}
\newcommand{\awXAuditASValueLo}{0.0014}
\newcommand{\awXAuditASValueHi}{0.0089}
\newcommand{\awXAuditASValueCI}{[\awXAuditASValueLo{},\awXAuditASValueHi{}]}

\newcommand{\awXAuditAUValue}{-0.0004}
\newcommand{\awXAuditAUValueLo}{-0.0006}
\newcommand{\awXAuditAUValueHi}{-0.0001}
\newcommand{\awXAuditAUValueCI}{[\awXAuditAUValueLo{},\awXAuditAUValueHi{}]}
\newcommand{\awXAuditAVValue}{0.0045}
\newcommand{\awXAuditAVValueLo}{-0.0074}
\newcommand{\awXAuditAVValueHi}{0.0156}
\newcommand{\awXAuditAVValueCI}{[\awXAuditAVValueLo{},\awXAuditAVValueHi{}]}

\newcommand{\bbXAuditEValue}{0.0542}
\newcommand{\bbXAuditEValueLo}{0.0457}
\newcommand{\bbXAuditEValueHi}{0.0626}
\newcommand{\bbXAuditEValueCI}{[\bbXAuditEValueLo{},\bbXAuditEValueHi{}]}

\newcommand{\bbXAuditMValue}{0.0610}
\newcommand{\bbXAuditMValueLo}{0.0524}
\newcommand{\bbXAuditMValueHi}{0.0693}
\newcommand{\bbXAuditMValueCI}{[\bbXAuditMValueLo{},\bbXAuditMValueHi{}]}

\newcommand{\bbXFAuditOValueLo}{1.4061}
\newcommand{\bbXFAuditOValueHi}{1.7128}
\newcommand{\bbXFAuditOValueCI}{[\bbXFAuditOValueLo{},\bbXFAuditOValueHi{}]}

\newcommand{\bbXFAuditRValue}{1.0129}

\newcommand{\gSevNative}{0.860}
\newcommand{\gSevSched}{0.916}
\newcommand{\gSevGain}{+0.056}
\newcommand{\gSevGainCI}{[+0.027,+0.084]}
\newcommand{\gSevP}{0.00035}

\newcommand{\eFourNEp}{300}
\newcommand{\eFourNative}{0.583}
\newcommand{\eFourBase}{0.583}
\newcommand{\eFourOracle}{0.687}

\newcommand{\eFourBOneRate}{0.560}
\newcommand{\eFourBTwoRate}{0.584}
\newcommand{\eFourRecipeRate}{0.673}
\newcommand{\eFourBOneRec}{-0.113}
\newcommand{\eFourBOneRecCI}{[-0.154,-0.073]}
\newcommand{\eFourBOneRecP}{<0.0001}
\newcommand{\eFourBTwoRec}{-0.089}
\newcommand{\eFourBTwoRecCI}{[-0.131,-0.047]}
\newcommand{\eFourBTwoRecP}{0.0001}
\newcommand{\eFourBOneNat}{-0.023}
\newcommand{\eFourBOneNatCI}{[-0.084,+0.033]}

\newcommand{\eFourBTwoNat}{+0.001}
\newcommand{\eFourBTwoNatCI}{[-0.055,+0.056]}

\newcommand{\eFourRecNat}{+0.090}
\newcommand{\eFourRecNatCI}{[+0.032,+0.149]}
\newcommand{\eFourBOneBTwo}{-0.024}
\newcommand{\eFourBOneBTwoCI}{[-0.060,+0.010]}

\newcommand{\eFourHBOneRec}{-0.113}

\newcommand{\eFourHBTwoRec}{-0.090}

\newcommand{\eFvdNContexts}{236}
\newcommand{\eFvdNative}{92.6}
\newcommand{\eFvdFull}{150.2}
\newcommand{\eFvdSched}{94.4}
\newcommand{\eFvdFloor}{22.2}

\newcommand{\eFvdSchedNat}{+1.8}
\newcommand{\eFvdSchedNatCI}{[-7.3,+9.1]}
\newcommand{\eFvdFullNat}{+57.6}
\newcommand{\eFvdFullNatCI}{[+44.3,+73.4]}
\newcommand{\eFvdSchedOverFloor}{+0.08}
\newcommand{\eFvdFactNative}{181.6}
\newcommand{\eFvdFactFull}{278.5}
\newcommand{\eFvdFactSched}{173.8}

\newcommand{\eAbsNContexts}{236}
\newcommand{\eAbsObjNat}{0.0683}
\newcommand{\eAbsObjSched}{0.0616}
\newcommand{\eAbsObjFull}{0.0636}
\newcommand{\eAbsObjRelSched}{9.7}
\newcommand{\eAbsObjRelSchedCI}{[7.5,12.1]}
\newcommand{\eAbsObjRelFull}{6.8}
\newcommand{\eAbsObjRelFullCI}{[4.3,9.3]}
\newcommand{\eAbsContactNat}{0.0649}
\newcommand{\eAbsContactSched}{0.0575}
\newcommand{\eAbsContactFull}{0.0585}
\newcommand{\eAbsContactRelSched}{11.4}
\newcommand{\eAbsContactRelSchedCI}{[7.9,15.4]}
\newcommand{\eAbsContactRelFull}{9.9}
\newcommand{\eAbsContactRelFullCI}{[6.1,13.9]}
\newcommand{\eAbsGlobalNat}{0.0537}
\newcommand{\eAbsGlobalSched}{0.0414}
\newcommand{\eAbsGlobalFull}{0.0421}
\newcommand{\eAbsGlobalRelSched}{22.9}
\newcommand{\eAbsGlobalRelSchedCI}{[21.1,24.6]}
\newcommand{\eAbsGlobalRelFull}{21.6}
\newcommand{\eAbsGlobalRelFullCI}{[19.7,23.4]}

\newcommand{\eFlowNContexts}{236}
\newcommand{\eFlowDirNat}{0.106}
\newcommand{\eFlowDirSched}{0.382}
\newcommand{\eFlowDirFull}{0.378}
\newcommand{\eFlowDirSchedNat}{+0.276}
\newcommand{\eFlowDirSchedNatCI}{[+0.244,+0.308]}

\newcommand{\eFlowGtDir}{0.974}

\newcommand{\eCutNContexts}{236}

\newcommand{\eCutKThree}{+0.0766}

\newcommand{\eCutKFour}{+0.0955}

\newcommand{\eCutLate}{+0.0004}

\newcommand{\eCutPairLate}{-0.0062}

\newcommand{\eCutFracFour}{0.996}

\newcommand{\pNullNatPair}{0.551}

\newcommand{\pNullNatMarg}{0.016}

\newcommand{\foldGripGain}{+0.007}
\newcommand{\foldGripGainCI}{[-0.010,+0.024]}
\newcommand{\foldGripOwn}{+0.201}
\newcommand{\foldGripOwnCI}{[+0.178,+0.225]}
\newcommand{\foldGripRest}{-0.193}
\newcommand{\foldGripRestCI}{[-0.213,-0.174]}

\newcommand{\foldHoldGain}{+0.101}
\newcommand{\foldHoldGainCI}{[+0.087,+0.114]}
\newcommand{\foldHoldOwn}{+0.019}
\newcommand{\foldHoldOwnCI}{[+0.014,+0.025]}
\newcommand{\foldHoldRest}{+0.081}
\newcommand{\foldHoldRestCI}{[+0.069,+0.095]}
\newcommand{\foldHoldShare}{0.19}
\newcommand{\foldFlipGain}{+0.219}
\newcommand{\foldFlipGainCI}{[+0.182,+0.261]}
\newcommand{\foldFlipOwn}{+0.125}
\newcommand{\foldFlipOwnCI}{[+0.102,+0.149]}
\newcommand{\foldFlipRest}{+0.094}
\newcommand{\foldFlipRestCI}{[+0.076,+0.114]}
\newcommand{\foldFlipShare}{0.57}
\newcommand{\foldHalfGain}{+0.090}
\newcommand{\foldHalfGainCI}{[+0.075,+0.106]}
\newcommand{\foldHalfOwn}{+0.122}
\newcommand{\foldHalfOwnCI}{[+0.105,+0.141]}
\newcommand{\foldHalfRest}{-0.032}
\newcommand{\foldHalfRestCI}{[-0.045,-0.020]}
\newcommand{\foldHalfShare}{1.35}
\newcommand{\foldPoolGain}{+0.104}
\newcommand{\foldPoolGainCI}{[+0.092,+0.117]}
\newcommand{\foldPoolOwn}{+0.117}
\newcommand{\foldPoolOwnCI}{[+0.105,+0.129]}
\newcommand{\foldPoolRest}{-0.012}
\newcommand{\foldPoolRestCI}{[-0.015,-0.010]}
\newcommand{\foldPoolShare}{1.12}

\newcommand{\eCutSixFourMargin}{+0.0004}
\newcommand{\eCutSixFourMarginCI}{[-0.0002,+0.0010]}
\newcommand{\eCutSixFourPair}{-0.0064}
\newcommand{\eCutSixFourPairCI}{[-0.0097,-0.0032]}
\newcommand{\eCutFourThree}{+0.0190}
\newcommand{\eCutFourThreeCI}{[+0.0157,+0.0226]}

\newcommand{\aEGtfitMargRead}{lower than the recipe}

\newcommand{\aExGtnVsRecMarg}{-0.0432}
\newcommand{\aExGtnVsRecMargCI}{[-0.0491,-0.0373]}

\newcommand{\aExGtnVsRecMargTaskCI}{[-0.0511,-0.0358]}

\newcommand{\aExGtnVsRecPair}{-0.0654}
\newcommand{\aExGtnVsRecPairCI}{[-0.0826,-0.0493]}

\newcommand{\aExGtnVsRecPairTaskCI}{[-0.0849,-0.0462]}

\newcommand{\aExGtnVsNatMarg}{+0.0523}
\newcommand{\aExGtnVsNatMargCI}{[+0.0421,+0.0635]}

\newcommand{\aExGtnVsNatPair}{+0.0972}
\newcommand{\aExGtnVsNatPairCI}{[+0.0663,+0.1274]}

\newcommand{\aExNodirVsRecMarg}{-0.0429}
\newcommand{\aExNodirVsRecMargCI}{[-0.0503,-0.0362]}

\newcommand{\aExNodirVsRecMargTaskCI}{[-0.0502,-0.0360]}

\newcommand{\aExNodirVsRecPair}{-0.0840}
\newcommand{\aExNodirVsRecPairCI}{[-0.1010,-0.0671]}

\newcommand{\aExNodirVsRecPairTaskCI}{[-0.1023,-0.0658]}

\newcommand{\aExNodirVsNatMarg}{+0.0526}
\newcommand{\aExNodirVsNatMargCI}{[+0.0405,+0.0648]}

\newcommand{\aExNodirVsNatPair}{+0.0787}
\newcommand{\aExNodirVsNatPairCI}{[+0.0489,+0.1075]}

\newcommand{\aExGtfitVsRecMarg}{-0.0910}
\newcommand{\aExGtfitVsRecMargCI}{[-0.1038,-0.0779]}

\newcommand{\aExGtfitVsRecMargTaskCI}{[-0.1150,-0.0715]}

\newcommand{\aExGtfitVsRecPair}{-0.1578}
\newcommand{\aExGtfitVsRecPairCI}{[-0.1888,-0.1266]}

\newcommand{\aExGtfitVsRecPairTaskCI}{[-0.1973,-0.1247]}

\newcommand{\aExGtfitVsNatMarg}{+0.0045}
\newcommand{\aExGtfitVsNatMargCI}{[+0.0030,+0.0061]}

\newcommand{\aExGtfitVsNatPair}{+0.0048}
\newcommand{\aExGtfitVsNatPairCI}{[-0.0048,+0.0143]}

\newcommand{\aExRecVsNatMarg}{+0.0956}
\newcommand{\aExRecVsNatMargCI}{[+0.0834,+0.1087]}

\newcommand{\aExRecVsNatPair}{+0.1626}
\newcommand{\aExRecVsNatPairCI}{[+0.1307,+0.1940]}

\newcommand{\aExNFresh}{117}
\newcommand{\aExNSealed}{119}
\newcommand{\tObjFull}{-4.7}
\newcommand{\tObjSched}{-6.7}
\newcommand{\tContFull}{-6.4}
\newcommand{\tContSched}{-7.4}
\newcommand{\tObjNatAbs}{68.3}
\newcommand{\tContNatAbs}{64.9}
\newcommand{\tPairFull}{+0.156}

\newcommand{\tGtnMarg}{+0.052}
\newcommand{\tGtnPair}{+0.097}

\newcommand{\tNodirMarg}{+0.053}
\newcommand{\tNodirPair}{+0.079}

\newcommand{\tGtfitMarg}{+0.0045}
\newcommand{\tGtfitPair}{+0.005}
\newcommand{\tRecMarg}{+0.096}
\newcommand{\tRecPair}{+0.163}
\newcommand{\rpDiff}{+0.008}
\newcommand{\rpDiffCI}{[-0.048,+0.063]}
\newcommand{\rpP}{0.82}

\newcommand{\rpNatThis}{0.623}

\newcommand{\rpSchedThis}{0.632}

\newcommand{\pbDirSched}{+0.290}
\newcommand{\pbDirSchedCI}{[+0.221,+0.362]}

\newcommand{\pbDirFull}{+0.295}
\newcommand{\pbDirFullCI}{[+0.223,+0.367]}

\newcommand{\pbDirGateKeep}{-0.004}
\newcommand{\pbDirGateKeepCI}{[-0.014,+0.005]}

\newcommand{\pbRetSched}{+0.072}
\newcommand{\pbRetSchedCI}{[+0.027,+0.116]}

\newcommand{\pbRetFull}{+0.063}
\newcommand{\pbRetFullCI}{[+0.021,+0.106]}

\newcommand{\pbRetGateKeep}{+0.008}
\newcommand{\pbRetGateKeepCI}{[-0.005,+0.022]}

\newcommand{\pbRsq}{0.92}
\newcommand{\pbMedErr}{1.6}
\newcommand{\pbDispCos}{0.91}
\newcommand{\pbRetGT}{0.88}
\newcommand{\pbNEligDir}{183}
\newcommand{\pbNEligRet}{157}
\newcommand{\gsScAThreeQ}{+0.069}
\newcommand{\gsScAThreeQCI}{[+0.059,+0.079]}
\newcommand{\gsScAHalf}{+0.042}
\newcommand{\gsScAHalfCI}{[+0.035,+0.049]}
\newcommand{\gsScAQuarter}{+0.017}
\newcommand{\gsScAQuarterCI}{[+0.014,+0.020]}
\newcommand{\gsScGateKeep}{-0.0004}
\newcommand{\gsScGateKeepCI}{[-0.0010,+0.0002]}

\newcommand{\pcN}{117}
\newcommand{\pcMarg}{+0.0272}
\newcommand{\pcMargCI}{[+0.0177,+0.0366]}
\newcommand{\pcPair}{+0.171}
\newcommand{\pcPairCI}{[+0.128,+0.213]}
\newcommand{\pcLpips}{-0.035}
\newcommand{\pcFvd}{0.914}
\newcommand{\pcFvdCI}{[0.81,0.97]}
\newcommand{\pcFloor}{0.0004}
\newcommand{\ccEpisodes}{1500}
\newcommand{\ccNI}{0.03}
\newcommand{\ccNat}{+0.0123}
\newcommand{\ccNatCI}{[-0.0067,+0.0333]}
\newcommand{\ccBase}{+0.0277}
\newcommand{\ccBaseCI}{[+0.0067,+0.0487]}
\newcommand{\ccOracle}{+0.066}
\newcommand{\ceEpisodes}{2000}
\newcommand{\ceRateBase}{0.812}
\newcommand{\ceRateNat}{0.813}
\newcommand{\ceRateSched}{0.832}
\newcommand{\ceRateOracle}{0.8845}
\newcommand{\ceNat}{+0.0193}
\newcommand{\ceNatCI}{[+0.0013,+0.0383]}
\newcommand{\ceBase}{+0.0203}
\newcommand{\ceBaseCI}{[+0.0027,+0.0378]}

\newcommand{\hbN}{117}
\newcommand{\hbOpieGain}{+0.107}
\newcommand{\hbOpieFvd}{0.99}

\newcommand{\hbWsGain}{-0.004}
\newcommand{\hbWsDiff}{+0.111}
\newcommand{\hbWsCI}{[+0.093,+0.131]}
\newcommand{\hbWsFvd}{1.15}
\newcommand{\hbFrgGain}{-0.001}
\newcommand{\hbFrgDiff}{+0.108}
\newcommand{\hbFrgCI}{[+0.090,+0.128]}
\newcommand{\hbFrgFvd}{1.00}
\newcommand{\hbFroGain}{+0.002}
\newcommand{\hbFroDiff}{+0.105}
\newcommand{\hbFroCI}{[+0.086,+0.124]}
\newcommand{\hbFroFvd}{0.96}
\newcommand{\hbBOneGain}{+0.071}
\newcommand{\hbBOneDiff}{+0.036}
\newcommand{\hbBOneCI}{[+0.024,+0.049]}
\newcommand{\hbBOneFvd}{0.71}
\newcommand{\hbBTwoGain}{+0.154}
\newcommand{\hbBTwoDiff}{-0.047}
\newcommand{\hbBTwoCI}{[-0.058,-0.035]}
\newcommand{\hbBTwoFvd}{2.63}
\newcommand{\hbCwGain}{+0.047}
\newcommand{\hbCwDiff}{+0.060}
\newcommand{\hbCwCI}{[+0.047,+0.073]}
\newcommand{\hbCwFvd}{0.79}
\newcommand{\hbGainGain}{-0.001}
\newcommand{\hbGainDiff}{+0.108}
\newcommand{\hbGainCI}{[+0.091,+0.126]}
\newcommand{\hbGainFvd}{1.18}
\newcommand{\hbContrGain}{-0.003}
\newcommand{\hbContrDiff}{+0.110}
\newcommand{\hbContrCI}{[+0.090,+0.132]}

\newcommand{\hbFaGain}{-0.042}
\newcommand{\hbFaDiff}{+0.149}
\newcommand{\hbFaCI}{[+0.131,+0.168]}

\newcommand{\hbParamBig}{22.9M}
\newcommand{\hbParamSmall}{1.25M}
\newcommand{\hbParamRatio}{18}
\newcommand{\mcPoint}{+0.096}
\newcommand{\mcCtx}{[+0.082,+0.109]}
\newcommand{\mcDemo}{[+0.084,+0.109]}
\newcommand{\mcTask}{[+0.076,+0.117]}
\newcommand{\mcN}{236}
\newcommand{\hmM}{8}
\newcommand{\hmMaxPCtx}{0.0098}

\newcommand{\hmMaxPDemo}{0.0125}

\newcommand{\hmMaxPTask}{0.0358}

\newcommand{\idrGain}{+0.129}
\newcommand{\idrGainCI}{[+0.106,+0.152]}
\newcommand{\idrGainDemoCI}{[+0.104,+0.153]}
\newcommand{\idrGainTaskCI}{[+0.088,+0.163]}
\newcommand{\idrFull}{+0.127}
\newcommand{\idrFullCI}{[+0.104,+0.152]}
\newcommand{\idrFullDemoCI}{[+0.104,+0.151]}
\newcommand{\idrFullTaskCI}{[+0.092,+0.157]}
\newcommand{\idrKeep}{+0.002}
\newcommand{\idrKeepCI}{[-0.008,+0.011]}
\newcommand{\idrKeepDemoCI}{[-0.009,+0.012]}
\newcommand{\idrKeepTaskCI}{[-0.008,+0.012]}

\newcommand{\idrScHalf}{+0.068}
\newcommand{\idrScHalfCI}{[+0.049,+0.087]}
\newcommand{\idrScHalfDemoCI}{[+0.048,+0.089]}
\newcommand{\idrScHalfTaskCI}{[+0.036,+0.098]}

\newcommand{\idrFresh}{+0.124}
\newcommand{\idrFreshCI}{[+0.094,+0.154]}
\newcommand{\idrFreshDemoCI}{[+0.096,+0.153]}
\newcommand{\idrFreshTaskCI}{[+0.093,+0.151]}
\newcommand{\idrSealed}{+0.134}
\newcommand{\idrSealedCI}{[+0.100,+0.170]}
\newcommand{\idrSealedDemoCI}{[+0.096,+0.172]}
\newcommand{\idrSealedTaskCI}{[+0.077,+0.180]}
\newcommand{\idrN}{236}
\newcommand{\idrNFresh}{117}
\newcommand{\idrNSealed}{119}
\newcommand{\idrBetter}{160}
\newcommand{\idrWorse}{50}
\newcommand{\idrNat}{0.369}
\newcommand{\idrSched}{0.499}
\newcommand{\idrGT}{0.664}
\newcommand{\idrGap}{0.44}
\newcommand{\idrSeedA}{+0.133}
\newcommand{\idrSeedB}{+0.125}
\newcommand{\idrTrainMin}{+0.125}
\newcommand{\idrTrainMax}{+0.133}
\newcommand{\idrRho}{0.95}
\newcommand{\idrNull}{-0.025}

\newcommand{\idrDevAcc}{0.682}
\newcommand{\idrDevMinRec}{0.462}
\newcommand{\idrDevBin}{0.833}
\newcommand{\idrDevBlur}{0.677}
\newcommand{\idrDevBlurRatio}{0.992}

\newcommand{\idrTrainCtx}{781}
\newcommand{\idrTrainSamples}{3905}

\newcommand{\cfEpisodes}{2000}

\newcommand{\cfNI}{0.03}
\newcommand{\cfNat}{+0.032}
\newcommand{\cfNatCI}{[+0.014,+0.050]}
\newcommand{\cfP}{0.0004}
\newcommand{\cfNIP}{<0.001}
\newcommand{\cfSeedA}{+0.029}
\newcommand{\cfSeedB}{+0.035}
\newcommand{\cfRateBase}{0.808}
\newcommand{\cfRateNat}{0.799}
\newcommand{\cfRateOracle}{0.857}
\newcommand{\cfRateSched}{0.831}
\newcommand{\cfBase}{+0.023}
\newcommand{\cfBaseCI}{[+0.005,+0.041]}

\newcommand{\cfNoOne}{+0.020}
\newcommand{\cfNoOneCI}{[+0.002,+0.038]}
\newcommand{\cfNoOneP}{0.015}
\newcommand{\cfOne}{+0.14}
\newcommand{\cfOneCI}{[+0.07,+0.22]}

\newcommand{\cfHead}{+0.058}
\newcommand{\cfHeadCI}{[+0.016,+0.101]}

\newcommand{\cfTasksPos}{8}

\newcommand{\cfNatBase}{-0.009}
\newcommand{\cfNatBaseCI}{[-0.030,+0.012]}

\newcommand{\rlFdSched}{1.36}
\newcommand{\rlFdSchedCI}{[1.24,1.41]}
\newcommand{\rlFdFull}{1.63}
\newcommand{\rlFdFullCI}{[1.48,1.69]}
\newcommand{\rlLpSched}{0.72}
\newcommand{\rlLpSchedCI}{[0.70,0.73]}
\newcommand{\rlLpFull}{0.75}
\newcommand{\rlLpFullCI}{[0.73,0.76]}
\newcommand{\rlShSched}{0.93}
\newcommand{\rlShSchedCI}{[0.92,0.93]}
\newcommand{\rlShFull}{0.87}
\newcommand{\rlShFullCI}{[0.87,0.88]}
\newcommand{\rlFdNat}{62.2}

\newcommand{\rlFdQuarter}{0.97}
\newcommand{\rlLpQuarter}{0.92}
\newcommand{\rlShQuarter}{0.99}
\newcommand{\rlFdHalf}{0.99}
\newcommand{\rlLpHalf}{0.84}
\newcommand{\rlShHalf}{0.98}
\newcommand{\rlFdThreeQ}{1.12}
\newcommand{\rlLpThreeQ}{0.75}
\newcommand{\rlShThreeQ}{0.96}
\newcommand{\rlLpNat}{0.153}
\newcommand{\rlShShareSched}{72}
\newcommand{\rlShShareFull}{95}
\newcommand{\rlShShareNull}{3}

\newcommand{\rlLpShareNull}{28}
\newcommand{\ctN}{236}
\newcommand{\ctMargin}{+0.125}
\newcommand{\ctMarginCI}{[+0.114,+0.136]}
\newcommand{\ctSeedA}{+0.124}
\newcommand{\ctSeedB}{+0.125}
\newcommand{\ctGuard}{0.83}
\newcommand{\ctGuardHi}{0.84}
\newcommand{\ctPair}{+0.246}
\newcommand{\ctPairCI}{[+0.215,+0.277]}

\newcommand{\ctMarginNat}{-0.023}
\newcommand{\ctMarginAd}{+0.102}

\newcommand{\cdCos}{0.13}
\newcommand{\cdPair}{0.48}
\newcommand{\cdPairCI}{[0.41,0.55]}
\newcommand{\cdMargin}{-0.010}
\newcommand{\wdCos}{0.49}
\newcommand{\wdPair}{0.83}
\newcommand{\wdPairCI}{[0.77,0.89]}
\newcommand{\wdMargin}{0.19}
\newcommand{\wdSnr}{3.0}
\newcommand{\wdN}{35}

\newcommand{\fsTraj}{70}
\newcommand{\fsQueries}{1607}

\newcommand{\fsRegNatFour}{0.036}

\newcommand{\fsShrinkEight}{0.07}

\newcommand{\fsSeedsPosEight}{4}

\newcommand{\fsShrinkSixteen}{0.09}

\newcommand{\fsRegNatSixteen}{0.065}
\newcommand{\fsSeedsPosSixteen}{4}

\newcommand{\cfTaskP}{0.04}
\newcommand{\suN}{423}
\newcommand{\suPos}{341}
\newcommand{\suMargin}{+0.060}
\newcommand{\suMarginCI}{[+0.053,+0.067]}
\newcommand{\suKeep}{-0.001}

\newcommand{\suPair}{+0.144}
\newcommand{\suPairCI}{[+0.126,+0.162]}
\newcommand{\suFvd}{1.15}
\newcommand{\suFvdCI}{[1.03,1.22]}
\newcommand{\suFvdFull}{2.01}

\newcommand{\suSpatialN}{105}
\newcommand{\suSpatialMargin}{+0.043}
\newcommand{\suSpatialPair}{+0.154}
\newcommand{\suSpatialFvd}{1.07}
\newcommand{\suObjectN}{101}
\newcommand{\suObjectMargin}{+0.076}
\newcommand{\suObjectPair}{+0.147}
\newcommand{\suObjectFvd}{1.04}
\newcommand{\suGoalN}{110}
\newcommand{\suGoalMargin}{+0.047}
\newcommand{\suGoalPair}{+0.137}
\newcommand{\suGoalFvd}{1.21}
\newcommand{\suNinetyN}{107}
\newcommand{\suNinetyMargin}{+0.073}
\newcommand{\suNinetyPair}{+0.140}
\newcommand{\suNinetyFvd}{1.11}
\newcommand{\hxRaters}{16}
\newcommand{\hxValid}{11}
\newcommand{\hxExcl}{5}
\newcommand{\hxN}{74}
\newcommand{\hxK}{55}
\newcommand{\hxOther}{19}
\newcommand{\hxTies}{4}
\newcommand{\hxNone}{2}
\newcommand{\hxShare}{0.74}
\newcommand{\hxCI}{[0.63,0.84]}
\newcommand{\hxP}{<0.001}

\newcommand{\hxLooMin}{0.61}
\newcommand{\hxLooRuns}{11}
\newcommand{\hxCT}{44}

\newcommand{\hxRaterMin}{0.57}
\newcommand{\hxRaterMax}{0.87}

\newcommand{\fuEpisodes}{1500}

\newcommand{\fuRateBase}{0.619}
\newcommand{\fuRateNat}{0.607}
\newcommand{\fuRateOracle}{0.736}
\newcommand{\fuRateSched}{0.678}
\newcommand{\fuNat}{+0.071}
\newcommand{\fuNatCI}{[+0.043,+0.099]}

\newcommand{\fuP}{0.010}
\newcommand{\fuSeedA}{+0.067}
\newcommand{\fuSeedB}{+0.075}
\newcommand{\fuShare}{55}

\newcommand{\fuBase}{+0.058}
\newcommand{\fuBaseCI}{[+0.029,+0.088]}
\newcommand{\fuNatBase}{-0.013}
\newcommand{\fuNatBaseCI}{[-0.045,+0.020]}

\newcommand{\fuOracleBase}{+0.117}
\newcommand{\fuTasksPos}{8}

\newcommand{\fuOne}{+0.180}
\newcommand{\fuOneCI}{[+0.083,+0.273]}
\newcommand{\fuNoOne}{+0.059}
\newcommand{\fuNoOneCI}{[+0.029,+0.089]}
\newcommand{\fuNoOneP}{0.020}

\newcommand{\fuPilotHead}{0.16}

\newcommand{\fuPilotDefHead}{0.077}

\newcommand{\dscN}{1500}
\newcommand{\dscRhoNat}{0.09}
\newcommand{\dscRhoSch}{0.35}
\newcommand{\dscRhoDiff}{+0.26}
\newcommand{\dscRhoCI}{[+0.21,+0.32]}
\newcommand{\dscTopNat}{0.32}
\newcommand{\dscTopSch}{0.44}
\newcommand{\dscTopDiff}{+0.13}

\newcommand{\dscTopTasks}{9}

\newcommand{\dsdN}{2000}
\newcommand{\dsdRhoNat}{0.06}
\newcommand{\dsdRhoSch}{0.26}
\newcommand{\dsdRhoDiff}{+0.20}
\newcommand{\dsdRhoCI}{[+0.16,+0.25]}
\newcommand{\dsdTopNat}{0.29}
\newcommand{\dsdTopSch}{0.39}
\newcommand{\dsdTopDiff}{+0.10}

\newcommand{\dsdTopTasks}{8}

\newcommand{\fqEpisodes}{1000}

\newcommand{\fqRateBase}{0.626}
\newcommand{\fqRateNat}{0.613}
\newcommand{\fqRateOracle}{0.728}
\newcommand{\fqRateSched}{0.656}
\newcommand{\fqNat}{+0.043}
\newcommand{\fqNatCI}{[+0.009,+0.078]}

\newcommand{\fqP}{0.068}

\newcommand{\fqAdMin}{+0.021}
\newcommand{\fqAdMax}{+0.060}

\newcommand{\fqBase}{+0.030}
\newcommand{\fqBaseCI}{[-0.003,+0.064]}
\newcommand{\fqNatBase}{-0.013}
\newcommand{\fqNatBaseCI}{[-0.052,+0.026]}

\newcommand{\fqOne}{+0.265}
\newcommand{\fqOneCI}{[+0.168,+0.362]}
\newcommand{\fqNoOne}{+0.018}
\newcommand{\fqNoOneCI}{[-0.018,+0.053]}
\newcommand{\fqNoOneP}{0.137}

\newcommand{\fqRhoNat}{0.08}
\newcommand{\fqRhoSch}{0.34}
\newcommand{\fqRhoDiff}{+0.26}
\newcommand{\fqRhoCI}{[+0.20,+0.32]}

\newcommand{\fqTopNat}{0.30}
\newcommand{\fqTopSch}{0.43}

\newcommand{\rzEpisodes}{2500}

\newcommand{\rzRateBase}{0.637}
\newcommand{\rzRateNat}{0.620}
\newcommand{\rzRateOracle}{0.733}
\newcommand{\rzRateSched}{0.667}
\newcommand{\rzNat}{+0.047}
\newcommand{\rzNatCI}{[+0.026,+0.069]}

\newcommand{\rzP}{0.014}

\newcommand{\rzAdMin}{+0.039}
\newcommand{\rzAdMax}{+0.059}

\newcommand{\rzBase}{+0.030}
\newcommand{\rzBaseCI}{[+0.008,+0.053]}
\newcommand{\rzNatBase}{-0.017}
\newcommand{\rzNatBaseCI}{[-0.042,+0.007]}

\newcommand{\rzOne}{+0.146}
\newcommand{\rzOneCI}{[+0.083,+0.210]}

\newcommand{\rzRhoNat}{0.06}
\newcommand{\rzRhoSch}{0.36}
\newcommand{\rzRhoDiff}{+0.29}
\newcommand{\rzRhoCI}{[+0.25,+0.34]}
\newcommand{\rzRhoTasks}{10}
\newcommand{\rzRhoP}{0.001}
\newcommand{\rzTopNat}{0.30}
\newcommand{\rzTopSch}{0.45}
\newcommand{\rzTwoDiff}{+0.036}
\newcommand{\rzTwoCI}{[+0.014,+0.059]}
\newcommand{\rzTwoP}{0.027}

\newcommand{\pqMean}{+0.047}
\newcommand{\pqCI}{[+0.016,+0.083]}
\newcommand{\pqP}{0.012}
\newcommand{\pqPos}{8}

\newcommand{\pqNoOne}{+0.034}
\newcommand{\pqNoOneP}{0.023}

\newcommand{\dzContexts}{2000}

\newcommand{\dzSlope}{+0.065}
\newcommand{\dzSlopeCI}{[+0.036,+0.094]}

\newcommand{\dzSignP}{0.006}
\newcommand{\dzTasksPos}{9}

\newcommand{\dzSumHalf}{78}
\newcommand{\dzSumFull}{123}
\newcommand{\dzGainHalf}{+0.039}
\newcommand{\dzGainHalfCI}{[+0.012,+0.066]}
\newcommand{\dzGainFull}{+0.061}
\newcommand{\dzGainFullCI}{[+0.034,+0.089]}

\newcommand{\dzRateBase}{0.635}
\newcommand{\dzRateNat}{0.601}
\newcommand{\dzRateOracle}{0.727}
\newcommand{\dzRateHalf}{0.639}
\newcommand{\dzRateFull}{0.662}

\newcommand{\dzTwoSlope}{+0.060}
\newcommand{\dzTwoSlopeCI}{[+0.030,+0.091]}
\newcommand{\dzTwoSignP}{0.012}

\newcommand{\dzEnd}{+0.061}
\newcommand{\dzEndCI}{[+0.034,+0.089]}
\newcommand{\dzEndP}{0.007}

\newcommand{\dzEndAdA}{+0.058}

\newcommand{\dzEndAdD}{+0.040}

\newcommand{\dzEndTwoDiff}{+0.054}
\newcommand{\dzEndTwoCI}{[+0.025,+0.083]}

\newcommand{\dzRhoFull}{0.34}
\newcommand{\dzRhoFullDiff}{+0.25}
\newcommand{\dzRhoFullCI}{[+0.20,+0.29]}
\newcommand{\dzRhoNat}{0.10}

\newcommand{\awcPairShare}{0.626}
\newcommand{\awcContexts}{2000}
\newcommand{\awcClusters}{500}

\newcommand{\awcRateA}{0.679}
\newcommand{\awcRateBase}{0.635}
\newcommand{\awcRateNat}{0.601}
\newcommand{\awcRateRep}{0.662}
\newcommand{\awcRateOracle}{0.727}

\newcommand{\awcTwo}{+0.043}

\newcommand{\awcShare}{0.78}
\newcommand{\awcShareCI}{[0.45,1.34]}

\newcommand{\awcRhoA}{0.75}
\newcommand{\awcRhoNat}{0.10}
\newcommand{\awcRhoRep}{0.33}
\newcommand{\awcTopA}{0.71}
\newcommand{\awcTopNat}{0.34}
\newcommand{\awcTopRep}{0.43}

\newcommand{\ovPoolN}{800}
\newcommand{\ovPoolGain}{+0.028}
\newcommand{\ovPoolGainCI}{[-0.016,+0.071]}
\newcommand{\ovPoolBase}{-0.014}
\newcommand{\ovPoolBaseCI}{[-0.056,+0.030]}
\newcommand{\ovOtherN}{1200}
\newcommand{\ovOtherGain}{+0.084}
\newcommand{\ovOtherGainCI}{[+0.050,+0.120]}
\newcommand{\ovOtherBase}{+0.053}
\newcommand{\ovOtherBaseCI}{[+0.019,+0.087]}

\newcommand{\ctlQContexts}{1790}
\newcommand{\ctlRhoCtl}{0.15}
\newcommand{\ctlRhoNat}{0.10}
\newcommand{\ctlRhoRep}{0.32}
\newcommand{\ctlRhoDiff}{+0.18}
\newcommand{\ctlRhoCI}{[+0.14,+0.21]}
\newcommand{\ctlRhoTasksPos}{10}

\newcommand{\ctlTopCtl}{0.32}
\newcommand{\ctlTopNat}{0.34}
\newcommand{\ctlTopRep}{0.43}
\newcommand{\ctlTopDiff}{+0.12}
\newcommand{\ctlTopCI}{[+0.09,+0.14]}
\newcommand{\ctlOfN}{236}
\newcommand{\ctlOfDemos}{60}
\newcommand{\ctlOfCtlMarg}{-0.053}
\newcommand{\ctlOfCtlMargCI}{[-0.067,-0.039]}
\newcommand{\ctlOfRepMarg}{+0.096}
\newcommand{\ctlOfRepMargCI}{[+0.083,+0.108]}
\newcommand{\ctlOfDiff}{-0.149}
\newcommand{\ctlOfDiffCI}{[-0.160,-0.138]}
\newcommand{\ctlOfPair}{-0.087}
\newcommand{\ctlOfPairCI}{[-0.119,-0.054]}
\newcommand{\ctlOfPairRep}{+0.163}
\newcommand{\ctlOfClsCtl}{-0.004}

\newcommand{\ctlOfClsCtlCI}{[-0.027,+0.021]}
\newcommand{\ctlOfLpCtl}{0.82}
\newcommand{\ctlOfLpRep}{0.72}

\newcommand{\ctlOfFvdCtl}{1.88}

\newcommand{\ctlOfFvdRep}{1.02}

\newcommand{\gsContexts}{236}
\newcommand{\gsRollouts}{7080}
\newcommand{\gsSchedFullMarg}{+0.094}
\newcommand{\gsSchedFullCls}{+0.124}
\newcommand{\gsSchedFullFd}{1.34}
\newcommand{\gsSchedFullSh}{0.92}
\newcommand{\gsSchedFullFvd}{1.01}

\newcommand{\gsSchedThreeQMarg}{+0.068}
\newcommand{\gsSchedThreeQCls}{+0.108}
\newcommand{\gsSchedThreeQFd}{1.11}
\newcommand{\gsSchedThreeQSh}{0.96}

\newcommand{\gsSchedHalfMarg}{+0.042}
\newcommand{\gsSchedHalfCls}{+0.067}
\newcommand{\gsSchedHalfFd}{0.98}
\newcommand{\gsSchedHalfSh}{0.98}

\newcommand{\gsAlwaysFullMarg}{+0.095}
\newcommand{\gsAlwaysFullCls}{+0.124}
\newcommand{\gsAlwaysFullFd}{1.61}
\newcommand{\gsAlwaysFullSh}{0.87}
\newcommand{\gsAlwaysFullFvd}{1.58}

\newcommand{\gsAlwaysThreeQMarg}{+0.070}
\newcommand{\gsAlwaysThreeQCls}{+0.108}
\newcommand{\gsAlwaysThreeQFd}{1.21}
\newcommand{\gsAlwaysThreeQSh}{0.92}

\newcommand{\gsAlwaysHalfMarg}{+0.043}
\newcommand{\gsAlwaysHalfCls}{+0.073}
\newcommand{\gsAlwaysHalfFd}{1.01}
\newcommand{\gsAlwaysHalfSh}{0.95}

\newcommand{\gsCutThreeMarg}{+0.053}
\newcommand{\gsCutThreeCls}{+0.087}
\newcommand{\gsCutThreeFd}{1.34}
\newcommand{\gsCutThreeSh}{1.06}

\newcommand{\gsDiffThreeQFd}{-0.107}
\newcommand{\gsDiffThreeQFdCI}{[-0.119,-0.082]}
\newcommand{\gsDiffThreeQSh}{+0.039}
\newcommand{\gsDiffThreeQShCI}{[+0.034,+0.044]}
\newcommand{\gsDiffThreeQMarg}{-0.002}
\newcommand{\gsDiffThreeQMargCI}{[-0.002,-0.001]}
\newcommand{\gsDiffHalfFd}{-0.031}
\newcommand{\gsDiffHalfFdCI}{[-0.038,-0.021]}
\newcommand{\gsDiffHalfSh}{+0.025}
\newcommand{\gsDiffHalfShCI}{[+0.022,+0.028]}
\newcommand{\gsDiffHalfMarg}{-0.001}
\newcommand{\gsDiffHalfMargCI}{[-0.001,-0.001]}

\newcommand{\sdSeedPooledLo}{+0.094}
\newcommand{\sdSeedPooledHi}{+0.099}

\newcommand{\qlCosNatHi}{0.05}
\newcommand{\qlCosOurLo}{0.47}
\newcommand{\qlCosOurHi}{0.70}
\providecommand{\rwTBD}{\textcolor{red}{??}}

\newcommand{\rwNEval}{50}

\newcommand{\rwSceneA}{Lemon in Bowl}\newcommand{\rwSceneB}{Cup and Banana}
\newcommand{\rwSceneC}{Pot on Stove}\newcommand{\rwSceneD}{Spoon on Towel}
\newcommand{\rwArmA}{w/o \ours{}}\newcommand{\rwArmB}{w/ \ours{}}
\newcommand{\rwScAA}{24}\newcommand{\rwScAB}{40}
\newcommand{\rwScBA}{7}\newcommand{\rwScBB}{31}
\newcommand{\rwScCA}{35}\newcommand{\rwScCB}{43}
\newcommand{\rwScDA}{26}\newcommand{\rwScDB}{39}
\edef\rwSumA{\the\numexpr\rwScAA+\rwScBA+\rwScCA+\rwScDA\relax}
\edef\rwSumB{\the\numexpr\rwScAB+\rwScBB+\rwScCB+\rwScDB\relax}
\edef\rwNAll{\the\numexpr4*\rwNEval\relax}
\newcommand{\rwPct}[2]{\pgfmathparse{100*((#1)/(#2))}\pgfmathprintnumber[fixed,precision=1,zerofill]{\pgfmathresult}}
\newcommand{\rwCellN}[2]{\ifnum#1<0 \rwTBD\else#1\,(\rwPct{#1}{#2})\fi}
\newcommand{\rwDelta}[3]{\ifnum#1<0 \rwTBD\else\ifnum#2<0 \rwTBD\else\pgfmathparse{100*(((#2)-(#1))/(#3))}\pgfmathprintnumber[fixed,precision=1,zerofill,showpos]{\pgfmathresult}\fi\fi}

\newcommand{\rwWil}[5]{\ifnum#4<0 \pgfmathsetmacro{#1}{0}\pgfmathsetmacro{#2}{0}\pgfmathsetmacro{#3}{0}\else
  \pgfmathsetmacro{\rwtp}{#4/#5}\pgfmathsetmacro{\rwtc}{(\rwtp+1.9208/#5)/(1+3.8416/#5)}%
  \pgfmathsetmacro{\rwth}{1.96*sqrt(\rwtp*(1-\rwtp)/#5+(0.9604/#5)/#5)/(1+3.8416/#5)}%
  \pgfmathsetmacro{#1}{100*\rwtp}\pgfmathsetmacro{#2}{100*(\rwtp-(\rwtc-\rwth))}\pgfmathsetmacro{#3}{100*((\rwtc+\rwth)-\rwtp)}\fi}
\newif\ifrwAll\rwAlltrue
\def\rwchk#1{\ifnum#1<0 \rwAllfalse\fi}
\rwchk{\rwScAA}\rwchk{\rwScAB}\rwchk{\rwScBA}\rwchk{\rwScBB}\rwchk{\rwScCA}\rwchk{\rwScCB}\rwchk{\rwScDA}\rwchk{\rwScDB}

\newif\ifrwFresh\rwFreshtrue
\def\rwcmp#1#2{\ifnum#1=#2 \else\rwFreshfalse\fi}
\rwcmp{\rwScAA}{24}\rwcmp{\rwScAB}{40}\rwcmp{\rwScBA}{7}\rwcmp{\rwScBB}{31}\rwcmp{\rwScCA}{35}\rwcmp{\rwScCB}{43}\rwcmp{\rwScDA}{26}\rwcmp{\rwScDB}{39}
\newcommand{\rwStat}[1]{\ifrwFresh#1\else\rwTBD\fi}

\newcommand{\rwPvCmh}{$<$0.001}

\newcommand{\ceRateOracleR}{0.885}
\newcommand{\ceNatR}{+0.019}

\newcommand{\mtMargWs}{-0.004}
\newcommand{\mtPairWs}{-0.007}
\newcommand{\mtObjWs}{+0.6}
\newcommand{\mtContWs}{+0.5}
\newcommand{\mtMargFrg}{-0.001}
\newcommand{\mtPairFrg}{-0.004}
\newcommand{\mtObjFrg}{-0.8}
\newcommand{\mtContFrg}{-1.2}
\newcommand{\mtMargFro}{+0.002}
\newcommand{\mtPairFro}{+0.011}
\newcommand{\mtObjFro}{0.0}
\newcommand{\mtContFro}{-0.6}
\newcommand{\mtMargBOne}{+0.071}
\newcommand{\mtPairBOne}{+0.130}
\newcommand{\mtObjBOne}{+1.5}
\newcommand{\mtContBOne}{+1.4}
\newcommand{\mtMargBTwo}{+0.154}
\newcommand{\mtPairBTwo}{+0.246}
\newcommand{\mtObjBTwo}{-14.9}
\newcommand{\mtContBTwo}{-17.6}
\newcommand{\mtMargCw}{+0.047}
\newcommand{\mtPairCw}{+0.075}
\newcommand{\mtObjCw}{+3.6}
\newcommand{\mtContCw}{+2.6}
\newcommand{\mtMargGain}{-0.001}
\newcommand{\mtPairGain}{-0.050}
\newcommand{\mtObjGain}{+0.9}
\newcommand{\mtContGain}{+1.1}
\newcommand{\mtMargContr}{-0.003}
\newcommand{\mtPairContr}{-0.002}
\newcommand{\mtObjContr}{+15.3}
\newcommand{\mtContContr}{+25.6}
\newcommand{\mtMargFa}{-0.042}
\newcommand{\mtPairFa}{-0.140}
\newcommand{\mtObjFa}{+29.2}
\newcommand{\mtContFa}{+12.5}
\newcommand{\mtMargOpie}{+0.107}
\newcommand{\mtPairOpie}{+0.185}
\newcommand{\mtObjOpie}{-7.9}
\newcommand{\mtContOpie}{-9.0}

\newcommand{\mtNatMarg}{0.012}
\newcommand{\mtNatPair}{0.537}
\newcommand{\mtNatObj}{70.0}
\newcommand{\mtNatCont}{66.7}

\newcommand{\mtAblGtnObj}{+1.9}
\newcommand{\mtAblGtnCont}{+0.3}
\newcommand{\mtAblNodirObj}{-6.5}
\newcommand{\mtAblNodirCont}{-7.3}
\newcommand{\mtAblGtfitObj}{-2.4}
\newcommand{\mtAblGtfitCont}{-2.9}
\newcommand{\mtAblRecObj}{-6.6}
\newcommand{\mtAblRecCont}{-7.4}

\begin{document}
\title{A Case of Mistaken Identity:\\Repairing Consequence Identity in World Action Models}
\author{%
Zhiruo Zhou$^{1,2}$\quad Zelin Li$^{1}$\quad Chen Xiwen$^{1,3}$\quad Yucheng Chen$^{2}$\quad Xiaojun Zhu$^{1}$\quad Houde Liu$^{1}$\\[2pt]
$^{1}$Tsinghua University\quad $^{2}$Wuhan University of Technology\quad $^{3}$Shanghai Jiao Tong University}
\maketitle
\begin{strip}
\vspace{-22pt}
\centering
\includegraphics{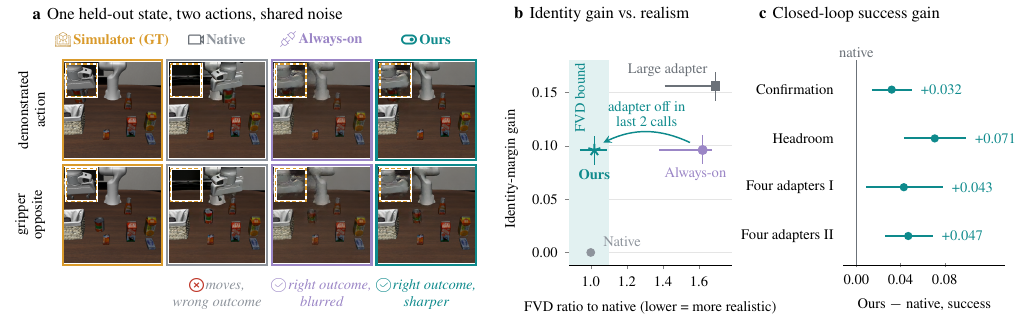}
\captionsetup{hypcap=false}
\captionof{figure}{\textbf{Responding to an action is not predicting its consequence.} (a) Native Prophet moves the arm yet reaches another action's object outcome, always-on repair reaches the simulator's outcome with blurred frames, and ours, the same adapter switched off for the last denoising calls, keeps the outcome with sharper frames; dashed boxes mark the simulator's gripper. (b) Identity gain against FVD ratio to native on 236 held-out contexts; shaded: registered FVD bound. (c) Gain of ours over native in closed-loop success, four registered tests.}
\captionsetup{hypcap=true}
\label{fig:teaser}
\end{strip}

\begin{abstract}
World action models (WAMs) rank robot actions by imagining their consequences, so each action must produce its own outcome. We find that WAMs for policy reinforcement learning move the arm as commanded yet can show the object outcome of a different action. We propose OPIE (\underline{O}n-\underline{P}olicy \underline{I}ntervention-\underline{E}ffect tuning), an offline repair of the action pathway. OPIE trains a small action-embedder adapter on branch differences at the states the sampler itself visits. It switches the adapter on only in the early high-noise denoising steps, where the outcome forms. It adds $0.06\%$ of the parameters and no denoising call, and it transfers across WAM backbones. On LIBERO-10, OPIE lifts the identity margin sevenfold (from $0.016$ to $0.112$) within a preset FVD bound. A DINOv2 classifier assigns video endpoints to the correct action $35\%$ more often, and blind raters prefer the repaired outcomes in $74\%$ of judged cases. On a separately trained Cosmos Policy, OPIE improves pairwise accuracy by $56\%$. In closed-loop action selection, OPIE raises success over native in all four registered tests, by $4$--$12\%$ in relative terms. In a pre-specified readout, the rank correlation of its imagined scores with simulator outcomes rises from $\rzRhoNat{}$ to $\rzRhoSch{}$. On a real robot, policies trained in the repaired WAM raise success in all four scenes (\rwPct{\rwSumA}{\rwNAll}$\%$ to \rwPct{\rwSumB}{\rwNAll}$\%$ pooled).
\end{abstract}

\section{Introduction}
\label{sec:intro}
World action models (WAMs) build on video generators to predict how the world evolves under robot actions, often together with the actions themselves~\cite{dreamzero2026,uwm2025,cosmospolicy2026}. They are used to post-train policies~\cite{prophrl2025}, evaluate behavior~\cite{worldeval2025,worldgym2025,veo_eval2025}, rank candidate actions~\cite{gpc2025,forewarn2025,nwm2025,ttswam2026}, and synthesize training data~\cite{dreamgen2025}. These uses require more than plausible videos: when an action changes, its predicted consequence must change in the right way. A model can move the arm as commanded yet show the object outcome of a different action. The consequence is the signal in each of these uses; swapped consequences reward wrong behavior and misrank actions. On Prophet~\cite{prophrl2025}, the imagined scores of candidate actions are almost uncorrelated with their simulated outcomes (rank correlation $\rzRhoNat{}$).

\begin{figure*}[t]
\centering
\includegraphics[width=\textwidth]{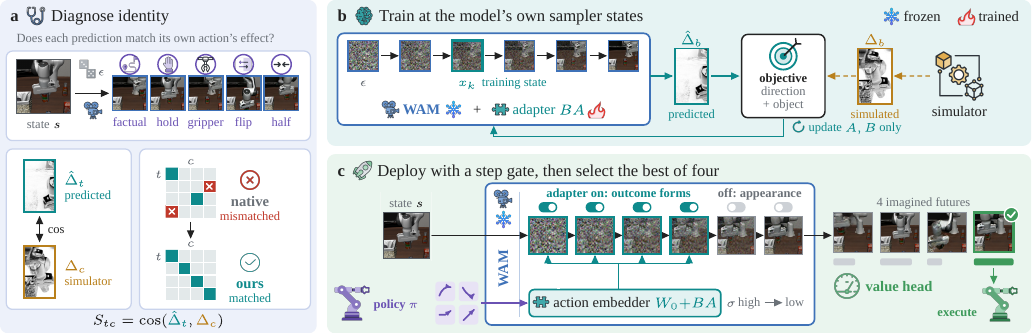}
\caption{\textbf{OPIE repairs the action pathway of a WAM and uses it only at high noise.} (a) Diagnose, (b) train, (c) deploy. Frames come from one held-out context; matrices, noise levels and scores are schematic.}
\label{fig:method}
\end{figure*}

Perceptual, physics and action-fidelity benchmarks measure plausibility or action following on single rollouts~\cite{vbench2024,worldmodelbench2025,physicsiq2025,mirabench2026,ewmbench2025}, and recent protocols ask whether the model responds to a changed action~\cite{worldsync2026,coco2026}. None checks whether each action's predicted change is its own, so a model can pass them all and still swap the outcomes of two actions. We separate \emph{responsiveness} from \emph{consequence fidelity}. From one simulator state, the model imagines several candidate actions under the same noise, and we compare each predicted change with the simulator's (Sec.~\ref{sec:diagnosis}). A model is \emph{responsive} if these changes have the right size, and \emph{consequence-correct} if each matches its own action's change rather than another's. We call the latter property \emph{identity}. On three released WAMs, the diagnostic gives a first finding: Prophet responds at about the right magnitude but partly aligned directions; UWM's~\cite{uwm2025} effects are below seed noise; A2World~\cite{a2world2026} has the highest identity (Table~\ref{tab:diag}).

This diagnosis motivates \ours{}. It trains a small action-embedder LoRA~\cite{lora2022} on branch-difference directions and object consequences, at the states the model's own sampler visits. The standard signal, denoising a noised ground-truth video, gives little supervision here, and the diagnosis shows why. At low noise the input already reveals much of the outcome, and it differs from the states the sampler visits at test time; this is the exposure mismatch known from diffusion and autoregressive video~\cite{ning2023input,selfforcing2025}. However, keeping the adapter on through the final calls raises FVD~\cite{fvd2018}. The repair must change the outcome but not the appearance.

Our answer is \textbf{step-scheduled deployment}. The frozen adapter is on at high noise, where the branch outcome forms, and off in the final calls, where appearance is rendered. We select the cutoff per model on development data and freeze it before reading held-out data; no denoiser call or training step is added. In registered tests on held-out contexts, this keeps the margin gain, brings FVD back within a preset bound of native and carries over to policy-proposed actions. In closed-loop action selection, it raises ten-task success over native on new policy seeds (Fig.~\ref{fig:teaser}, Sec.~\ref{sec:exp}). The same loss also repairs a separately trained WAM, Cosmos Policy~\cite{cosmospolicy2026}. On a real robot, policies trained in the repaired WAM succeed more often in all four scenes (Fig.~\ref{fig:real}).

Our contributions are the following:
\begin{itemize}
\item We show that a WAM can respond to actions yet swap their consequences, an error of direction that current benchmarks miss, and give a five-action test that catches it.
\item We present \ours{}, which repairs identity by two choices: training at the sampler's own states and turning the adapter off at low noise. Each answers a finding: noised ground truth reveals the outcome; appearance forms late.
\item Within a preset FVD bound, \ours{} gives the largest identity gain of all baselines. The repaired WAM ranks candidates better, picks better actions in four closed-loop tests and closes most of the gap to a WAM built to link actions to their consequences.
\end{itemize}

\section{Related Work}
\label{sec:related}

\begin{table}[t]
\centering
\caption{\textbf{Native diagnosis} of three WAMs on the 39 development contexts. Bold: best per column; $^\ddagger$latent space.}
\label{tab:diag}
\footnotesize
\setlength{\tabcolsep}{3.4pt}
\begin{tabular}{@{}lcccccc@{}}
\toprule
 & \multicolumn{2}{c}{Responsiveness} & \multicolumn{2}{c}{Direction} & \multicolumn{2}{c}{Identity} \\
\cmidrule(lr){2-3}\cmidrule(lr){4-5}\cmidrule(l){6-7}
Model & SNR & Amp. & Cos.$\uparrow$ & Cont./zero$\downarrow$ & PairAcc$\uparrow$ & Margin$\uparrow$ \\
\midrule
A2World & $\dgASnr$ & $\dgAAmp$ & $\mathbf{\dgACos}$ & $\dgAZero$ & $\mathbf{\dgAPairAcc}$ & $\mathbf{\dgAMargin}$ \\
Prophet & $\dgPSnr$ & $\dgPAmp$ & $\dgPCos$ & $\dgPZero$ & $\dgPPairAcc$ & $\dgPMargin$ \\
UWM$^\ddagger$ & $\dgUSnr$ & $\dgUAmp$ & $\dgUCos$ & $\mathbf{\dgUZero}$ & $\dgUPairAcc$ & $\dgUMargin$ \\
\bottomrule
\end{tabular}
\end{table}

\paragraph{World action models.}
Video generators conditioned on low-level actions serve as interactive simulators~\cite{unisim2024,genie2024}. Robot WAMs build on large video backbones~\cite{cosmospredict25_2025}, inject actions through adapters or adaLN~\cite{avid2024,dws2025,irasim2025,vid2world2026,ctrlworld2026,genieenvisioner2026}, couple video and action diffusion~\cite{uwm2025,worldvla2025,dreamzero2026}, use video prediction as the policy backbone~\cite{vpp2025}, or pretrain on millions of trajectories~\cite{a2world2026}. Recent systems strengthen generation or control~\cite{cosmospolicy2026,dreamtrue2026,pavxplorerl2026}, or address lost objects and passive action conditioning with new architectures, object-consistency losses or preference data~\cite{dreamxphi2026,abotphysworld2026}. These efforts build or retrain models; we diagnose and repair consequence identity in the frozen checkpoints already in use.

\paragraph{Evaluating action fidelity.}
Perceptual and physics benchmarks score single rollouts~\cite{vbench2024,worldmodelbench2025,physicsiq2025}; embodied ones use factual references~\cite{ewmbench2025}, human judgments~\cite{mirabench2026} or functional utility~\cite{worldarena2026,worldinworld2026}. The closest protocols compare several actions from one state: WorldSync's WorldEcho~\cite{worldsync2026} measures end-effector alignment, CoCo~\cite{coco2026} scores responses to zero and inverse actions, Twin Rollouts~\cite{twinrollouts2026} couple noise across counterfactual streams, and group-action world models~\cite{groupactions2026} formalize action transformations as operators. These protocols ask whether the model \emph{responds}; we ask whether it can \emph{tell its actions apart}, that is, whether each action's predicted change matches its own simulator change rather than another's, scored all-or-nothing per pair as in~\cite{eyeswideshut2024}.

\paragraph{Strengthening action conditioning.}
Classifier-free guidance~\cite{cfg2022} scales the whole conditional signal, so it amplifies the action's effect together with every other conditional effect~\cite{autoguidance2024}, and it needs an unconditional path. FreeAction~\cite{freeaction2025} replaces that path with the negated action, and Vid2World~\cite{vid2world2026} also guides on actions. Such guidance amplifies the response to an action but does not teach the model which consequence belongs to which action, and it leaves identity unchanged (Sec.~\ref{sec:exp-baselines}). Fine-tuning approaches align video models with rewards, preferences or teacher features~\cite{vader2024,videorepa2025}, none of which compares two actions from one state. Conditioning only part of the noise range has served generation quality and personalization~\cite{guidanceinterval2024,ediffi2022,tsm2025,tclora2025}. We use the denoising trajectory's division of labor for repair: early, high-noise calls decide what happens and late calls decide how it looks, so a frozen action adapter runs only while the outcome forms and stops before appearance is rendered, restoring identity within a preset FVD bound.

\paragraph{Intervention-effect supervision and on-policy training.}
Two lines of work shape the training signal. The first supervises intervention effects: WorldSync~\cite{worldsync2026} aligns the predicted change between two actions with the ground-truth change, and the concurrent CureWM~\cite{curewm2026} repairs a released checkpoint with counterfactual data; CoCo~\cite{coco2026}, CtrlWAM~\cite{ctrlwam2026} and the concurrent R$^2$-WAM~\cite{r2wam2026} and RAP-WAM~\cite{rapwam2026} add counterfactual-consistency losses, simulated action perturbations or repair post-training. The second trains on the model's own predictions to close the exposure gap in sequence models~\cite{scheduledsampling2015,dagger2011}, diffusion~\cite{ning2023input} and autoregressive video~\cite{selfforcing2025,persistworld2026}. In the first line, effects are supervised at noised ground-truth states~\cite{worldsync2026,ctrlwam2026}; the second targets the fidelity of a single sequence, not the difference between actions. \ours{} brings the two lines together: it supervises between-action differences at the sampler's own states, and a step gate holds FVD within a preset bound. We compare with both lines in Sec.~\ref{sec:exp-ablation} and Supp.~S9.

\section{Problem and Diagnosis}
\label{sec:diagnosis}

\subsection{Counterfactual branch contexts}
\label{sec:diag-setup}
We measure identity on \emph{contexts}: a simulator state with five actions branching off. The state comes from a LIBERO-10~\cite{libero2023} human demonstration at an event step, such as first contact. The actions are the demonstrated (factual) one, \br{hold\_pose}, \br{gripper\_opposite}, \br{translation\_axis\_flip} and \br{translation\_half}. The simulator renders every branch's future, so each prediction can be checked against its own ground truth and the others'. Repairs are trained on one set of demonstrations, a development set guides all choices, and two held-out sets, \emph{fresh} and \emph{sealed}, were locked until confirmation; the sets share no contexts, frames or branch-point states (Supp.~S1). ``Held-out'' means unused in training and selection.

\subsection{Metrics}
\label{sec:diag-metrics}
Let $\hat\Delta_b$ be the difference between the last two frames of the predicted video of branch $b$ and those of the factual branch $f$ under shared noise, and $\Delta_b$ the same difference in the simulator-rendered video. We compare the non-factual branches by cosine similarity, $S_{tc}=\cos(\hat\Delta_t,\Delta_c)$. \textbf{\pairacc{}} is the fraction of branches $t$ whose own target is the strictly best match, $S_{tt}>\max_{c\neq t}S_{tc}$, and the \textbf{identity margin} is the mean of $S_{tt}-\max_{c\neq t}S_{tc}$. \textbf{Object-endpoint} and \textbf{contact-event MAE} measure consequence fidelity inside the object mask; \textbf{global MAE} and a \textbf{denoise-MSE ratio} guard against damage elsewhere. Our \textbf{acceptance test}, fixed before the first repair was trained (Supp.~S2), requires reliable gains on these measures, bounded damage elsewhere, and agreement across training seeds.

\subsection{Responsive but not consequence-correct}
\label{sec:diag-spectrum}
We take Prophet~\cite{prophrl2025}, a Cosmos-based~\cite{cosmospredict25_2025} simulator for VLA reinforcement learning, as our main case; UWM~\cite{uwm2025} and A2World~\cite{a2world2026} serve for the diagnosis and for transfer. Table~\ref{tab:diag} compares the three native models on the development contexts. Besides identity, it reports the \textbf{branch/seed SNR}, the effect \textbf{amplitude}, its \textbf{direction cosine} $\cos(\hat\Delta_b,\Delta_b)$ with the ground truth, and the contact MAE relative to a \textbf{zero-response} predictor (Supp.~S2). \emph{Prophet's} object-region effects clear seed noise with about the right amplitude, but their cosine with the ground-truth effect is only \dgPCos{} (A2World \dgACos), so its contact error exceeds that of the zero-response predictor. \emph{UWM's} effects are below seed noise. \emph{A2World} is responsive and mostly correctly oriented. The distinguishing axis is direction, not amplitude. Repeating the diagnosis on the held-out contexts reproduces each of these readings, with native margins of \freshMarginNative{} (fresh) and \sealedMarginNative{} (sealed), and more sampling steps leave Prophet's native margin unchanged (Supp.~S3).

\subsection{Design constraints for a repair}
\label{sec:diag-constraints}
First, in Prophet, actions enter only through two action-embedder MLPs that feed the timestep embedding and the shared adaLN modulation~\cite{dit2023}; a residual adapter placed in the middle of the network learned an action-independent output even when it received the action through its own encoder (Supp.~S3). Second, on states made by noising ground-truth latents, the native intervention-effect error $\|\hat\Delta - \Delta\|^2/\|\Delta\|^2$ (1 means predicting no branch difference) falls from $1.15$ at $\sigma{=}80$ to $0.012$ at $\sigma{=}0.085$ (Supp.~S3): these inputs already contain the branch outcome, so a loss on them says little about the action, whereas on the model's own sampler states the error stays near 1 at every step. Useful supervision must therefore come from these states.

\section{OPIE with Step-Scheduled Deployment}
\label{sec:method}

\ours{} follows from the constraints of Sec.~\ref{sec:diag-constraints}: train a small action-pathway adapter at the sampler's own states, supervise branch \emph{differences}, and gate the adapter by noise level at deployment (Fig.~\ref{fig:method}b,c; training loop in Supp.~S2).

\textbf{Setup.} A context provides a state $s$ and branches $b\in\mathcal{B}=\{f,1,\dots,4\}$ with actions $a_b$, where $f$ is the factual branch. The WAM denoises a video latent with $D_\theta(x,\sigma,s,a)$ along a fixed noise schedule $\sigma_0{>}\dots{>}\sigma_5$ parameterized as in~\cite{edm2022}. We write $x_b$ for the simulator latent of branch $b$ and $\Delta_b=x_b-x_f$ for its ground-truth effect, the latent-space counterpart of the pixel difference in Sec.~\ref{sec:diag-metrics}.

\textbf{Where to intervene.}
Because actions enter Prophet only through two action-embedder MLPs (Sec.~\ref{sec:diag-constraints}), we repair exactly that pathway: we add zero-initialized rank-$r$ updates~\cite{lora2022} to their four linear layers (Eq.~\ref{eq:lora}),
\begin{equation}
  W = W_0 + BA,\qquad B\in\mathbb{R}^{d_\mathrm{out}\times r},\; A\in\mathbb{R}^{r\times d_\mathrm{in}},
  \label{eq:lora}
\end{equation}
and freeze all other weights. With $r{=}30$ this is $1.25$M parameters, $0.06\%$ of the $2.06$B-parameter DiT~\cite{dit2023}; at initialization the model is bitwise identical to native.

\begin{table*}[t]
\centering
\caption{\textbf{Within the FVD bound, ours gives the largest identity gain and the largest drop in consequence error.} Top: registered held-out test (236 contexts). Bottom: baselines on the fresh split (117 contexts). Gains over native; native in parentheses as absolute values; FVD as ratio to native, not computed for the guidance baselines. Bold: best identity and error per block within the FVD bound ($\le1.10$).}
\label{tab:results}
\footnotesize
\setlength{\tabcolsep}{7pt}
\renewcommand{\arraystretch}{1.02}
\begin{tabular}{@{}lcccccc@{}}
\toprule
 & & \multicolumn{2}{c}{Identity} & \multicolumn{2}{c}{Consequence error ($10^{-3}$)} & Realism \\
\cmidrule(lr){3-4}\cmidrule(lr){5-6}\cmidrule(l){7-7}
Method & Trained params & Margin$\uparrow$ & PairAcc$\uparrow$ & Object$\downarrow$ & Contact$\downarrow$ & FVD ratio$\downarrow$ \\
\midrule
\multicolumn{7}{@{}l}{\textit{Registered held-out test}} \\
Native Prophet & -- & ($\pNullNatMarg{}$) & ($\pNullNatPair{}$) & ($\tObjNatAbs{}$) & ($\tContNatAbs{}$) & 1.00 \\
Always-on repair & 1.25M & $\scXFullMargin{}$ & $\tPairFull{}$ & $\tObjFull{}$ & $\tContFull{}$ & $\scXFullFVD{}$ \\
Ours (scheduled) & 1.25M & $\mathbf{\scXMargin{}}$ & $\mathbf{\scXPair{}}$ & $\mathbf{\tObjSched{}}$ & $\mathbf{\tContSched{}}$ & $\scXFVD{}$ \\
\midrule
\multicolumn{7}{@{}l}{\textit{Baselines on the fresh split}} \\
Native Prophet & -- & ($\mtNatMarg{}$) & ($\mtNatPair{}$) & ($\mtNatObj{}$) & ($\mtNatCont{}$) & 1.00 \\
Action gain $\times1.5$ & 0 & $\mtMargGain{}$ & $\mtPairGain{}$ & $\mtObjGain{}$ & $\mtContGain{}$ & $\hbGainFvd{}$ \\
Contrastive guidance & 0 & $\mtMargContr{}$ & $\mtPairContr{}$ & $\mtObjContr{}$ & $\mtContContr{}$ & -- \\
FreeAction~\cite{freeaction2025} & 0 & $\mtMargFa{}$ & $\mtPairFa{}$ & $\mtObjFa{}$ & $\mtContFa{}$ & -- \\
WorldSync-style~\cite{worldsync2026} & 71.7M & $\mtMargWs{}$ & $\mtPairWs{}$ & $\mtObjWs{}$ & $\mtContWs{}$ & $\hbWsFvd{}$ \\
Full-rank, GT fit & 71.7M & $\mtMargFrg{}$ & $\mtPairFrg{}$ & $\mtObjFrg{}$ & $\mtContFrg{}$ & $\hbFrgFvd{}$ \\
Full-rank, our loss & 71.7M & $\mtMargFro{}$ & $\mtPairFro{}$ & $\mtObjFro{}$ & $\mtContFro{}$ & $\hbFroFvd{}$ \\
CureWM-style~\cite{curewm2026} & 22.9M & $\mtMargCw{}$ & $\mtPairCw{}$ & $\mtObjCw{}$ & $\mtContCw{}$ & $\hbCwFvd{}$ \\
Backbone adapter, GT fit & 22.9M & $\mtMargBOne{}$ & $\mtPairBOne{}$ & $\mtObjBOne{}$ & $\mtContBOne{}$ & $\hbBOneFvd{}$ \\
Backbone adapter, our loss & 22.9M & $\mtMargBTwo{}$ & $\mtPairBTwo{}$ & $\mtObjBTwo{}$ & $\mtContBTwo{}$ & $\hbBTwoFvd{}$ \\
Ours (scheduled) & 1.25M & $\mathbf{\mtMargOpie{}}$ & $\mathbf{\mtPairOpie{}}$ & $\mathbf{\mtObjOpie{}}$ & $\mathbf{\mtContOpie{}}$ & $\hbOpieFvd{}$ \\
\bottomrule
\end{tabular}
\end{table*}

\begin{figure*}[t]
\centering
\includegraphics[width=\textwidth]{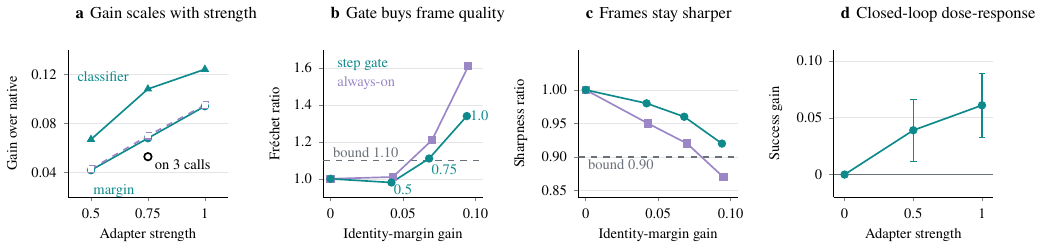}
\caption{\textbf{The step gate is not a lower dose.} (a) Margin and classifier gain by adapter strength. (b, c) Fr\'echet ratio and sharpness ratio against margin gain. (d) Closed-loop success gain by strength.}
\label{fig:gate}
\end{figure*}

\textbf{On-policy branch states.}
Noised ground-truth latents already contain the branch outcome at low noise (the exposure mismatch, Sec.~\ref{sec:diag-constraints}), so we train where the sampler actually goes. For one context we draw a single noise $\epsilon$, share it across all five branches, and run the model's own sampler to step $k$, giving states $x^{(k)}_b$ that differ only in the action, so every branch difference the adapter sees is caused by the action. The adapter is trained on the clean-latent predictions at these states (Eq.~\ref{eq:onpolicy}),
\begin{equation}
  \hat x_b = D_\theta\bigl(x^{(k)}_b,\sigma_k,s,a_b\bigr),\qquad \hat\Delta_b=\hat x_b-\hat x_f .
  \label{eq:onpolicy}
\end{equation}
Each update uses one context and one noise level, cycled so that every context visits every level; gradients flow only through $BA$, and our sampler reproduces the inference endpoints exactly (schedule: Supp.~S2).

\textbf{Objective.} Because Prophet's effects have about the right size but a different orientation (Sec.~\ref{sec:diag-spectrum}), the identity term, Eq.~\ref{eq:dir}, is scale-free:
\begin{equation}
  \mathcal{L}_{\mathrm{dir}}=\frac{1}{4}\sum_{b\neq f}\Bigl(1-\cos\bigl(\hat\Delta_b,\Delta_b\bigr)\Bigr),
  \label{eq:dir}
\end{equation}
with the cosine over all generated latent frames. Two object terms, Eqs.~\ref{eq:obj} and~\ref{eq:cons}, tie the corrected effect to the manipulated object. With $M$ the union over branches of the target-object mask at the last frame $[\cdot]_T$ and $\|\cdot\|^2_M$ the mean squared error inside $M$,
\begin{align}
  \mathcal{L}_{\mathrm{obj}}&=\frac{1}{5}\sum_{b}\bigl\|[\hat x_b]_T-[x_b]_T\bigr\|^2_M, \label{eq:obj}\\
  \mathcal{L}_{\mathrm{cons}}&=\frac{1}{4}\sum_{b\neq f}\bigl\|[\hat\Delta_b]_T-[\Delta_b]_T\bigr\|^2_M . \label{eq:cons}
\end{align}
Two regularizers (Eq.~\ref{eq:reg}) keep the repaired prediction close to the model and to the data: an anchor to the native prediction $\hat x_b^{0}$ at the same state, computed with the adapter off, and a denoising term to the ground truth,
\begin{equation}
  \mathcal{L}_{\mathrm{anc}}=\frac{1}{5}\sum_{b}\bigl\|\hat x_b-\hat x_b^{0}\bigr\|^2,\qquad
  \mathcal{L}_{\mathrm{den}}=\frac{1}{5}\sum_{b}\bigl\|\hat x_b-x_b\bigr\|^2,
  \label{eq:reg}
\end{equation}
both averaged over the generated latent frames. The full objective (Eq.~\ref{eq:loss}) is
\begin{equation}
  \mathcal{L}=\mathcal{L}_{\mathrm{dir}}+\mathcal{L}_{\mathrm{obj}}+\mathcal{L}_{\mathrm{cons}}+0.3\,\mathcal{L}_{\mathrm{anc}}+0.1\,\mathcal{L}_{\mathrm{den}} .
  \label{eq:loss}
\end{equation}
The direction term answers the misaligned directions of Sec.~\ref{sec:diag-spectrum}, the object terms the object and contact errors, and the two regularizers guard the rest of the video; the acceptance test checks each of them. Term weights were fixed before confirmation (Supp.~S2) and are ablated in Sec.~\ref{sec:exp-ablation}.

\textbf{Training and confirmation.}
The repair takes 1024 updates (learning rate $3\times10^{-4}$, 781 training contexts), about $1.8$ GPU-hours on one A100-40GB. We froze one candidate and a decision rule before generating confirmation data.

\textbf{Step-scheduled deployment.} Kept active at every call, the adapter raises FVD (Sec.~\ref{sec:exp-main}). We therefore follow the division of labor of the denoising trajectory: the branch outcome forms in the early, high-noise calls, and the late calls mainly render appearance. The frozen update runs only in the early calls (Eq.~\ref{eq:gate}),
\begin{equation}
  W(\sigma)=W_0+\mathbb{1}[\sigma\ge\tau]\,BA .
  \label{eq:gate}
\end{equation}
Prophet uses $\tau=0.5$ (the first four of six denoiser calls) and A2World $\tau=5$, each selected on the development set. The late calls run the native weights unchanged, and the gate adds no denoiser call, parameter or training step.

\section{Experiments}
\label{sec:exp}
\textbf{Setup.} We test identity within the FVD bound, closed-loop selection, ablations and baselines, transfer and a real robot (Secs.~\ref{sec:exp-main}--\ref{sec:exp-real}). Offline tests use held-out contexts, unseen states of the same ten tasks: 117 fresh and 119 sealed (Sec.~\ref{sec:diag-setup}), opened after the candidate was frozen. All selections used the development set. A readout is \emph{registered} when its endpoints, contrasts and decision rule were hashed before its outcome was read; other levels are in Supp.~S1.

\textbf{Arms and statistics.} Native is the unmodified model, always-on the adapter at every call, and ours the same weights under the step gate. In closed loop a fixed UWM policy proposes four action chunks, the WAM imagines each, a frozen value head scores the final frames and the best is executed; the policy alone uses no WAM and the oracle executes all four. Each task uses its 50 official initial states. Gains are paired within one evaluation process; intervals are bootstrapped, clustered by context offline and by task and initial state in closed loop, and every registered offline primary test passes Holm's procedure~\cite{holm1979} (Supp.~S1).

\subsection{Identity within the FVD bound}
\begin{figure*}[t]
\centering
\includegraphics[width=0.83\textwidth]{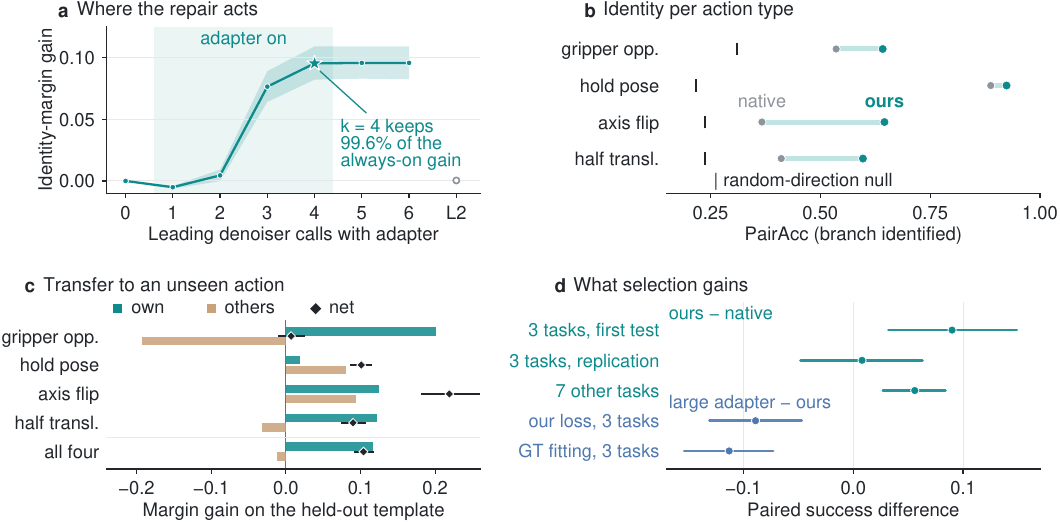}
\caption{\textbf{The gain forms in the high-noise calls and covers every action type.} (a) Margin gain by adapter-on range. (b) PairAcc per action. (c) Gain on an unseen template. (d) Closed-loop success differences.}
\label{fig:analysis}
\end{figure*}

\label{sec:exp-main}
Does switching the adapter off late keep the identity gain within the FVD bound? Table~\ref{tab:results} summarizes the registered held-out test over four training and two evaluation seeds. Always-on repair raises the identity margin but pays a clear realism cost; switching the adapter off for the last two calls keeps the gain and brings FVD back inside the preset bound, so all three registered endpoints pass (Sec.~\ref{sec:diag-metrics}). Object and contact errors also fall, and the factual-branch flow cosine, unused by our loss, rises from $\eFlowDirNat{}$ to $\eFlowDirSched{}$.

\textbf{Frame quality.} Per frame (Supp.~S5), ours is closer to the simulator than native in LPIPS~\cite{lpips2018} and passes the sharpness bound. The gate is not a lower dose. In Fig.~\ref{fig:gate}a--c the margin gain follows the adapter strength, with or without the gate. At the same margin gain, however, the gate gives a lower Fr\'echet ratio~\cite{steinfd2023} and higher sharpness, so strength sets the operating point (Supp.~S5). The gain is positive for every seed and both held-out sets (Supp.~S4).

\textbf{Identity read without our loss.} A registered readout uses neither generated frames nor our loss: a five-way classifier on frozen DINOv2 features~\cite{dinov22024}, trained only on simulator frames of the training pool, names the action that produced a final frame. It assigns clearly more of our endpoints to their own action than of native's; the gain grows with adapter strength and tracks the margin gain across all variants. In a control adapter, each branch is supervised by another branch's future. This adapter gains neither margin nor accuracy, so the gain comes from the identity of the branch rather than from branch-wise supervision (Supp.~S6).

\textbf{Object probe and human study.} A registered object-displacement probe matches the classifier. In a registered blind two-choice study on 80 held-out items, raters judged ours closer to the simulator on $\hxK{}$ of $\hxN{}$ items with a clear majority, and every valid rater favored ours (Supp.~S6).

\begin{figure*}[t]
\centering
\includegraphics[width=0.80\textwidth]{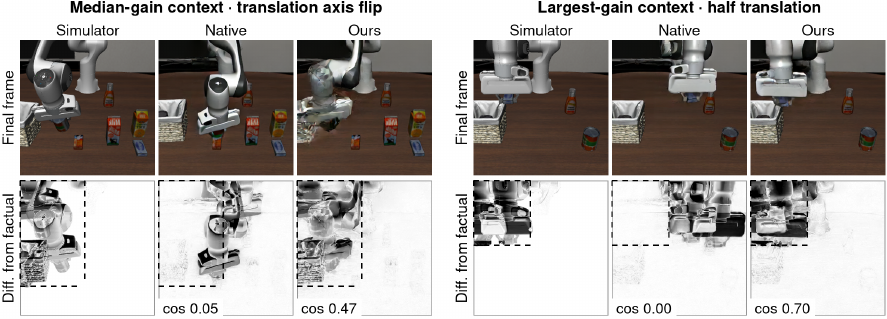}
\caption{\textbf{Same state, a different action.} Median- and largest-gain held-out contexts. Top: final frames (simulator, native, ours); bottom: their difference from the factual branch with its cosine to the simulator's; dashed box: where the simulator moved.}
\label{fig:qual}
\end{figure*}

\textbf{Qualitative view.} Fig.~\ref{fig:qual} shows one counterfactual action in the held-out contexts with the median and the largest gain; the difference from the factual branch isolates what the action changed. Native's difference is nearly unrelated to the simulator's, whereas ours moves the arm to where the simulator places it and its difference follows the simulator's. All five actions are in Supp.~S10.

\textbf{Where the repair acts.} With the adapter on for only the first $k$ of six calls, the gain appears within the first few high-noise calls (Fig.~\ref{fig:analysis}a), and all four action types gain PairAcc (Fig.~\ref{fig:analysis}b), so the repair is not tied to one kind of intervention.

\subsection{Closed-loop selection}
\begin{table}[t]
\centering
\caption{\textbf{The repair raises closed-loop selection success.} Best-of-four success of each arm in each test, and the gain of ours over native with and without task~1. Bold: best non-oracle arm. $^\dagger$Earlier batches, outside the four registered tests.}
\label{tab:closed}
\footnotesize
\setlength{\tabcolsep}{2.6pt}
\begin{tabular}{@{}lcccccc@{}}
\toprule
Test & Policy & Native & Ours & Oracle & Gain & No task 1 \\
\midrule
Confirmation & $\cfRateBase{}$ & $\cfRateNat{}$ & $\mathbf{\cfRateSched{}}$ & $\cfRateOracle{}$ & $\cfNat{}$ & $\cfNoOne{}$ \\
Higher headroom & $\fuRateBase{}$ & $\fuRateNat{}$ & $\mathbf{\fuRateSched{}}$ & $\fuRateOracle{}$ & $\fuNat{}$ & $\fuNoOne{}$ \\
Four adapters I & $\fqRateBase{}$ & $\fqRateNat{}$ & $\mathbf{\fqRateSched{}}$ & $\fqRateOracle{}$ & $\fqNat{}$ & $\fqNoOne{}$ \\
Four adapters II & $\rzRateBase{}$ & $\rzRateNat{}$ & $\mathbf{\rzRateSched{}}$ & $\rzRateOracle{}$ & $\rzNat{}$ & $\rzTwoDiff{}$ \\
Earlier, seeds 4--7$^\dagger$ & $\ceRateBase{}$ & $\ceRateNat{}$ & $\mathbf{\ceRateSched{}}$ & $\ceRateOracleR{}$ & $\ceNatR{}$ & -- \\
Three tasks$^\dagger$ & $\eFourBase{}$ & $\eFourNative{}$ & $\mathbf{\eFourRecipeRate{}}$ & $\eFourOracle{}$ & $\eFourRecNat{}$ & -- \\
\bottomrule
\end{tabular}
\end{table}

\label{sec:exp-down}
Does identity change which action a planner picks? Table~\ref{tab:closed} lists every arm in each test, and Fig.~\ref{fig:teaser}c shows the gains. Each test uses new policy seeds and re-runs all arms (Supp.~S7).

\textbf{Ten-task confirmation.} We registered a superiority test of ours against native over all ten tasks. Ours passes it and exceeds the policy alone, while native stays near the latter, so imagined rollouts help only after the repair; the gain stays positive without task~1 and in both seeds (Supp.~S7).

\textbf{Follow-up tests.} A second registered test used the setting with most oracle headroom in a WAM-free pilot; ours again gains over native and over the policy alone, also without task~1 and for each training seed. Two further registered tests ran all four independently trained adapters at one setting: every adapter gains in both. The later test repeats the earlier one on five new policy seeds and meets all registered criteria, also without task~1. Pooled over the four batches, the gain relies on no single batch nor task~1 (Supp.~S7).

\textbf{Dose-response.} In a registered test the four adapters ran at strength $0$ (native), $0.5$ and $1.0$ on four new policy seeds, rescaled rather than retrained. The gain over native grows with the strength and the registered dose route is supported, also without task~1 (Fig.~\ref{fig:gate}d, Supp.~S7). The gain is larger on initial states never used in training (Supp.~S7).

\begin{figure}[t]
\centering
\includegraphics{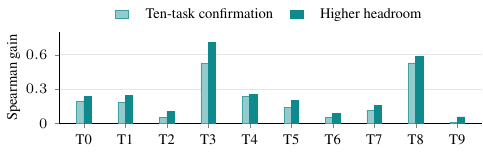}
\caption{\textbf{Ours ranks the candidates better in every task.} Spearman gain over native per task (first query).}
\label{fig:rank}
\end{figure}

\textbf{Ranking of the candidates.} At each episode's first query all arms share the four candidates, and the oracle runs each in the simulator. Native's imagined scores rank them close to chance; ours ranks them clearly better, positively in all ten tasks (Fig.~\ref{fig:rank}), and picks the simulator's best candidate more often; in the readout pre-specified for both four-adapter tests, the rank correlation rises from $\fqRhoNat{}$ to $\fqRhoSch{}$ and from $\rzRhoNat{}$ to $\rzRhoSch{}$ (Supp.~S7). An adapter trained with shuffled branch targets does not reproduce the ranking gain (Supp.~S6), and the repaired Prophet recovers most of the selection gap to native A2World, which is built to link each action to its consequence (Supp.~S7). With up to sixteen candidates, ours also shrinks native's regret for every seed (Supp.~S7).

\subsection{Ablations and baselines}
\begin{figure*}[t]
\centering
\includegraphics{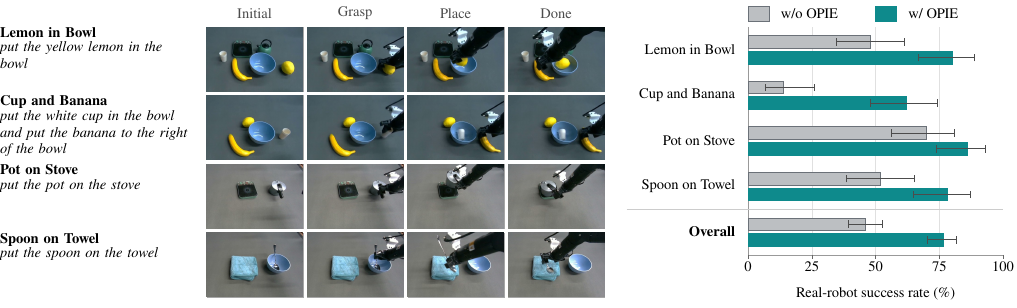}
\caption{\textbf{Real-robot validation.} Left: example successful rollouts. Right: success of policies trained in the original (\rwArmA{}) or the \ours{}-repaired (\rwArmB{}) world model, \rwNEval{} real trials per scene and arm, with 95\% Wilson intervals.}
\label{fig:real}
\end{figure*}

\label{sec:exp-ablation}
\label{sec:exp-baselines}
\begin{table}[t]
\centering
\caption{\textbf{Each recipe choice carries part of the gain.} Ablations on 236 held-out contexts, gains over native; errors in $10^{-3}$.}
\label{tab:ablations}
\footnotesize
\setlength{\tabcolsep}{4pt}
\begin{tabular*}{\linewidth}{@{}l@{\extracolsep{\fill}}cccc@{}}
\toprule
Variant & Margin$\uparrow$ & PairAcc$\uparrow$ & Object$\downarrow$ & Contact$\downarrow$ \\
\midrule
Ours (scheduled) & $\mathbf{\tRecMarg{}}$ & $\mathbf{\tRecPair{}}$ & $\mathbf{\mtAblRecObj{}}$ & $\mathbf{\mtAblRecCont{}}$ \\
GT-noised states & $\tGtnMarg{}$ & $\tGtnPair{}$ & $\mtAblGtnObj{}$ & $\mtAblGtnCont{}$ \\
w/o $\mathcal{L}_{\mathrm{dir}}$ & $\tNodirMarg{}$ & $\tNodirPair{}$ & $\mtAblNodirObj{}$ & $\mtAblNodirCont{}$ \\
GT fitting only & $\tGtfitMarg{}$ & $\tGtfitPair{}$ & $\mtAblGtfitObj{}$ & $\mtAblGtfitCont{}$ \\
\bottomrule
\end{tabular*}
\vspace{8pt}
\caption{\textbf{Transfer to other actions, models and suites.} Gains over native in margin and PairAcc, and the FVD ratio.}
\label{tab:transfer}
\scriptsize
\setlength{\tabcolsep}{3.5pt}
\renewcommand{\arraystretch}{0.94}
\begin{tabular*}{\linewidth}{@{}l@{\extracolsep{\fill}}rccc@{}}
\toprule
Setting & $n$ & Margin$\uparrow$ & PairAcc$\uparrow$ & FVD$\downarrow$ \\
\midrule
\multicolumn{5}{@{}l}{\textit{Other actions}} \\
\quad Unseen template & 236 & $\htXGain{}$ & -- & -- \\
\quad Policy candidates & $\pcN{}$ & $\pcMarg{}$ & $\pcPair{}$ & $\pcFvd{}$ \\
\addlinespace[1pt]
\multicolumn{5}{@{}l}{\textit{Other models}} \\
\quad Cosmos Policy & $\ctN{}$ & $\ctMargin{}$ & $\ctPair{}$ & -- \\
\quad A2World & 236 & $\awXMargin{}$ & $\awXPair{}$ & $\awXFVD{}$ \\
\addlinespace[1pt]
\multicolumn{5}{@{}l}{\textit{Other suites}} \\
\quad LIBERO-Spatial & $\suSpatialN{}$ & $\suSpatialMargin{}$ & $\suSpatialPair{}$ & $\suSpatialFvd{}$ \\
\quad LIBERO-Object & $\suObjectN{}$ & $\suObjectMargin{}$ & $\suObjectPair{}$ & $\suObjectFvd{}$ \\
\quad LIBERO-Goal & $\suGoalN{}$ & $\suGoalMargin{}$ & $\suGoalPair{}$ & $\suGoalFvd{}$ \\
\quad LIBERO-90 subset & $\suNinetyN{}$ & $\suNinetyMargin{}$ & $\suNinetyPair{}$ & $\suNinetyFvd{}$ \\
\bottomrule
\end{tabular*}
\end{table}

Which ingredients carry the gain (Table~\ref{tab:ablations})? Training on noised ground truth instead of the sampler's own states, or dropping the direction term, removes much of the gain, and GT fitting alone barely moves it; each is below ours at the task level, so all three choices matter. Without $\mathcal{L}_{\mathrm{dir}}$ the error drops remain, so this term carries identity; rank~8 and smaller training sets stay close (Supp.~S9).

\textbf{Baselines.} The bottom of Table~\ref{tab:results} scores every baseline on the fresh split. Ours gains more margin than eight of the nine baselines, with every contrast above zero (Supp.~S9), and within the FVD bound lowers object and contact error the most. The exception, the 22.9M-parameter backbone adapter trained with our loss, reaches a higher margin gain but at a higher FVD ratio and with far more parameters. In closed loop both large adapters select worse than ours and neither beats native (Fig.~\ref{fig:analysis}d, Supp.~S9).

\subsection{Transfer}
\label{sec:exp-transfer}
Does the gain reach other tasks, models and actions (Table~\ref{tab:transfer})?

\textbf{Other tasks.} On four LIBERO suites never used in training, the registered margin and PairAcc rise in every suite and training seed, and the gate again cuts the realism cost of always-on repair on new scenes (Supp.~S8).

\textbf{Other models.} Cosmos Policy~\cite{cosmospolicy2026} shares Prophet's low native direction cosine, and a rank-5 attention adapter with the same loss clearly raises its held-out margin and PairAcc. UWM, repaired with its own adapter, gains in latent-space margin, whereas A2World, whose native identity is already high, gains little, as expected (Supp.~S3, S8). A Wan2.2-based WAM~\cite{wan2025,wovr2026} already has high native identity, so the diagnosis tells which models need repair (Supp.~S8).

\textbf{Other actions.} On policy-proposed actions the gated adapter raises the margin with both realism guards held, and holding out one intervention template at a time gives a positive pooled gain over the four folds (Fig.~\ref{fig:analysis}c, Supp.~S8).

\subsection{Real-robot validation}
\label{sec:exp-real}
We train policies in the original and in the \ours{}-repaired world model and evaluate each with \rwNEval{} real trials in each of four scenes (Fig.~\ref{fig:real}). Both policies run on the same pre-drawn object placements, one after the other in random order, and the operator does not know which policy is running (Supp.~S12). The repaired world model gives higher success in every scene; pooled over the scenes, success rises from \rwPct{\rwSumA}{\rwNAll}\% to \rwPct{\rwSumB}{\rwNAll}\%, and the difference is significant in a test stratified by scene. The gain is largest in the two-step \rwSceneB{} scene, \rwPct{\rwScBA}{\rwNEval}\% $\to$ \rwPct{\rwScBB}{\rwNEval}\%, where a trial counts as a success only if both steps are completed, and smallest in \rwSceneC{}, \rwPct{\rwScCA}{\rwNEval}\% $\to$ \rwPct{\rwScCB}{\rwNEval}\%, where the original model already succeeds most often.

\section{Outlook}
\label{sec:limitations}
\textbf{Identity, not margin alone.} The large backbone adapter reaches a higher margin at a higher FVD ratio, yet selects less well than ours (Sec.~\ref{sec:exp-ablation}). Selection thus follows identity held within the FVD bound, so WAM training should target the two together; we will test this with a second selection policy and value head.

\textbf{Real counterfactuals.} Learned from simulator counterfactuals, the repair already yields better policies on a real robot. Real counterfactual video, recorded from one state under several actions, would let it learn on the robot itself.

\textbf{Which models need repair?} The diagnosis tells before any training (Sec.~\ref{sec:exp-transfer}), and we plan to use it as a checkpoint test while WAMs are trained.

\section{Conclusion}
\label{sec:conclusion}
A WAM is judged by how real its rollouts look and whether it responds to an action, yet it can pass both and still show the consequence of a different action. Current benchmarks do not test this consequence identity, guidance amplifies the response without telling actions apart, and supervision at noised ground-truth states recovers only about half of the gain. Our five-action diagnostic measures identity directly. \ours{} repairs it at the sampler's own states with a $0.06\%$-parameter adapter, and step-scheduled deployment keeps FVD within a preset bound, where ours gives the largest identity gain of all baselines. The repaired WAM picks better actions in four registered closed-loop tests and trains better policies on a real robot in all four scenes.

\clearpage
\section*{LLM Usage Statement}
{\small
We used Claude through Claude Code to edit and restructure the text of this manuscript and its supplement, to format the LaTeX source, to assist with literature search, and to check that the experimental numbers are consistent across the text, tables and supplement; every retrieved reference was checked against its arXiv or publisher record. The authors designed the method, ran every experiment, derived the results and conclusions, reviewed all text, and take full responsibility for every claim and number in this paper and its supplement.
}

{\small
\phantomsection
\refstepcounter{section}
\label{sec:refs-start}
\bibliographystyle{ieeenat_fullname}
\bibliography{refs}
}

\clearpage
\onecolumn
\renewcommand{\thesection}{S\arabic{section}}
\renewcommand{\thetable}{S\arabic{table}}
\renewcommand{\thefigure}{S\arabic{figure}}
\renewcommand{\thealgorithm}{S\arabic{algorithm}}
\setcounter{section}{0}\setcounter{table}{0}\setcounter{figure}{0}\setcounter{algorithm}{0}
\renewcommand{\awFreshDecision}{confirmed, non-inferior}
\renewcommand{\awFreshCtrlDecision}{confirmed, non-inferior}
\renewcommand{\awSealedDecision}{confirmed, non-inferior}
\renewcommand{\awSealedCtrlDecision}{confirmed, non-inferior}
\providecommand{\ptci}[2]{$#1$\,{\scriptsize$#2$}}
\setlength{\emergencystretch}{3em}
\begin{center}{\large\bfseries A Case of Mistaken Identity: Repairing Consequence Identity in World Action Models\\[2pt] Supplementary Material}\end{center}\vspace{-4pt}
\noindent This supplement follows the order of the main paper; numbers prefixed with S refer to it, all others to the main paper, whose references it uses.

\noindent\textbf{Evidence.} A readout is \emph{registered} when its endpoints, contrasts and decision rule were fixed and hashed before its outcome was read, \emph{sequential} when that registration came after earlier outcomes were known, \emph{pre-specified} when it was named in such a registration as a secondary readout without a decision rule, and \emph{descriptive} otherwise. Offline numbers are on held-out contexts (fresh and sealed) unless a table names the development set; closed-loop numbers are best-of-four selection success on the ten LIBERO-10 tasks.

\section{Data, splits and statistics}
\label{supp:n1}

This section supports the data description of Sec.~\ref{sec:diag-setup}, the evidence levels and statistics of Sec.~\ref{sec:exp}, and the disjointness claim.

\paragraph{Splits.}
A context is made by replaying demonstration $k$ of a task's official LIBERO-10 file to an event step and simulating five 20-step action branches from that state.
All sets use the same ten tasks and differ in the demonstration index (Table~\ref{tab:s1-splits}).
The held-out data come from the same pipeline and filter as the training pool, without selection by model output, and were locked until the candidate and decision rule were frozen.

\begin{table}[!htb]
  \centering\scriptsize\setlength\tabcolsep{3pt}
  \caption{Data splits. Every context has five branches. Sets are defined by demonstration index.}
  \label{tab:s1-splits}
  \begin{tabular}{@{}l p{2.3cm} p{4.4cm} p{6.0cm}@{}}
    \toprule
    Split & Demonstrations & Contexts & Role \\
    \midrule
    Training pool & 20--34, 43--47 & 781 (3905 branch entries) & Repair training \\
    Pool validation & subset of 43--47 & 40 (8 per demonstration) & Baseline learning rates \\
    Development & 35 & 39 (195 branch entries) & All selections \\
    Locked & 36 & 39 & Unused \\
    Sealed & 37--39 & 119 (39, 40, 40) & Held out \\
    Fresh & 40--42 & 117 (38, 40, 39) & Held out, read first \\
    Other suites & Spatial, Object, Goal, LIBERO-90 subsample & 105, 101, 110, 107 & Sec.~\ref{sec:exp-transfer} \\
    \bottomrule
  \end{tabular}
\end{table}

\paragraph{Disjointness.}
The demonstration indices of the training pool and of 35--42 do not intersect, and neither do their (task, demonstration) pairs or context identifiers.
Between the pool and demonstration 35 we also compared initial states, action sequences, branch-point states and seven file hashes per context: none is shared, except the \br{hold\_pose} action arrays, which are identical by construction.
For demonstrations 37--42, disjointness rests on the disjoint indices and a recorded authorization before generation.
The Prophet and UWM checkpoints were fine-tuned on LIBERO-10 and may have seen all 50 demonstrations.
The four further suites were fixed by a registered rule before rendering: three hash-ordered demonstrations per task for Spatial, Object and Goal, one demonstration for each of thirty LIBERO-90 tasks, at most four events per task and demonstration, and at least 40 contexts per suite.
The adapters were trained on LIBERO-10 contexts only; the UWM checkpoint was pretrained on LIBERO-90, so its results on that suite are flagged as in-suite and not pooled.

\paragraph{Registration index.}
Table~\ref{tab:s1-reg} lists each readout on the held-out sets or the closed loop with its evidence level (Sec.~\ref{sec:exp}); demonstration 36 was read by none.

\begin{table}[!htb]
  \centering\scriptsize\setlength\tabcolsep{3pt}
  \caption{Registered readouts and their evidence level, in the order they were read.}
  \label{tab:s1-reg}
  \begin{tabular}{@{}p{10.5cm} l@{}}
    \toprule
    Readout & Evidence level \\
    \midrule
    Always-on repair on Prophet, fresh then sealed (seven-condition acceptance test) & Registered \\
    Goal selection on both sets, for Prophet and for the UWM adapter & Sequential \\
    UWM adapter, margin as primary endpoint & Sequential \\
    A2World non-inferiority confirmation & Sequential \\
    Frame-level quality and FVD, three models & Sequential \\
    Thirty-five-step readout of the frozen repair & Sequential \\
    Step gate on Prophet and on A2World & Sequential \\
    Held-out intervention templates & Sequential \\
    Backbone adapter against the recipe & Sequential \\
    Four further suites & Sequential \\
    Closed loop: training-seed replication, seven-task extension, new policy seeds, step-scheduled repair, large adapter & Sequential \\
    Closed loop, both four-adapter tests: ranking of the candidates at the first query (Spearman, top-1) & Pre-specified \\
    \bottomrule
  \end{tabular}
\end{table}

\paragraph{Nondeterminism and pairing.}
Native rollouts of Prophet are bitwise equal inside one evaluation process but not across processes.
Between two units on the same context and evaluation seed the largest native margin difference is \freshNativeMaxDiffMargin{} (fresh) and \sealedNativeMaxDiffMargin{} (sealed), and at most 6 of 234 context--seed pairs change PairAcc by one step.
Every reported gain is the mean over units of the adapter-on rows minus the native rows of the same unit, so each pair comes from one process.

\paragraph{Cluster units and multiplicity.}
We recomputed the registered offline endpoints from the stored per-context rows with three cluster units: context (as in the main text), demonstration (60 clusters) and task (10 clusters), each with 2000 percentile-bootstrap draws (Table~\ref{tab:s1-cluster}).
For scheduled minus native the pooled gain is \mcPoint{} with a context interval \mcCtx{}, a demonstration interval \mcDemo{} and a task interval \mcTask{}; every registered primary interval keeps its side of zero under each unit.
For Holm's step-down procedure at $0.05$ we recomputed each of the \hmM{} registered primary tests (the rows of Table~\ref{tab:s1-cluster} other than always-on minus native) by a one-sided percentile bootstrap with $20{,}000$ draws; all are rejected under each unit, and the largest adjusted $p$ is \hmMaxPCtx{} with contexts, \hmMaxPDemo{} with demonstrations and \hmMaxPTask{} with tasks (A2World against native in the context case).
Endpoints outside the family carry no adjustment.

\begin{table}[!htb]
\centering
\caption{Registered identity-margin endpoints under three clustering units, pooled over the $\mcN{}$ held-out contexts; 95\% percentile intervals.}
\label{tab:s1-cluster}
\scriptsize
\setlength{\tabcolsep}{2.5pt}
\begin{tabular}{@{}lcccc@{}}
\toprule
Endpoint & Est. & Context & Demonstration & Task\\
\midrule
Scheduled minus native & $+0.096$ & $[+0.082, +0.109]$ & $[+0.084, +0.109]$ & $[+0.076, +0.117]$\\
Scheduled minus always-on (non-inferiority) & $-0.000$ & $[-0.001, +0.000]$ & $[-0.001, +0.000]$ & $[-0.001, +0.000]$\\
Always-on minus native & $+0.096$ & $[+0.083, +0.110]$ & $[+0.083, +0.109]$ & $[+0.078, +0.117]$\\
A2World, scheduled minus native & $+0.005$ & $[+0.001, +0.009]$ & $[+0.001, +0.009]$ & $[+0.000, +0.010]$\\
A2World, scheduled minus always-on & $-0.000$ & $[-0.003, +0.002]$ & $[-0.003, +0.002]$ & $[-0.003, +0.003]$\\
Held-out templates, always-on & $+0.107$ & $[+0.095, +0.119]$ & $[+0.096, +0.118]$ & $[+0.093, +0.122]$\\
Held-out templates, gated & $+0.104$ & $[+0.092, +0.117]$ & $[+0.094, +0.116]$ & $[+0.091, +0.120]$\\
Backbone adapter minus ours, always-on & $+0.054$ & $[+0.046, +0.063]$ & $[+0.046, +0.063]$ & $[+0.041, +0.069]$\\
Backbone adapter minus ours, gated & $+0.061$ & $[+0.052, +0.069]$ & $[+0.052, +0.070]$ & $[+0.047, +0.077]$\\
\bottomrule
\end{tabular}

\end{table}

\section{Metrics, objective and implementation}
\label{supp:n2}

This section gives the metrics and acceptance test of Sec.~\ref{sec:diag-metrics}, the recipe of Sec.~\ref{sec:method}, the selection of the step gate, compute and the released code.

\paragraph{Metrics.}
Every metric compares the endpoint of a branch (the last two frames of its agent-view video, indices 19 and 20 of the 21-frame clip, values in $[0,1]$) with the simulator endpoint.
The object mask is the union, over the five branches and both endpoint frames, of the simulator-segmented pixels of the factual branch's target objects; the effect of a branch is its endpoint minus the factual endpoint (Table~\ref{tab:s2-defs}).

\begin{table}[t]
  \centering\scriptsize\setlength\tabcolsep{3pt}
  \caption{Metrics (lower is better for the error metrics and ratios, higher otherwise). $\hat\Delta_b$, $\Delta_b$: predicted and simulator effects of branch $b$; $f$: factual branch; $S_{tc}=\cos(\hat\Delta_t,\Delta_c)$.}
  \label{tab:s2-defs}
  \begin{tabular}{@{}p{2.7cm} p{12.6cm}@{}}
    \toprule
    Quantity & Definition \\
    \midrule
    \pairacc{} & Fraction of the four non-factual branches with $S_{tt}>\max_{c\neq t}S_{tc}$; chance $0.25$ \\
    Identity margin & Mean over the four branches of $S_{tt}-\max_{c\neq t}S_{tc}$ \\
    Object-endpoint MAE & Mean absolute error of the endpoint inside the object mask, five branches \\
    Contact-event MAE & Mean absolute error of the effect, $|(\hat P_b-\hat P_f)-(P_b-P_f)|$, inside the object mask, four branches \\
    Global MAE & Mean absolute error over the whole endpoint frames, five branches \\
    Denoise-MSE ratio & One-step denoising error of the ground-truth latent noised to $\sigma{=}5.84$, repaired over native, same noise \\
    Branch/seed SNR & Branch-to-factual difference inside the object mask over the difference between two sampling seeds of one branch; $>1$ is above seed noise \\
    Amplitude & Mean absolute predicted over simulator effect inside the object mask; near 1 is matched \\
    Direction cosine & $\cos(\hat\Delta_b,\Delta_b)$ over all pixels, averaged over the four branches \\
    Zero-response ratio & Contact-event MAE over the mean absolute simulator effect (the contact error of a predictor with no effect); $>1$ is worse than no effect \\
    IE-relative & $\|\hat\Delta-\Delta\|^2/\|\Delta\|^2$ with effects centered over the branches, on latent frames; 1 means no branch difference \\
    \bottomrule
  \end{tabular}
\end{table}

\paragraph{Acceptance test.}
The test, at one checkpoint, has seven conditions: the 95\% lower bound of the \pairacc{} gain (1), the identity-margin gain (2), the object-endpoint MAE reduction (3) and the contact-event MAE reduction (4) is above 0; the upper bound of the global-MAE change is at most $0.005$ (5) and that of the denoise-MSE ratio at most $1.05$ (6); and for every training seed the mean \pairacc{} gain and the mean margin gain are positive (7).
Gains are repaired minus native, paired within a process and averaged per context; intervals are 95\% percentile intervals from 2000 context-bootstrap draws.
Held-out confirmation pools four training seeds and two evaluation seeds, clusters by context and requires the same direction in all four training seeds.

\paragraph{Frozen recipe.}
The adapter is a rank-30 LoRA on the four linear layers of the two action embedders ($30\times41{,}520=1{,}245{,}600$ parameters).
It is trained on the 781 pool contexts with the loss of Eq.~\eqref{eq:loss} (weights $1.0$ direction, $1.0$ object, $1.0$ consequence, $0.3$ anchor, $0.1$ denoising), learning rate $3\times10^{-4}$ and 1024 updates; training seeds are 1701, 3407, 5501 and 7919 and evaluation seeds 3407 and 3413.
For $N$ training contexts, update $u$ uses one context at sampler step $k=\bigl((u{-}1)\bmod N+\lfloor(u{-}1)/N\rfloor\bigr)\bmod 6$.
All selections used demonstration 35; the held-out sets were opened only after the candidate was named.
The consequence weight was chosen by a rule fixed beforehand: each arm is judged separately against native and the earliest passing checkpoint is selected.
Weight 1 passes (\devGateR{} conditions after 1024 updates) and weight 3 passes at no checkpoint (\devConsThreeGateR{}); the second seed pair gives the same outcome (\devBGateR{} and \devBConsThreeGateR{}).

\paragraph{Algorithm.}
Algorithm~\ref{alg:opie} gives the training loop and the step-scheduled deployment of Sec.~\ref{sec:method}.

\begin{algorithm}[t]
\caption{OPIE: training and step-scheduled deployment}
\label{alg:opie}
\footnotesize
\begin{algorithmic}[1]
\State \textbf{Train} $A,B$ (rank $r$) on contexts $s$ with branches $b\in\mathcal{B}$; all other weights frozen.
\For{each update}
  \State pick a context $s$ and a noise level $k$ (cycled), draw one noise $\epsilon$ shared by all branches
  \For{$b\in\mathcal{B}$}
    \State run the model's own sampler to step $k$ to get $x^{(k)}_b$
    \State $\hat x_b\gets D_\theta(x^{(k)}_b,\sigma_k,s,a_b)$ with $W_0{+}BA$; $\hat x^{0}_b\gets$ the same call with $W_0$
  \EndFor
  \State $\hat\Delta_b\gets\hat x_b-\hat x_f$; \; step on $\mathcal{L}$ of Eq.~\ref{eq:loss}
\EndFor
\State \textbf{Deploy:} in denoiser call $i$, use $W_0{+}BA$ if $\sigma_i\ge\tau$, else $W_0$ (Eq.~\ref{eq:gate}).
\end{algorithmic}
\end{algorithm}

\paragraph{Selection of the step gate.}
The deployed schedule (adapter active only at high noise) was selected on the 39 development contexts with frozen checkpoints, two training and two evaluation seeds, and no further training.
The rule, written before the sweep: a variant is eligible if its margin-gain lower bound is above zero and its all-branch FVD ratio to native is at most $1.10$ (point estimate); the eligible variant with the largest margin gain is selected.
In Table~\ref{tab:s2-gate} three Prophet variants are eligible and the gate $\sigma\ge0.5$ (the four highest-noise of six calls) has the largest margin gain; for A2World only $\sigma\ge5$ is eligible, chosen on its own development set.

\begin{table}[t]
  \centering\scriptsize\setlength\tabcolsep{3pt}
  \caption{Step-gate selection on the 39 development contexts. Margin gain over native (95\% CI); all-branch FVD ratio to native (95\% bootstrap CI). $\sigma$ is the noise level of the denoiser call; the adapter is active where the condition holds. Selected variants are marked *.}
  \label{tab:s2-gate}
  \begin{tabular}{@{}l l l l c@{}}
    \toprule
    Model & Adapter active & Margin gain & FVD ratio & Eligible \\
    \midrule
    Prophet & every call & +0.083\ci{+0.054}{+0.113} & 1.42\ci{1.14}{1.65} & no \\
    Prophet & $\sigma\ge0.5$* & +0.082\ci{+0.055}{+0.112} & 1.07\ci{0.91}{1.18} & yes \\
    Prophet & $\sigma\ge5$ & +0.062\ci{+0.036}{+0.089} & 1.17\ci{1.02}{1.21} & no \\
    Prophet & every call, scale 0.75 & +0.060\ci{+0.039}{+0.083} & 1.08\ci{0.95}{1.16} & yes \\
    Prophet & every call, scale 0.5 & +0.034\ci{+0.020}{+0.049} & 1.02\ci{0.92}{1.10} & yes \\
    \midrule
    A2World & every call & +0.013\ci{+0.001}{+0.024} & 1.18\ci{1.09}{1.25} & no \\
    A2World & $\sigma\ge0.5$ & +0.009\ci{-0.002}{+0.020} & 1.05\ci{0.99}{1.07} & no \\
    A2World & $\sigma\ge5$* & +0.012\ci{+0.004}{+0.020} & 1.03\ci{0.99}{1.06} & yes \\
    \bottomrule
  \end{tabular}
\end{table}

\paragraph{Compute.}
The recipe adapter has $1{,}245{,}600$ trainable parameters on Prophet and on A2World (peak training memory 12.4 and 9.75 GiB), the full-rank embedder arms $71{,}720{,}960$, the backbone adapter (rank 16 on 280 projections) $22{,}937{,}600$, and the UWM adapter 944,850.
The median cost is $1.80$ GPU-h per training unit of the frozen recipe ($28{,}140$ forward passes) and $1.05$ GPU-h per development evaluation unit.
All project jobs with GPUs, including exploratory runs, allocated $3{,}329.5$ GPU-h (564 jobs).

\paragraph{Reproducibility and released code.}
Seeds are as above unless stated; every gain is paired within a process (S1).
We will release the five-action diagnostic, the evaluation code (held-out scoring, FVD, classifier readout, closed-loop selection, statistics) and a reference implementation of \ours{} for Prophet (adapter, losses, on-policy training, step gate), together with the remaining ports, adapters and the full protocol, upon publication; model weights and simulator data are not included.

\section{Diagnosis details}
\label{supp:n3}

This section gives the held-out repeat of the native diagnosis (Sec.~\ref{sec:diag-spectrum}) and the checks behind the design constraints (Sec.~\ref{sec:diag-constraints}).

\paragraph{Native diagnosis on the held-out contexts.}
The native diagnosis was repeated with the same code, seeds and metrics on the held-out contexts (Prophet and A2World 236, UWM 235; four evaluation seeds; native models only, nothing selected; Table~\ref{tab:s3-heldout}).
Four readouts were registered before the run: the A2World minus Prophet direction cosine is above zero, Prophet's object SNR is above one, UWM's latent SNR is below one, and Prophet's contact error exceeds that of the zero-response predictor ($1.36\times$, context-level $[1.31,1.41]$).
The first three hold at the context, demonstration and task level, the contact-error readout at the context level.
Held-out values sit close to the development values and the ordering is unchanged.

\begin{table}[t]
\centering\scriptsize
\setlength{\tabcolsep}{2.5pt}
\caption{Native diagnosis on development (39 contexts) and held-out contexts (236; UWM 235, latent space), context-clustered 95\% intervals.}
\label{tab:s3-heldout}
\begin{tabular}{@{}l cc cc cc@{}}
\toprule
 & \multicolumn{2}{c}{Prophet} & \multicolumn{2}{c}{A2World} & \multicolumn{2}{c}{UWM} \\
\cmidrule(lr){2-3}\cmidrule(lr){4-5}\cmidrule(l){6-7}
Metric & Dev. & Held-out & Dev. & Held-out & Dev. & Held-out \\
\midrule
Direction cosine & $0.166\,[0.123,0.209]$ & $0.136\,[0.119,0.152]$ & $0.587\,[0.555,0.619]$ & $0.574\,[0.555,0.591]$ & $0.159\,[0.141,0.177]$ & $0.147\,[0.138,0.155]$\\
Object SNR & $1.35\,[1.21,1.50]$ & $1.29\,[1.23,1.36]$ & $1.85\,[1.64,2.07]$ & $1.77\,[1.68,1.85]$ & $0.240\,[0.214,0.266]$ & $0.232\,[0.221,0.243]$\\
Object amplitude ratio & $0.95\,[0.80,1.12]$ & $0.91\,[0.85,0.97]$ & $1.23\,[1.11,1.41]$ & $1.16\,[1.12,1.21]$ & $0.249\,[0.224,0.276]$ & $0.241\,[0.230,0.253]$\\
\pairacc{} & $0.607\,[0.540,0.672]$ & $0.549\,[0.520,0.575]$ & $0.862\,[0.827,0.896]$ & $0.856\,[0.840,0.873]$ & $0.739\,[0.696,0.779]$ & $0.703\,[0.681,0.722]$\\
Identity margin & $0.034\,[0.011,0.057]$ & $0.016\,[0.007,0.026]$ & $0.234\,[0.211,0.256]$ & $0.218\,[0.205,0.229]$ & $0.054\,[0.047,0.061]$ & $0.048\,[0.044,0.052]$\\
\bottomrule
\end{tabular}
\end{table}

\paragraph{Residual adapter in the middle of the network.}
A residual adapter in the middle of Prophet, fed the action through its own small encoder and trained on-policy at the model's sampler states (20 units), learned a residual that did not depend on the action: its branch-difference cosine equals the native model's through 256 updates.

\paragraph{Thirty-five against five sampling steps.}
On four contexts (five branches, four seeds, no training), 35 instead of 5 sampling steps change the paired native margin by $+0.0005$\ci{+0.0003}{+0.0008}.
The branch/seed SNR on these contexts is $0.93$\ci{0.71}{1.16} at 5 steps and $0.80$\ci{0.51}{1.08} at 35 steps; the 39-context diagnosis gives \dgPSnr{}.

\paragraph{Ground-truth-noised against sampler states.}
Table~\ref{tab:s3-trace} gives the native intervention-effect error $\|\hat\Delta-\Delta\|^2/\|\Delta\|^2$ at each sampler noise level: on ground-truth-noised states it falls from 1.15 at $\sigma{=}80$ to 0.19 at $\sigma{=}0.965$ and 0.012 at $\sigma{=}0.085$, while on the sampler's own states it stays between 1.00 and 1.20.

\begin{table}[t]
  \centering\scriptsize\setlength\tabcolsep{4pt}
  \caption{Native intervention-effect error ratio (mean over four training seeds) at the six sampler noise levels of Prophet.}
  \label{tab:s3-trace}
  \begin{tabular}{@{}l cccccc@{}}
    \toprule
    $\sigma$ & 80 & 24.4 & 5.84 & 0.965 & 0.085 & 0.002 \\
    \midrule
    Ground-truth-noised states & 1.155 & 1.041 & 0.702 & 0.186 & 0.0120 & $6.7\times10^{-6}$ \\
    Sampler states & 1.175 & 1.111 & 1.065 & 0.999 & 1.081 & 1.195 \\
    \bottomrule
  \end{tabular}
\end{table}

\paragraph{How actions enter Prophet.} Two MLPs with 71.7M parameters turn the flattened 280-dimensional action chunk into an addition to the timestep embedding and an addition to the shared adaLN-LoRA modulation. The native sensitivity to the branch actions sits in the adaLN-LoRA path: the timestep part is nearly independent of the action (relative change against the factual branch $3.0\%$ for gripper opposite and $0.1$--$0.4\%$ for the other branches), whereas the modulation part changes by $0.36$ to $1.29$ between branches.

\section{Held-out identity results}
\label{supp:n4}
This section backs Table~\ref{tab:results} and Sec.~\ref{sec:exp-main}: the registered held-out test, the cut-off of the step gate, absolute values and the factual-branch flow cosine.

\paragraph{Registered test.}
The schedule (adapter off for the last two of six denoiser calls) was fixed before this read, after selection on demonstration 35. The candidate uses the same four update-1024 weights as the always-on repair (training seeds 1701, 3407, 5501, 7919), on 236 held-out contexts (fresh 117, sealed 119), two evaluation seeds (3407, 3413) and five branches, with shared context and sampling noise. The three registered endpoints are the scheduled-minus-native identity-margin gain (95\% lower bound above zero), the all-branch FVD ratio to native (upper bound at most $1.10$) and the scheduled-minus-always-on margin difference (lower bound above $-0.02$); all three pass in the pooled set (Table~\ref{tab:s4-endpoints}). Intervals use 2000 context-bootstrap draws (pooled FVD 1000); FVD uses I3D on frames 1--20; error gains are signed so that positive is a reduction. The margin gain is positive for each training seed, from $\sdSeedPooledLo{}$ to $\sdSeedPooledHi{}$, and in the fresh and the sealed set alone.

\begin{table}[h]
\centering\scriptsize
\setlength{\tabcolsep}{4pt}
\begin{tabular}{@{}p{4.6cm}lccc@{}}
\toprule
Endpoint & Registered rule & Pooled (236) & Fresh (117) & Sealed (119)\\
\midrule
Scheduled $-$ native, identity margin & lower bound $>0$ & \ptci{\scXAuditASValue}{\scXAuditASValueCI} & \ptci{\scXAuditTValue}{\scXAuditTValueCI} & \ptci{\scXAuditBRValue}{\scXAuditBRValueCI}\\
Scheduled $-$ always-on, identity margin & lower bound $>-0.02$ & \ptci{\scXAuditANValue}{\scXAuditANValueCI} & \ptci{\scXAuditOValue}{\scXAuditOValueCI} & \ptci{\scXAuditBMValue}{\scXAuditBMValueCI}\\
FVD ratio to native, all branches, scheduled & upper bound $\le1.10$ & \ptci{\scXAuditAIValue}{\scXAuditAIValueCI} & \ptci{\scXAuditJValue}{\scXAuditJValueCI} & \ptci{\scXAuditBHValue}{\scXAuditBHValueCI}\\
\addlinespace[2pt]
FVD ratio to native, all branches, always-on & secondary & \ptci{\scXAuditAHValue}{\scXAuditAHValueCI} & \ptci{\scXAuditIValue}{\scXAuditIValueCI} & \ptci{\scXAuditBGValue}{\scXAuditBGValueCI}\\
FVD ratio to native, factual branch, scheduled & secondary & \ptci{\scXAuditAKValue}{\scXAuditAKValueCI} & \ptci{\scXAuditLValue}{\scXAuditLValueCI} & \ptci{\scXAuditBJValue}{\scXAuditBJValueCI}\\
Scheduled $-$ native, \pairacc{} & secondary & \ptci{\scXAuditAVValue}{\scXAuditAVValueCI} & \ptci{\scXAuditWValue}{\scXAuditWValueCI} & \ptci{\scXAuditBUValue}{\scXAuditBUValueCI}\\
Scheduled $-$ native, object endpoint MAE & secondary & \ptci{\scXAuditAUValue}{\scXAuditAUValueCI} & \ptci{\scXAuditVValue}{\scXAuditVValueCI} & \ptci{\scXAuditBTValue}{\scXAuditBTValueCI}\\
Scheduled $-$ native, contact-event error & secondary & \ptci{\scXAuditAQValue}{\scXAuditAQValueCI} & \ptci{\scXAuditRValue}{\scXAuditRValueCI} & \ptci{\scXAuditBPValue}{\scXAuditBPValueCI}\\
\bottomrule
\end{tabular}
\caption{Registered held-out test of the step-scheduled repair, pooled and by split (95\% context-sampling intervals). The first three rows are the registered endpoints.}
\label{tab:s4-endpoints}
\end{table}

\paragraph{Cut-off of the step gate.}
The frozen adapter is enabled in the leading $k$ of the six denoiser calls ($k=1,\dots,6$; $k=4$ is the deployed schedule, $k=6$ the always-on repair) or only in the last two calls, for each of the four training seeds on the $\eCutNContexts{}$ held-out contexts and two evaluation seeds of the registered test. The sweep is descriptive; the cut-off stays at the value selected on the development demonstration (Table~\ref{tab:s4-tau}).

\begin{table}[h]
\centering
\caption{Cut-off sweep: identity-margin gain over native (context bootstrap interval) and FVD ratio to native, adapter on the first $k$ of six calls or the last two only.}
\label{tab:s4-tau}
\footnotesize
\begin{tabular}{@{}lccc@{}}
\toprule
Adapter on for & Gain & Context CI & FVD ratio\\
\midrule
first call only & $-0.005$ & $[-0.006, -0.004]$ & $1.013$\\
first 2 calls & $+0.004$ & $[-0.000, +0.009]$ & $1.178$\\
first 3 calls & $+0.077$ & $[+0.064, +0.089]$ & $1.327$\\
first 4 calls (deployed, $\tau=0.5$) & $+0.096$ & $[+0.082, +0.109]$ & $1.021$\\
first 5 calls & $+0.096$ & $[+0.083, +0.109]$ & $1.609$\\
all 6 calls (always-on) & $+0.096$ & $[+0.083, +0.109]$ & $1.623$\\
last 2 calls only & $+0.000$ & $[-0.000, +0.001]$ & $1.748$\\
\bottomrule
\end{tabular}

\end{table}

With the adapter on for the first four calls the gain keeps $\eCutFracFour{}$ of the always-on gain ($\scXFullMargin{}$) at an FVD ratio of $\scXFVD{}$, against $\scXFullFVD{}$ for always-on. The gain forms in the high-noise calls: it appears at the third call ($\eCutKThree{}$) and is complete at the fourth ($\eCutKFour{}$, an increment over $k=3$ of $\eCutFourThree{}\eCutFourThreeCI{}$); calls 5 and 6 add nothing to the margin ($k=6$ minus $k=4$: $\eCutSixFourMargin{}\eCutSixFourMarginCI{}$) and only raise the FVD ratio. Always-on minus deployed in \pairacc{} is $\eCutSixFourPair{}\eCutSixFourPairCI{}$. Enabling the adapter only in the last two calls adds a margin gain of $\eCutLate{}$ and a \pairacc{} change of $\eCutPairLate{}$. Fresh and sealed show the same curve. Only the deployed four-call gate ($\tau=0.5$) combines the full gain with an FVD ratio inside the registered bound.

\paragraph{Distribution distance and absolute endpoint errors.}
Table~\ref{tab:s4-fvd} gives the absolute values on the held-out contexts: identity readouts, FVD against the ground-truth videos ($\eFvdNContexts{}$ contexts, five branches each; native with both evaluation seeds stacked, repaired rows as the mean over four training seeds) and endpoint errors ($\eAbsNContexts{}$ contexts, lower is better).
The generation-seed floor, the FVD between the native sets of the two evaluation seeds, is $\eFvdFloor{}$; the scheduled-minus-native difference is $\eFvdSchedOverFloor{}$ times this floor and its interval contains zero, so the $1.10$ bound is a convention fixed in advance.

\begin{table}[t]
  \centering\scriptsize\setlength\tabcolsep{4pt}
  \caption{Absolute values on the held-out contexts, native against always-on and scheduled repair; differences and reductions (gain over the native mean) with 95\% context-bootstrap intervals.}
  \label{tab:s4-fvd}
  \begin{tabular}{@{} l c c c c c @{}}
    \toprule
    Quantity & Native & Always-on & Scheduled & Always-on vs.\ native & Scheduled vs.\ native \\
    \midrule
    \pairacc{} & 0.551 & 0.707 & 0.714 & --- & --- \\
    Identity margin & 0.016 & 0.112 & 0.112 & --- & --- \\
    FVD, all branches & \eFvdNative{} & \eFvdFull{} & \eFvdSched{} & \ptci{\eFvdFullNat}{\eFvdFullNatCI} & \ptci{\eFvdSchedNat}{\eFvdSchedNatCI} \\
    FVD, factual branch & \eFvdFactNative{} & \eFvdFactFull{} & \eFvdFactSched{} & --- & --- \\
    Object endpoint & \eAbsObjNat{} & \eAbsObjFull{} & \eAbsObjSched{} & \ptci{\eAbsObjRelFull}{\eAbsObjRelFullCI}\,\% & \ptci{\eAbsObjRelSched}{\eAbsObjRelSchedCI}\,\% \\
    Contact event (object) & \eAbsContactNat{} & \eAbsContactFull{} & \eAbsContactSched{} & \ptci{\eAbsContactRelFull}{\eAbsContactRelFullCI}\,\% & \ptci{\eAbsContactRelSched}{\eAbsContactRelSchedCI}\,\% \\
    Global endpoint & \eAbsGlobalNat{} & \eAbsGlobalFull{} & \eAbsGlobalSched{} & \ptci{\eAbsGlobalRelFull}{\eAbsGlobalRelFullCI}\,\% & \ptci{\eAbsGlobalRelSched}{\eAbsGlobalRelSchedCI}\,\% \\
    \bottomrule
  \end{tabular}
\end{table}

\paragraph{Flow cosine of the factual branch.}
Sec.~\ref{sec:exp-main} reports the direction cosine between the optical flow of the predicted factual video and that of the ground truth (Farneback flow, base-model metric functions; pixels where both flow magnitudes exceed $10^{-3}$; whole 21-frame clip) on the $\eFlowNContexts{}$ held-out contexts, four training seeds and two evaluation seeds. The cosine is $\eFlowDirNat{}$ for native, $\eFlowDirFull{}$ for always-on and $\eFlowDirSched{}$ for scheduled repair; scheduled minus native is $\eFlowDirSchedNat{}\eFlowDirSchedNatCI{}$ (paired context bootstrap). The ground truth scored against itself gives $\eFlowGtDir{}$ rather than 1 (a $10^{-6}$ term in the denominator), the ceiling for these values. The metric measures fit to the ground-truth factual video, not action identity.

\section{Frame-level realism and the gate against strength}
\label{supp:n5}
This section supports the realism paragraph of Sec.~\ref{sec:exp-main} and Fig.~\ref{fig:gate}a: frame-level readouts and the step gate against a lower adapter strength.

\paragraph{Frame-level realism.}
Three frame-level readouts of the stored held-out endpoints (final two frames of every branch; 236 contexts, two evaluation seeds) were registered before any endpoint was read: (i) the Fr\'echet distance between generated and simulator frame sets in frozen DINOv2 features (FD-DINOv2); (ii) LPIPS (AlexNet) between each generated frame and its simulator frame; (iii) sharpness, the variance of the Laplacian relative to the simulator frame. Ratios are to native, with 95\% demonstration-cluster bootstrap intervals (2000 draws). Registered rule: upper bound of the FD-DINOv2 and LPIPS ratios of ours below $1.10$, lower bound of the sharpness ratio above $0.90$. The same contexts were re-inferred with the scheduled adapter's strength scaled by $0.75$, $0.5$ and $0.25$ (registered before the contexts were read).

\begin{table}[h]
  \centering\scriptsize\setlength\tabcolsep{4pt}
  \caption{Frame-level readouts as ratios to native (95\% demonstration-cluster intervals) and identity-margin gain over native (95\% intervals over contexts); lower is better for FD-DINOv2 and LPIPS, $1$ is native for sharpness. Strength rows: scheduled adapter with scaled strength.}
  \label{tab:s5-realism}
  \begin{tabular}{@{} l c c c c @{}}
    \toprule
    Variant & FD-DINOv2 & LPIPS to simulator & Sharpness & Margin gain \\
    \midrule
    Native (absolute) & $\rlFdNat{}$ & $\rlLpNat{}$ & $1.00$ & --- \\
    Ours (scheduled) & $\rlFdSched{}\rlFdSchedCI{}$ & $\rlLpSched{}\rlLpSchedCI{}$ & $\rlShSched{}\rlShSchedCI{}$ & \scXMargin\scXMarginCI{} \\
    Always-on & $\rlFdFull{}\rlFdFullCI{}$ & $\rlLpFull{}\rlLpFullCI{}$ & $\rlShFull{}\rlShFullCI{}$ & \scXFullMargin\scXFullMarginCI{} \\
    Strength $0.75$ & $\rlFdThreeQ{}$ & $\rlLpThreeQ{}$ & $\rlShThreeQ{}$ & \gsScAThreeQ\gsScAThreeQCI{} \\
    Strength $0.5$ & $\rlFdHalf{}$ & $\rlLpHalf{}$ & $\rlShHalf{}$ & \gsScAHalf\gsScAHalfCI{} \\
    Strength $0.25$ & $\rlFdQuarter{}$ & $\rlLpQuarter{}$ & $\rlShQuarter{}$ & \gsScAQuarter\gsScAQuarterCI{} \\
    \bottomrule
  \end{tabular}
\end{table}

LPIPS and sharpness meet the registered rule. Ours is closer to the simulator frame than native in LPIPS, and in fewer than 1\% of contexts is it worse than native by more than the native seed-to-seed difference (null share $\rlLpShareNull\%$). The DINOv2 Fr\'echet ratio is the readout that moves with the adapter strength: it is $\rlFdHalf{}$ at strength $0.5$ and $\rlFdSched{}$ at full strength, against the $1.10$ reference. The deployed setting is full strength because the identity gain and the FVD bound are the primary endpoints, and the cost of the full gain is a lower sharpness at the endpoint frames ($\rlShShareSched\%$ of contexts against $\rlShShareFull\%$ for always-on and $\rlShShareNull\%$ between native seeds). The step gate recovers part of the always-on shift in every readout (Fig.~\ref{fig:teaser}a), and a lower strength is available where frame statistics matter more than the margin.

The margin gain falls roughly in proportion to the strength, whereas switching the adapter off for the last two calls leaves it unchanged (scheduled minus always-on $\gsScGateKeep\gsScGateKeepCI{}$); fresh and sealed show the same pattern. At strength $0.5$ the Fr\'echet distance matches native while the classifier gain of S6 is $\idrScHalf{}$.

\paragraph{Step gate against a lower strength.}
Because scaling the adapter down also moves its frame statistics towards native, the gate is compared with always-on repair at the same strength. In the same processes (paired contrasts) we ran native, the step gate and the always-on adapter at strengths $1.0$, $0.75$ and $0.5$, plus a gate that keeps the adapter on for the first three calls only (strength $0.75$), for training seeds 1701 and 3407 and evaluation seeds 3407 and 3413, on the $\gsContexts{}$ held-out contexts ($\gsRollouts{}$ rollouts), registered with its outcome rules before any rollout. The primary contrasts at strengths $0.75$ and $0.5$ are gate minus always-on in the DINOv2 Fr\'echet ratio (lower is better) and the sharpness ratio (higher is better), read only if the margin gains differ by less than $0.02$ (95\% demonstration-cluster intervals).

\begin{table}[h]
  \centering\scriptsize\setlength\tabcolsep{3pt}
  \caption{Step gate against always-on at the same strength (ratios to native; margin and classifier gains over native; 236 contexts, two training seeds).}
  \label{tab:s5-gate}
  \begin{tabular}{@{} l c c c c c @{}}
    \toprule
    Deployment & Strength & Margin & Classifier & FD-DINOv2 & Sharpness \\
    \midrule
    Step gate & $1.0$ & $\gsSchedFullMarg{}$ & $\gsSchedFullCls{}$ & $\gsSchedFullFd{}$ & $\gsSchedFullSh{}$  \\
    Always-on & $1.0$ & $\gsAlwaysFullMarg{}$ & $\gsAlwaysFullCls{}$ & $\gsAlwaysFullFd{}$ & $\gsAlwaysFullSh{}$  \\
    Step gate & $0.75$ & $\gsSchedThreeQMarg{}$ & $\gsSchedThreeQCls{}$ & $\gsSchedThreeQFd{}$ & $\gsSchedThreeQSh{}$  \\
    Always-on & $0.75$ & $\gsAlwaysThreeQMarg{}$ & $\gsAlwaysThreeQCls{}$ & $\gsAlwaysThreeQFd{}$ & $\gsAlwaysThreeQSh{}$  \\
    Step gate & $0.5$ & $\gsSchedHalfMarg{}$ & $\gsSchedHalfCls{}$ & $\gsSchedHalfFd{}$ & $\gsSchedHalfSh{}$  \\
    Always-on & $0.5$ & $\gsAlwaysHalfMarg{}$ & $\gsAlwaysHalfCls{}$ & $\gsAlwaysHalfFd{}$ & $\gsAlwaysHalfSh{}$  \\
    On for first three calls & $0.75$ & $\gsCutThreeMarg{}$ & $\gsCutThreeCls{}$ & $\gsCutThreeFd{}$ & $\gsCutThreeSh{}$  \\
    \bottomrule
  \end{tabular}
\end{table}

{\sloppy At strength $0.75$ the gate minus always-on is $\gsDiffThreeQFd\gsDiffThreeQFdCI{}$ in the Fr\'echet ratio and $\gsDiffThreeQSh\gsDiffThreeQShCI{}$ in sharpness, with margin gains that differ by $\gsDiffThreeQMarg\gsDiffThreeQMargCI{}$; at strength $0.5$ the differences are $\gsDiffHalfFd\gsDiffHalfFdCI{}$, $\gsDiffHalfSh\gsDiffHalfShCI{}$ and $\gsDiffHalfMarg\gsDiffHalfMargCI{}$. All four realism intervals exclude zero in the favorable direction and the margin gains agree within the tolerance, so the registered reading holds: the gate improves realism at matched strength. The FVD ratio of the gated adapter stays at or below $\gsSchedFullFvd{}$ at every strength, while always-on reaches $\gsAlwaysFullFvd{}$ at full strength. The first-three-calls gate gives a lower margin gain ($\gsCutThreeMarg{}$ against $\gsSchedThreeQMarg{}$) without a realism gain. The gate is therefore not equivalent to a lower dose: at the same strength it keeps the identity and classifier gains and lowers the frame-statistic cost.\par}

\section{Identity read without our loss}
\label{supp:n6}
This section supports the paragraph ``Identity read without our loss'' of Sec.~\ref{sec:exp-main}. The identity margin and the training loss are both cosines of predicted branch differences, so we registered four readouts that share neither their form nor their training data: a frame classifier, an object-position probe, a blind human study and a control adapter trained with shuffled branch targets. The first three were registered, with adequacy checks and decision rules, before any held-out endpoint or response was read. Classifier and probe use the endpoints of the adapter-strength re-inference run of Supp.~S5.

\paragraph{Classifier readout.}
A linear ridge head on frozen DINOv2 patch features of a branch's final frame and of its difference to the conditioning frame decides which of the five branches (factual, gripper opposite, hold pose, translation axis flip, translation half) a frame belongs to. It is trained only on ground-truth simulator frames of the training pool ($\idrTrainSamples{}$ samples, $\idrTrainCtx{}$ contexts $\times$ five branches, four registered blur and noise levels) and never sees a generated frame or the loss. On the 39 development ground-truth contexts all four registered checks passed: five-way accuracy $\idrDevAcc{}$ (threshold $0.60$; chance $0.20$), lowest per-class recall $\idrDevMinRec{}$ ($0.30$), factual-versus-gripper accuracy $\idrDevBin{}$ ($0.75$) and accuracy $\idrDevBlur{}$ under level-2 blur, $\idrDevBlurRatio{}$ of clean ($0.90$). On the $\idrN{}$ held-out ground-truth frame sets the accuracy is $\idrGT{}$ (registered floor $0.60$). The readout is the share of branches whose stored generated endpoint is assigned to that branch's own action.

\begin{table}[h]
  \centering\scriptsize\setlength\tabcolsep{3pt}
  \caption{Readouts without our loss on the stored held-out endpoints, gains over native. (a)~Five-way branch accuracy of the classifier (95\% intervals clustered by context, demonstration and task; pooled over $\idrN{}$ contexts unless stated); the top row is the registered primary test. (b)~Object-position probe on the 236 contexts (95\% intervals over contexts; $\pbNEligDir{}$ eligible branches for direction, $\pbNEligRet{}$ for retrieval); the first two rows are the registered primary contrasts.}
  \label{tab:s6-classifier}
  \par\vspace{4pt}{\footnotesize(a) Classifier}\\[1pt]
  \begin{tabular}{@{} l c c c c @{}}
    \toprule
    Contrast & Gain & Context interval & Demonstration interval & Task interval \\
    \midrule
    Scheduled $-$ native & \idrGain{} & \idrGainCI{} & \idrGainDemoCI{} & \idrGainTaskCI{} \\
    \quad fresh contexts ($\idrNFresh{}$) & \idrFresh{} & \idrFreshCI{} & \idrFreshDemoCI{} & \idrFreshTaskCI{} \\
    \quad sealed contexts ($\idrNSealed{}$) & \idrSealed{} & \idrSealedCI{} & \idrSealedDemoCI{} & \idrSealedTaskCI{} \\
    Always-on $-$ native & \idrFull{} & \idrFullCI{} & \idrFullDemoCI{} & \idrFullTaskCI{} \\
    Scheduled $-$ always-on & \idrKeep{} & \idrKeepCI{} & \idrKeepDemoCI{} & \idrKeepTaskCI{} \\
    Strength $0.5$ $-$ native & \idrScHalf{} & \idrScHalfCI{} & \idrScHalfDemoCI{} & \idrScHalfTaskCI{} \\
    \bottomrule
  \end{tabular}\\[4pt]
  {\footnotesize(b) Object-position probe}\\[1pt]
  \begin{tabular}{@{} l c c @{}}
    \toprule
    Contrast & Direction cosine & Retrieval accuracy \\
    \midrule
    Scheduled $-$ native & \pbDirSched\pbDirSchedCI{} & \pbRetSched\pbRetSchedCI{} \\
    Always-on $-$ native & \pbDirFull\pbDirFullCI{} & \pbRetFull\pbRetFullCI{} \\
    Scheduled $-$ always-on & \pbDirGateKeep\pbDirGateKeepCI{} & \pbRetGateKeep\pbRetGateKeepCI{} \\
    \bottomrule
  \end{tabular}
\end{table}

Native endpoints are assigned to their own branch with accuracy $\idrNat{}$ and scheduled endpoints with $\idrSched{}$; the scheduled model closes $\idrGap{}$ of the gap to the ground-truth frames ($\idrGT{}$). The gain passes the registered test: it exceeds $+0.05$, its demonstration interval stays above $+0.02$, and it is positive at both evaluation seeds ($\idrSeedA{}$ and $\idrSeedB{}$; the four training seeds give $\idrTrainMin{}$ to $\idrTrainMax{}$). It holds under all three clustering units, in the fresh and the sealed contexts separately, and in $\idrBetter{}$ of $\idrN{}$ contexts (worse in $\idrWorse{}$). Across the $20$ (variant, training seed) points the gain tracks the identity-margin gain with Spearman $\rho=\idrRho{}$ (registered threshold $0.80$). The difference between the two native evaluation seeds is $\idrNull{}$, of opposite sign. Object-level consequences are read by the probe below.

\paragraph{Object-position probe.}
The probe is a per-task ridge regression on frozen DINOv2 features (block 19, $6{\times}6$ pooling), trained on the training pool only. The five registered adequacy checks (position $R^2\ge0.90$, median position error $\le2.0$\,cm, displacement cosine $\ge0.80$, branch retrieval accuracy $\ge0.80$, relative error $\le0.50$) passed before any held-out endpoint was read. The probe estimates the object position from a branch's final predicted frame; the displacement relative to the factual branch is compared with the simulator's, on branches whose true displacement exceeds a registered threshold. The two primary readouts are the direction cosine of predicted against true displacement and the accuracy of retrieving the correct branch of a context by displacement. On the 236 held-out ground-truth frames the probe has position $R^2$ \pbRsq{}, median error \pbMedErr\,cm, displacement cosine \pbDispCos{} and retrieval \pbRetGT{}.

The scheduled model improves both primary readouts over native, with intervals above zero and above the generation-noise floor between two native evaluation seeds, and the step gate keeps the gain of the always-on adapter.

\paragraph{Blind human study.}
To read the repair without any learned metric, we registered a blind two-alternative study before any response existed. Each item shows the simulator's start frame, its final frame (the truth) and two generated final frames for the same context and action; the rater picks the one closer to the truth in where the outlined objects ended up and whether they moved (A, B or cannot tell). Outlines come from the start-frame object mask and are identical on all images. Eighty items compare native with ours (training seeds 1701 and 3407, evaluation seed 3407), drawn from held-out contexts in which an intervention moved the target by at least 1\,cm. Sides and order are randomized per rater and no condition label is visible. Eight attention checks and four practice items with feedback are included; a rater is valid with at most one failed check and at least 90\% of items answered. The primary test is the item-level majority over valid raters (ties excluded) preferring ours with a 95\% Clopper--Pearson lower bound above $0.5$, with at least eight valid raters. Consent, payment and anonymity are described in Supp.~S11.

$\hxRaters{}$ students outside robot learning and video generation took part; none was an author. $\hxValid{}$ were valid; the other $\hxExcl{}$ failed more than one attention check. On the $\hxN{}$ items with a clear majority, the majority preferred ours over native on $\hxK{}$ and native on $\hxOther{}$ ($\hxShare{}$, 95\% interval $\hxCI{}$, sign test $p\hxP{}$; $\hxTies{}$ ties and $\hxNone{}$ items without a decisive answer), so the registered primary is supported. Every valid rater individually leaned toward ours (shares $\hxRaterMin{}$ to $\hxRaterMax{}$). Leaving out one valid rater at a time keeps the lower bound at or above $\hxLooMin{}$ in all $\hxLooRuns{}$ runs. Cannot tell was answered on $\hxCT\%$ of items and does not enter the vote. Without any learned feature, readers judge our endpoints closer to the simulator's outcome than native's.

\paragraph{Control adapter with shuffled branch targets.}
A generic adapter could shift frame statistics without linking any action to its own consequence; the control removes that link. It is trained like the recipe adapter (781 contexts, 1024 updates, same optimizer, rank, seeds, data order and noise; Supp.~S2) except that within each context the targets of the four non-factual branches are permuted by a fixed derangement, so every counterfactual action is supervised with the consequence of a different one. Four control adapters are deployed with the same step gate and strength. The control and its analysis rules were registered before any control adapter existed.

On the $\ctlOfN{}$ held-out contexts (fresh and sealed pooled; $\ctlOfDemos{}$ demonstrations, four training seeds) the control does not raise the identity margin ($\ctlOfCtlMarg\ctlOfCtlMargCI{}$ against $\ctlOfRepMarg\ctlOfRepMargCI{}$ for the recipe adapter; difference $\ctlOfDiff\ctlOfDiffCI{}$) and lowers \pairacc{} ($\ctlOfPair\ctlOfPairCI{}$ against $\ctlOfPairRep{}$). The classifier gain is $\ctlOfClsCtl\ctlOfClsCtlCI{}$ for the control against $\idrGain{}$ for ours. The frame statistics move the same way for both (LPIPS ratio to native $\ctlOfLpCtl{}$ against $\ctlOfLpRep{}$ for ours; sharpness, Fr\'echet distance and the object, contact and global errors alike); only the FVD ratio differs: $\ctlOfFvdCtl{}$ for the control against $\ctlOfFvdRep{}$ for ours. The shifts in frame statistics are generic to training an adapter at these states; the identity gain and the FVD bound belong to the identity supervision.

The control adapters also ran in the closed-loop contexts of Supp.~S7 (ten tasks, 50 initial states, policy seeds 32 to 35, $K=4$, step gate, strength $1.0$). At the first query ($\ctlQContexts{}$ contexts with a defined ranking) the Spearman correlation between imagined scores and simulator outcomes of the four candidates is $\ctlRhoRep{}$ for ours, $\ctlRhoCtl{}$ for the control and $\ctlRhoNat{}$ for native (ours minus control $\ctlRhoDiff\ctlRhoCI{}$, positive in all $\ctlRhoTasksPos{}$ tasks), and the simulator's best candidate is picked in $\ctlTopRep{}$, $\ctlTopCtl{}$ and $\ctlTopNat{}$ of the decisions (ours minus control $\ctlTopDiff\ctlTopCI{}$; chance $0.25$). The identity supervision produces the identity-margin gain, the classifier gain, the FVD bound and the better candidate ranking.

\section{Closed-loop protocol and registered tests}
\label{supp:n7}

This section documents the closed-loop protocol of Sec.~\ref{sec:exp-down} and the tests behind Table~\ref{tab:closed} and Fig.~\ref{fig:teaser}c.
All numbers are best-of-four selection success with a learned value head in the simulator, not official benchmark success.

\paragraph{Protocol.}
A fixed UWM policy, fine-tuned on the ten LIBERO-10 tasks, proposes $K=4$ action chunks of 16 steps; candidate 0 is the policy alone.
The WAM imagines every chunk for 20 steps from the real history, a frozen value head scores the frame after 16 steps, and the highest-scoring chunk is executed.
An episode ends at success or at 700 control steps after 10 settling steps; $H=520$ is the official limit (secondary horizon). All arms share the policy's random draws and the WAM's noise, so they differ only in the choice among the same candidates.
Arms: policy alone; random choice; native; always-on repair; ours (the step-gated adapter); and the oracle, which executes each candidate for 16 steps from the restored simulator state and takes the best by the value head on the real frame (2.0 for task success).
The value head is a small convolutional network on $128\times128$ agent-view images, trained on real frames of 500 logged policy episodes only and frozen before any WAM frame was scored; its AUROC on real frames is $\apAurocReal{}$ against a registered gate of $0.90$.
Each task uses its 50 official initial states; intervals are cluster bootstraps over (task, initial state), and task-level tests are exact sign-flips over the ten per-task mean differences.
Tasks 0, 1 and 4 have headroom for selection (policy alone below 80\% in a reference run fixed beforehand).

\paragraph{Registered tests.}
Each registered test used new policy seeds and re-ran every arm.
The primary statement holds iff the 95\% cluster-bootstrap interval of ours minus native excludes zero and the one-sided task-level sign-flip gives $p<0.05$; the four-adapter tests add that the point estimate is positive for each of the four adapters (training seeds 1701, 3407, 5501, 7919).
The ten-task confirmation applies Holm's procedure over its primary test and a non-inferiority test with margin $\cfNI{}$.
Table~\ref{tab:s7-reg} lists the registered tests and their contrasts.

\begin{table}[t]
  \centering\scriptsize\setlength\tabcolsep{3pt}
  \caption{Registered closed-loop tests (ten tasks, 50 initial states). (a)~Success per arm; ours is the mean over the adapters of the test: two training seeds (ten-task confirmation, higher headroom) or four (the other rows; dose-response at strength $1.0$, adapters in rotation). Gain: ours minus native (95\% cluster-bootstrap interval); $p$: task-level sign-flip. Higher headroom, four-adapter and dose-response tests use the higher-headroom candidate setting. (b)~Further contrasts; nine-task $p$: sign-flip over the tasks other than task~1; adapter range: smallest to largest gain over native among the adapters of the test.}
  \label{tab:s7-reg}
  \par\vspace{4pt}{\footnotesize(a) Success per arm}\\[1pt]
  \begin{tabular}{@{} l c c c c c c c c @{}}
    \toprule
    Test & Policy seeds & Episodes & Policy alone & Native & Ours & Oracle & Gain & $p$ \\
    \midrule
    Ten-task confirmation & 10--13 & \cfEpisodes{} & \cfRateBase{} & \cfRateNat{} & \cfRateSched{} & \cfRateOracle{} & \cfNat\cfNatCI{} & \cfTaskP{} \\
    Higher headroom & 20--22 & \fuEpisodes{} & \fuRateBase{} & \fuRateNat{} & \fuRateSched{} & \fuRateOracle{} & \fuNat\fuNatCI{} & \fuP{} \\
    Four adapters, first & 23--24 & \fqEpisodes{} & \fqRateBase{} & \fqRateNat{} & \fqRateSched{} & \fqRateOracle{} & \fqNat\fqNatCI{} & \fqP{} \\
    Four adapters, replication & 27--31 & \rzEpisodes{} & \rzRateBase{} & \rzRateNat{} & \rzRateSched{} & \rzRateOracle{} & \rzNat\rzNatCI{} & \rzP{} \\
    Dose-response, strength $1.0$ & 32--35 & \dzContexts{} & \dzRateBase{} & \dzRateNat{} & \dzRateFull{} & \dzRateOracle{} & \dzEnd\dzEndCI{} & \dzEndP{} \\
    \bottomrule
  \end{tabular}\\[4pt]
  {\footnotesize(b) Further contrasts}\\[1pt]
  \begin{tabular}{@{} l c c c c c c @{}}
    \toprule
    Test & Task~1 & Without task~1 & Nine-task $p$ & Ours $-$ policy alone & Native $-$ policy alone & Adapter range \\
    \midrule
    Ten-task confirmation & \cfOne\cfOneCI{} & \cfNoOne\cfNoOneCI{} & \cfNoOneP{} & \cfBase\cfBaseCI{} & \cfNatBase\cfNatBaseCI{} & \cfSeedA{} to \cfSeedB{} \\
    Higher headroom & \fuOne\fuOneCI{} & \fuNoOne\fuNoOneCI{} & \fuNoOneP{} & \fuBase\fuBaseCI{} & \fuNatBase\fuNatBaseCI{} & \fuSeedA{} to \fuSeedB{} \\
    Four adapters, first & \fqOne\fqOneCI{} & \fqNoOne\fqNoOneCI{} & \fqNoOneP{} & \fqBase\fqBaseCI{} & \fqNatBase\fqNatBaseCI{} & \fqAdMin{} to \fqAdMax{} \\
    Four adapters, replication & \rzOne\rzOneCI{} & \rzTwoDiff\rzTwoCI{} & \rzTwoP{} & \rzBase\rzBaseCI{} & \rzNatBase\rzNatBaseCI{} & \rzAdMin{} to \rzAdMax{} \\
    Dose-response, strength $1.0$ & --- & \dzEndTwoDiff\dzEndTwoCI{} & --- & --- & --- & \dzEndAdA{} to \dzEndAdD{} \\
    \bottomrule
  \end{tabular}
\end{table}

In the ten-task confirmation the Holm-adjusted one-sided $p$ is $\cfP{}$ and the non-inferiority test holds ($p\cfNIP{}$); ours gains on $\cfTasksPos{}$ of ten tasks, and on the three headroom tasks ours exceeds native by $\cfHead{}\cfHeadCI{}$.
The higher-headroom setting was chosen in a WAM-free pilot of four sampler settings of the same policy, by the oracle's advantage over the policy alone: an initial-noise scale of $1.5$ gave $\fuPilotHead{}$ against $\fuPilotDefHead{}$ at the default. There the oracle exceeds the policy alone by $\fuOracleBase{}$, and ours gains on $\fuTasksPos{}$ of ten tasks and closes $\fuShare\%$ of the gap between native and the oracle.
In the first four-adapter test every adapter is positive (task-level sign-flip $p=\fqP{}$); the replication on five new policy seeds meets all three registered criteria.
Pooled per task over the four registered batches that share tasks, initial states and value head (weighted by episodes), the gain over native is $\pqMean\pqCI$ with $\pqPos{}$ of ten tasks positive (sign-flip $p=\pqP{}$), and $\pqNoOne{}$ without task~1 ($p=\pqNoOneP{}$).

\paragraph{Earlier batches.}
Table~\ref{tab:s7-earlier} lists the batches that preceded the registered tests; the main text cites them as earlier seeds. All used their own policy seeds and the same value head and protocol.
A second value head, trained on frozen DINOv2 features, met the registered validity, value-ranking and agreement criteria; all closed-loop tests use the first head so that the batches share one scorer.

\begin{table}[t]
  \centering\scriptsize\setlength\tabcolsep{3pt}
  \caption{Earlier closed-loop batches (success, repaired minus native with 95\% cluster-bootstrap intervals; $p$ where a test was registered or computed). Rows 1--2: always-on repair; rows 3--6: scheduled repair. Seeds: policy seeds.}
  \label{tab:s7-earlier}
  \begin{tabular}{@{} l c c c c @{}}
    \toprule
    Batch & Native & Repaired & Difference & $p$ \\
    \midrule
    Tasks 0, 1, 4, training seed 1701 (seeds 0--1, 300 episodes) & \tsRateNative{} & \tsRateRepaired{} & \tsRepNat\tsRepNatCI{} & \tsRepNatP{} \\
    Seven other tasks, training seed 1701 (700 episodes) & \tsXNativeRate{} & \tsXLoraRate{} & \tsXLoraNativeDiff\tsXLoraNativeCI{} & \tsXLoraNativeP{} \\
    \midrule
    Tasks 0, 1, 4, four adapters (seeds 0--1, 300 episodes) & \eFourNative{} & \eFourRecipeRate{} & \clXAuditAWValue\clXAuditAWValueCI{} & --- \\
    Replication, tasks 0, 1, 4 (seeds 2--3, 300 episodes) & \rpNatThis{} & \rpSchedThis{} & \rpDiff\rpDiffCI{} & \rpP{} \\
    Seven other tasks, four adapters (seeds 0--1, 700 episodes) & \gSevNative{} & \gSevSched{} & \gSevGain\gSevGainCI{} & \gSevP{} \\
    Registered ten-task test, seeds 4--7 (\ceEpisodes{} episodes) & \ceRateNat{} & \ceRateSched{} & \ceNat\ceNatCI{} & --- \\
    \bottomrule
  \end{tabular}
\end{table}

\paragraph{Ranking of the candidates against their simulator outcomes.}
At query 0 of every episode the four candidates are common to all arms; the oracle scores the real final frame of each with the frozen value head, and the WAM arms score the imagined final frame of the same candidate.
For each (task, initial state, policy seed) we compute the Spearman correlation between the two score vectors and the top-1 agreement (the WAM's best candidate is the simulator's best; chance $0.25$). Only query 0 is read, because later queries follow different trajectories in different arms.
Native's scores are almost uncorrelated with the simulator's (Table~\ref{tab:s7-rank}); ours raises the rank correlation by $\dscRhoDiff\dscRhoCI{}$ and $\dsdRhoDiff\dsdRhoCI{}$ in the two batches, positive in all ten tasks in both (sign-flip $p=0.001$), and top-1 agreement by $\dscTopDiff{}$ and $\dsdTopDiff{}$ (positive in $\dscTopTasks{}$ and $\dsdTopTasks{}$ tasks).
The comparison is within one state and one set of candidates, so a shift that moves all four imagined frames alike cannot produce it, and both measures are invariant to monotone recalibration of the value head.
In both four-adapter tests the readout was pre-specified in the registration before any episode ran: ours raises the rank correlation from $\fqRhoNat{}$ to $\fqRhoSch{}$ ($\fqRhoDiff\fqRhoCI{}$) and, in the replication, from $\rzRhoNat{}$ to $\rzRhoSch{}$ ($\rzRhoDiff\rzRhoCI{}$, positive in all $\rzRhoTasks{}$ tasks, sign-flip $p=\rzRhoP{}$).

\begin{table}[t]
  \centering\scriptsize\setlength\tabcolsep{3pt}
  \caption{Agreement of the WAM's candidate scores with the simulator outcomes of the same candidates at query 0 (ours is the mean over its adapters). Spearman difference with a 95\% cluster-bootstrap interval. $^{\ast}$Pre-specified secondary readout of the registered test; the rows above are descriptive.}
  \label{tab:s7-rank}
  \begin{tabular}{@{} l c c c c c @{}}
    \toprule
    Batch & Spearman, native & Spearman, ours & Difference & Top-1, native & Top-1, ours \\
    \midrule
    Higher headroom (\dscN{} contexts) & \dscRhoNat{} & \dscRhoSch{} & \dscRhoDiff\dscRhoCI{} & \dscTopNat{} & \dscTopSch{} \\
    Default setting (\dsdN{} contexts) & \dsdRhoNat{} & \dsdRhoSch{} & \dsdRhoDiff\dsdRhoCI{} & \dsdTopNat{} & \dsdTopSch{} \\
    Dose-response, strength $1.0$ & \dzRhoNat{} & \dzRhoFull{} & \dzRhoFullDiff\dzRhoFullCI{} & --- & --- \\
    \midrule
    Four adapters, first$^{\ast}$ (\fqEpisodes{} contexts) & \fqRhoNat{} & \fqRhoSch{} & \fqRhoDiff\fqRhoCI{} & \fqTopNat{} & \fqTopSch{} \\
    Four adapters, replication$^{\ast}$ (\rzEpisodes{} contexts) & \rzRhoNat{} & \rzRhoSch{} & \rzRhoDiff\rzRhoCI{} & \rzTopNat{} & \rzTopSch{} \\
    \bottomrule
  \end{tabular}
\end{table}

\paragraph{Dose-response on the repair strength.}
The same frozen adapters ran at strength $0$ (native), $0.5$ and $1.0$; the strength rescales the adapter's contribution and is not a statement about training steps.
Ten tasks, 50 initial states and policy seeds 32--35 give $\dzContexts{}$ contexts, each run with one of the four adapters in rotation.
With $g_{0.5}$ and $g_{1.0}$ the per-context success differences to native, the slope is $(0.5\,g_{0.5}+1.0\,g_{1.0})/1.25$; the dose route is supported iff the pooled-slope interval excludes 0, the task-level sign-flip over the ten per-task slopes gives $p<0.05$, and $\sum g_{1.0}>\sum g_{0.5}>0$.
Success is $\dzRateNat{}$ for native, $\dzRateHalf{}$ at strength $0.5$ and $\dzRateFull{}$ at $1.0$; the gain over native is $\dzGainHalf\dzGainHalfCI{}$ at $0.5$ and $\dzGainFull\dzGainFullCI{}$ at $1.0$.
The pooled slope is $\dzSlope\dzSlopeCI{}$, the sign-flip gives $p=\dzSignP{}$ with $\dzTasksPos{}$ of ten task slopes positive, and the sums are $\dzSumFull{}$ and $\dzSumHalf{}$, so the registered dose route is supported; without task~1 the slope is $\dzTwoSlope\dzTwoSlopeCI{}$ ($p=\dzTwoSignP{}$).

\paragraph{Overlap with the repair training states.}
Twenty of the 50 official initial states of each task are also initial states of contexts used to train the adapters (Sec.~S1).
Splitting the dose-response contexts by this overlap: on the 20 overlapping states ($\ovPoolN{}$ contexts) ours at strength $1.0$ gains $\ovPoolGain\ovPoolGainCI{}$ over native and $\ovPoolBase\ovPoolBaseCI{}$ over the policy alone; on the other 30 states ($\ovOtherN{}$ contexts) it gains $\ovOtherGain\ovOtherGainCI{}$ and $\ovOtherBase\ovOtherBaseCI{}$.
The gain is not carried by the states the adapters saw in training.

\paragraph{Selection with up to sixteen candidates on a fixed trajectory.}
To read which candidate a WAM ranks first apart from how the choice changes later states, the executed chunk is always the policy's own sample (candidate~0).
At every query of the policy-alone run (policy seed 14; ten tasks; initial states $0,8,\dots,48$; $\fsTraj{}$ trajectories, $\fsQueries{}$ queries) the policy proposes 16 candidates, each scored by the oracle and by every WAM with shared noise.
For the first $K\in\{2,4,8,16\}$ candidates, a scorer's regret is the oracle score of the best candidate minus that of the candidate it ranks first, and the shrink is $1-R_{\text{ours}}(K)/R_{\text{native}}(K)$, averaged over the four training seeds.
Ours shrinks native's regret by $\fsShrinkEight{}$ at $K=8$ and $\fsShrinkSixteen{}$ at $K=16$, with $\fsSeedsPosEight{}$ and $\fsSeedsPosSixteen{}$ of four training seeds positive.
Native's regret grows with $K$, from $\fsRegNatFour{}$ at $K=4$ to $\fsRegNatSixteen{}$ at $K=16$.
Scores are value-head units, not success rates.
\paragraph{Native A2World as the WAM.}
A2World has high identity natively (Sec.~S8), so if linking each action to its own consequence is what best-of-four selection needs, native A2World should select better than native Prophet, with the repaired Prophet between the two.
We ran two arms on the contexts of the dose-response batch ($\awcContexts{}$ contexts, $\awcClusters{}$ task--state clusters): the policy alone, to check the contexts, and native A2World; native Prophet, the repaired Prophet at full strength and the oracle are the stored arms of that batch.
$\awcPairShare{}$ of the contexts reproduce the stored policy-alone outcome, so the comparison with native Prophet is an unpaired cluster contrast and that with the policy alone is paired.
Success is $\awcRateA{}$ for native A2World, $\awcRateBase{}$ for the policy alone, $\awcRateNat{}$ for native Prophet, $\awcRateRep{}$ for the repaired Prophet and $\awcRateOracle{}$ for the oracle.
The repaired Prophet recovers $\awcShare{}$ of the difference between native A2World and native Prophet (interval $\awcShareCI{}$); native A2World exceeds the policy alone by $\awcTwo{}$.
At the first query the Spearman correlation between imagined scores and simulator outcomes is $\awcRhoA{}$ for native A2World, $\awcRhoNat{}$ for native Prophet and $\awcRhoRep{}$ for the repaired Prophet, and the simulator's best candidate is picked in $\awcTopA{}$, $\awcTopNat{}$ and $\awcTopRep{}$ of decisions (chance $0.25$; descriptive).
A2World differs from Prophet in architecture, training and sampler, so this is not a controlled test of action identity alone.

\section{Transfer to other tasks, models and actions}
\label{supp:n8}

This section supports Table~\ref{tab:transfer} and Sec.~\ref{sec:exp-transfer}: four further LIBERO suites, further WAMs, policy-proposed actions and a held-out action template. Intervals are 95\% context-clustered bootstrap intervals unless stated.

\paragraph{Other tasks and scenes: four further LIBERO suites.}
The adapter was trained on LIBERO-10 only; a one-time registered readout used $\suN{}$ contexts of LIBERO-Spatial, LIBERO-Object, LIBERO-Goal and a LIBERO-90 subsample.
Native, always-on and ours (the same weights under the step gate) were run with four training seeds and evaluation seeds 3407 and 3413; intervals resample (suite, task, demonstration) clusters.
The registered tests were the margin gain over native (lower bound above zero, positive in every suite and training seed) and ours keeping the always-on gain (lower bound of ours minus always-on above $-0.02$), with Holm over these two, and the pooled FVD ratio against the $1.10$ bound.
Ours raises the identity margin by $\suMargin{}\suMarginCI{}$ over native, positive on $\suPos{}$ of $\suN{}$ contexts, in all four suites and for all training seeds; PairAcc rises by $\suPair{}\suPairCI{}$. Ours minus always-on is $\suKeep{}$, and both registered tests reject after Holm correction.
The step gate lowers the pooled FVD ratio to native from $\suFvdFull{}$ (always-on) to $\suFvd{}\suFvdCI{}$, on scenes never seen in training, which is the price of a margin gain in every suite from an adapter trained on LIBERO-10 only; the $1.10$ bound is the LIBERO-10 registration. Per-suite values are in Table~\ref{tab:transfer}.

\paragraph{Other models: Cosmos Policy and Wan.}
The same-state diagnosis of Sec.~\ref{sec:diagnosis} was applied to two further action-conditioned video models (frozen evaluator, frame after 16 actions, three noise seeds).
Cosmos Policy (2B parameters, action as latent frames; 39 development contexts) shows the deficit of Prophet: direction cosine $\cdCos{}$, PairAcc $\cdPair{}\cdPairCI{}$, identity margin $\cdMargin{}$, against a ground-truth ceiling of $1.0$.
A Wan2.2-5B WAM released for LIBERO-Goal ($\wdN{}$ development contexts) has high identity: direction cosine $\wdCos{}$, PairAcc $\wdPair{}\wdPairCI{}$, margin $\wdMargin{}$, SNR $\wdSnr{}$; it was trained on policy rollouts, so its reading also reflects a distribution shift.
By the registered room-to-repair rule (direction cosine below $0.4$ and PairAcc interval lower bound below $0.75$) Cosmos Policy has room and the Wan model does not, so only Cosmos Policy was repaired.

The Cosmos Policy adapter is a rank-5 LoRA on the query and value projections of the self-attention in all 28 blocks ($1{,}146{,}880$ parameters, $0.059\%$ of the network), trained for 1024 updates on the same 781-context pool with the Prophet loss and the sampler's own states; there is no step schedule because the model produces its single future frame in one sequence.
The registered held-out test ($\ctN{}$ contexts, three noise seeds) required the pooled margin-gain interval above zero, both training seeds positive and the upper bound of the global-error ratio below $1.10$. All three hold: the gain is $\ctMargin{}\ctMarginCI{}$ ($\ctSeedA{}$ and $\ctSeedB{}$ per seed), and the global-error ratio is $\ctGuard{}$ (upper bound $\ctGuardHi{}$).
The margin rises from $\ctMarginNat{}$ to $\ctMarginAd{}$ and PairAcc by $\ctPair{}\ctPairCI{}$; the contexts are held out for the adapter only, since Cosmos Policy was trained on all LIBERO demonstrations.

\paragraph{UWM and A2World.}
UWM's action information reaches the future tokens only through attention, so its adapter covers the action encoder and the query, key and value projections of blocks 0--7 at the action-token positions, and the objective adds a branch-contrast term (weight $0.5$); the configuration (1024 updates, checkpoint at 768) was selected on 40 pool-validation contexts only.
The registered confirmation had the margin as primary endpoint, against native and against a matched control that fits the ground truth with the same module, data and budget. It is confirmed on both held-out sets (Table~\ref{tab:s8-models}); the native UWM margin is $\dgUMargin{}$ in latent space.

A2World already has high identity, so the question was whether the unchanged recipe does harm; the registered rule is non-inferiority with margins of $1.75$ times the interval half-widths of Prophet's development readout, for the adapter and its matched control alike.
Under the always-on recipe both held-out sets are confirmed non-inferior (Table~\ref{tab:s8-models}).
The schedule transfer (236 contexts, two training seeds) disables the last three of six denoiser calls by $\sigma\ge5$, selected for this model by the development rule used for Prophet.
The scheduled adapter gains $\awXMargin{}$ in margin at an FVD ratio of $\awXFVD{}$; A2World already has the highest identity (Sec.~\ref{sec:diag-spectrum}), so the adapter has the least to add there: PairAcc, object and contact errors stay close to native (sign-adjusted gains $\awXAuditAUValue{}\awXAuditAUValueCI{}$ and $\awXAuditAQValue{}\awXAuditAQValueCI{}$, pooled).

\begin{table}[htbp]
  \centering\scriptsize\setlength\tabcolsep{3pt}
  \caption{Further models and actions: gains over native (95\% intervals over contexts) and, where a matched control was trained, the margin gain over that control. UWM and A2World held-out rows are the always-on recipe with two (A2World) or four (UWM) training seeds; the A2World scheduled row pools fresh and sealed; policy candidates are read under the step gate.}
  \label{tab:s8-models}
  \begin{tabular}{@{} l r c c c l @{}}
    \toprule
    Setting & $n$ & Margin gain & Margin, minus control & PairAcc gain & Decision \\
    \midrule
    Cosmos Policy, repaired & \ctN{} & \ctMargin\ctMarginCI{} & --- & \ctPair\ctPairCI{} & registered test met \\
    UWM, fresh & \cpFreshN{} & \cpFreshNat\cpFreshNatCI{} & \cpFreshCtrl\cpFreshCtrlCI{} & \cpFreshPairAcc\cpFreshPairAccCI{} & confirmed \\
    UWM, sealed & \cpSealedN{} & \cpSealedNat\cpSealedNatCI{} & \cpSealedCtrl\cpSealedCtrlCI{} & \cpSealedPairAcc\cpSealedPairAccCI{} & confirmed \\
    A2World, fresh & 117 & \awFreshMargin\awFreshMarginCI{} & \awFreshMarginCtrl\awFreshMarginCtrlCI{} & \awFreshPairAcc\awFreshPairAccCI{} & \awFreshDecision{} \\
    A2World, sealed & 119 & \awSealedMargin\awSealedMarginCI{} & \awSealedMarginCtrl\awSealedMarginCtrlCI{} & \awSealedPairAcc\awSealedPairAccCI{} & \awSealedDecision{} \\
    A2World, scheduled & 236 & \awXAuditASValue\awXAuditASValueCI{} & --- & \awXAuditAVValue\awXAuditAVValueCI{} & --- \\
    Policy candidates, fresh & \pcN{} & \pcMarg\pcMargCI{} & --- & \pcPair\pcPairCI{} & both guards held \\
    \bottomrule
  \end{tabular}
\end{table}

\paragraph{Other actions: policy-proposed chunks and the closed loop.}
The template-trained adapter under its step gate (training seeds 1701 and 3407, no retraining) was applied to the four chunks proposed by the closed-loop policy, replayed in the simulator, on $\pcN{}$ contexts not used for any selection (Table~\ref{tab:s8-models}).
Scheduled minus native gives identity margin $\pcMarg{}\pcMargCI{}$, PairAcc $\pcPair{}\pcPairCI{}$, LPIPS $\pcLpips{}$ and FVD ratio $\pcFvd{}\pcFvdCI{}$; both realism guards held, and the margin's noise floor is $\pcFloor{}$.
The registered ten-task closed-loop test (policy seeds 4--6, $\ccEpisodes{}$ episodes per arm, two training seeds pooled) reads \emph{improves} if the lower bound is above zero and $p<0.05$, \emph{non-inferior} if it is above $-\ccNI{}$, and \emph{not shown} otherwise.
Ours minus native is $\ccNat{}\ccNatCI{}$ (non-inferior), ours minus the policy alone $\ccBase{}\ccBaseCI{}$, and oracle minus native $\ccOracle{}$.
Pooling policy seeds 4--7 (seed 7 fixed in advance) gives $\ceEpisodes{}$ episodes per arm with success $\ceRateBase{}$ (policy alone), $\ceRateNat{}$ (native), $\ceRateSched{}$ (ours) and $\ceRateOracle{}$ (oracle); ours minus native is $\ceNat{}\ceNatCI{}$ and ours minus the policy alone $\ceBase{}\ceBaseCI{}$, and the two training seeds point the same way (descriptive).

\paragraph{Other actions: a held-out template.}
Four leave-one-template-out checkpoints (each trained on three of the four counterfactual templates, training seed 1701) were read on the 117 fresh and 119 sealed contexts with evaluation seeds 3407 and 3413.
The registered primary endpoint averages the margin gain over native on the held-out template across contexts and the four folds, with a gated analog; the pooled gain is $\htXGain{}$.
Table~\ref{tab:s8-folds} splits the gain on the pairs involving the held-out branch into an own part (fold model minus a hybrid whose held-out branch comes from the native rollout) and a rest (hybrid minus native).
The pooled gain is carried by the own-branch change, that is, by the model predicting the held-out action's own consequence.

\begin{table}[htbp]
  \centering\scriptsize\setlength\tabcolsep{4pt}
  \caption{Held-out template, scheduled deployment, pooled over fresh and sealed: identity-margin gain over native with its own and rest parts (95\% context-bootstrap intervals). The own share is own divided by gain and is undefined when the interval of the gain contains zero.}
  \label{tab:s8-folds}
  \begin{tabular}{@{} l c c c c @{}}
    \toprule
    Held-out template & Gain & Own branch & Rest & Own share \\
    \midrule
    \br{gripper\_opposite} & \ptci{\foldGripGain}{\foldGripGainCI} & \ptci{\foldGripOwn}{\foldGripOwnCI} & \ptci{\foldGripRest}{\foldGripRestCI} & undefined \\
    \br{hold\_pose} & \ptci{\foldHoldGain}{\foldHoldGainCI} & \ptci{\foldHoldOwn}{\foldHoldOwnCI} & \ptci{\foldHoldRest}{\foldHoldRestCI} & \foldHoldShare{} \\
    \br{translation\_axis\_flip} & \ptci{\foldFlipGain}{\foldFlipGainCI} & \ptci{\foldFlipOwn}{\foldFlipOwnCI} & \ptci{\foldFlipRest}{\foldFlipRestCI} & \foldFlipShare{} \\
    \br{translation\_half} & \ptci{\foldHalfGain}{\foldHalfGainCI} & \ptci{\foldHalfOwn}{\foldHalfOwnCI} & \ptci{\foldHalfRest}{\foldHalfRestCI} & \foldHalfShare{} \\
    \midrule
    Pooled over four folds & \ptci{\foldPoolGain}{\foldPoolGainCI} & \ptci{\foldPoolOwn}{\foldPoolOwnCI} & \ptci{\foldPoolRest}{\foldPoolRestCI} & \foldPoolShare{} \\
    \bottomrule
  \end{tabular}
\end{table}

\section{Ablations and baselines}
\label{supp:n9}

This section supports Table~\ref{tab:ablations}, the baseline rows of Table~\ref{tab:results} and Secs.~\ref{sec:exp-ablation} and~\ref{sec:exp-baselines}: recipe ablations, baseline settings and the closed-loop test of the backbone adapters. Gains are paired inside each evaluation process; intervals are 95\% context-clustered bootstrap intervals unless stated.

\paragraph{Recipe ablations on held-out contexts.}
Three ablations were read once on the $\eFvdNContexts{}$ held-out contexts (fresh $\aExNFresh{}$, sealed $\aExNSealed{}$) with four training seeds (1701, 3407, 5501, 7919), update 1024 and evaluation seeds 3407 and 3413: (i) GT-noised states instead of sampler states; (ii) no direction term; (iii) GT-fit only, the compact adapter trained with the ground-truth flow-matching loss and the anchor term only (same rank, rate, pool and updates).
Each runs under the step schedule with the scheduled recipe and native re-run in the same process, so every contrast is paired; the read was registered before any result existed.
The primary endpoints are each ablation minus the scheduled recipe on the identity margin and on PairAcc; the registered reading is ``lower than the recipe'' if the upper end of the interval is below zero and ``not shown'' otherwise, also at the demonstration (60 clusters) and task level (10 clusters).

\begin{table}[htbp]
  \centering\scriptsize\setlength\tabcolsep{4pt}
  \caption{Recipe ablations; positive is better. (a)~Held-out read (236 contexts, step schedule): gain over native and the primary contrast with the scheduled recipe with its context-level and task-level intervals. (b)~Development readouts at 1024 updates (39 contexts, four seeds; 95\% intervals over contexts): gains over native and, for the ablation arms, arm minus the frozen recipe on the same seeds.}
  \label{tab:s9-held}
  \par\vspace{4pt}{\footnotesize(a) Held-out contexts}\\[1pt]
  \begin{tabular}{@{} l l c c c @{}}
    \toprule
    Variant & Metric & Gain over native & Minus recipe [context CI] & Task-level CI \\
    \midrule
    Recipe (scheduled) & margin & \ptci{\aExRecVsNatMarg}{\aExRecVsNatMargCI} & --- & --- \\
     & PairAcc & \ptci{\aExRecVsNatPair}{\aExRecVsNatPairCI} & --- & --- \\
    GT-noised states & margin & \ptci{\aExGtnVsNatMarg}{\aExGtnVsNatMargCI} & \ptci{\aExGtnVsRecMarg}{\aExGtnVsRecMargCI} & $\aExGtnVsRecMargTaskCI$ \\
     & PairAcc & \ptci{\aExGtnVsNatPair}{\aExGtnVsNatPairCI} & \ptci{\aExGtnVsRecPair}{\aExGtnVsRecPairCI} & $\aExGtnVsRecPairTaskCI$ \\
    No direction term & margin & \ptci{\aExNodirVsNatMarg}{\aExNodirVsNatMargCI} & \ptci{\aExNodirVsRecMarg}{\aExNodirVsRecMargCI} & $\aExNodirVsRecMargTaskCI$ \\
     & PairAcc & \ptci{\aExNodirVsNatPair}{\aExNodirVsNatPairCI} & \ptci{\aExNodirVsRecPair}{\aExNodirVsRecPairCI} & $\aExNodirVsRecPairTaskCI$ \\
    GT-fit loss only & margin & \ptci{\aExGtfitVsNatMarg}{\aExGtfitVsNatMargCI} & \ptci{\aExGtfitVsRecMarg}{\aExGtfitVsRecMargCI} & $\aExGtfitVsRecMargTaskCI$ \\
     & PairAcc & \ptci{\aExGtfitVsNatPair}{\aExGtfitVsNatPairCI} & \ptci{\aExGtfitVsRecPair}{\aExGtfitVsRecPairCI} & $\aExGtfitVsRecPairTaskCI$ \\
    \bottomrule
  \end{tabular}\\[4pt]
  {\footnotesize(b) Development contexts}\\[1pt]
  \begin{tabular}{@{} l c c c c @{}}
    \toprule
    Arm & PairAcc gain & Margin gain & PairAcc, minus recipe & Margin, minus recipe \\
    \midrule
    Recipe & \devPairAccGain\devPairAccGainCI{} & \devMarginGain\devMarginGainCI{} & --- & --- \\
    Consequence weight 3 & \devConsThreePairAccGain\devConsThreePairAccGainCI{} & \devConsThreeMarginGain\devConsThreeMarginGainCI{} & --- & --- \\
    \midrule
    GT-noised training states & \abGtNPairAcc\abGtNPairAccCI{} & \abGtNMargin\abGtNMarginCI{} & \abGtNDPairAcc\abGtNDPairAccCI{} & \abGtNDMargin\abGtNDMarginCI{} \\
    No direction term & \abNoDirPairAcc\abNoDirPairAccCI{} & \abNoDirMargin\abNoDirMarginCI{} & \abNoDirDPairAcc\abNoDirDPairAccCI{} & \abNoDirDMargin\abNoDirDMarginCI{} \\
    Rank 8 & \abRankEightPairAcc\abRankEightPairAccCI{} & \abRankEightMargin\abRankEightMarginCI{} & \abRankEightDPairAcc\abRankEightDPairAccCI{} & \abRankEightDMargin\abRankEightDMarginCI{} \\
    Rank 4 & \abRankFourPairAcc\abRankFourPairAccCI{} & \abRankFourMargin\abRankFourMarginCI{} & \abRankFourDPairAcc\abRankFourDPairAccCI{} & \abRankFourDMargin\abRankFourDMarginCI{} \\
    192 training contexts & \abDataSmallPairAcc\abDataSmallPairAccCI{} & \abDataSmallMargin\abDataSmallMarginCI{} & \abDataSmallDPairAcc\abDataSmallDPairAccCI{} & \abDataSmallDMargin\abDataSmallDMarginCI{} \\
    389 training contexts & \abDataMidPairAcc\abDataMidPairAccCI{} & \abDataMidMargin\abDataMidMarginCI{} & \abDataMidDPairAcc\abDataMidDPairAccCI{} & \abDataMidDMargin\abDataMidDMarginCI{} \\
    \bottomrule
  \end{tabular}
\end{table}

All six primary contrasts are negative with intervals that exclude zero at the context, demonstration and task level, every leave-one-task-out estimate is negative, and all six read ``\aEGtfitMargRead{}''; sign and reading are the same on the fresh and sealed sets.
The GT-noised and no-direction arms lose about the same margin ($\aExGtnVsRecMarg{}$ and $\aExNodirVsRecMarg{}$) and keep about half of the recipe's gain; the GT-fit arm loses the most ($\aExGtfitVsRecMarg{}$) and is close to native. The GT-fit arm is not a single-factor change from the recipe.

\paragraph{Development-split ablations.}
Every development arm was trained with four seeds and read on the 39 development contexts of demonstration 35 with two evaluation seeds (Table~\ref{tab:s9-held}b, 1024 updates).
Training on GT-noised states lowers PairAcc, margin and both error reductions with intervals that exclude zero. Removing the direction term lowers PairAcc and margin but leaves the object and contact error reductions unchanged (contrasts $\abNoDirDObj{}$ and $\abNoDirDCont{}$, intervals including zero): the direction term carries identity, and the object and consequence terms carry the error reductions. Rank 8 and the two smaller training sets stay within about $0.01$ of the recipe; rank 4 is lower.
The weight-1 recipe was kept by the registered route rule.

\paragraph{Baselines: settings.}
The two full-rank arms update all weights of the two action embedders ($71{,}720{,}960$ parameters, $57.6$ times the $1{,}245{,}600$ of the recipe's adapter); the learning rate was chosen on 40 pool-validation contexts by the largest mean margin gain among admissible rates (global-error change at most $0.005$, denoising ratio at most $1.05$), giving $10^{-5}$ for ground-truth fitting and \ftOursLr{} for the recipe's objective.
\emph{Action gain} multiplies the translation and rotation entries of the action chunk by $g$. \emph{Contrastive guidance} moves every predicted clean latent of the five branches away from their mean, $x_a \leftarrow x_a + \lambda (x_a - \bar{x})$. \emph{FreeAction} replaces the missing unconditional condition by the negated action in the initial three of five sampler steps with weight $w=\lambda\|a\|_2$; the full variant also truncates the initial noise to a radius that grows with $\|a\|_2$. Classifier-free guidance is a no-op on Prophet, which has no unconditional path.
Each setting was chosen on 40 pool-validation contexts, disjoint from demonstration 35, by the same admissibility rule (global endpoint error change at most $0.005$, largest mean margin gain) over grids of $g$ and $\lambda$; the selected settings are $g=1.5$ and $\lambda=0.25$, each evaluated once on the development contexts.
None raises PairAcc or margin significantly there and none meets more than two of the seven acceptance conditions; the fresh-split contrasts of all baselines are in Table~\ref{tab:heldout}.

\begin{table}[t]
\centering
\caption{\textbf{On unseen states, ours has the largest margin gain within the FVD bound.} Fresh split, $\hbN{}$ contexts. Gain: identity margin over native (ours scheduled, baselines always-on, two training seeds); FVD ratio to native (guidance baselines decode no video). Bold: highest gain with FVD ratio $\le1.10$. Baselines are adaptations to our setting.}
\label{tab:heldout}
\footnotesize
\setlength{\tabcolsep}{3pt}
\renewcommand{\arraystretch}{1.03}
\begin{tabular}{@{}lccc@{}}
\toprule
Method & Gain & Ours $-$ method [95\% CI] & FVD \\
\midrule
Ours (scheduled) & $\mathbf{\hbOpieGain{}}$ & -- & $\hbOpieFvd{}$ \\
WorldSync-style & $\hbWsGain{}$ & $\hbWsDiff{}\,\hbWsCI{}$ & $\hbWsFvd{}$ \\
Full-rank, GT fit & $\hbFrgGain{}$ & $\hbFrgDiff{}\,\hbFrgCI{}$ & $\hbFrgFvd{}$ \\
Full-rank, our loss & $\hbFroGain{}$ & $\hbFroDiff{}\,\hbFroCI{}$ & $\hbFroFvd{}$ \\
Backbone, GT fit & $\hbBOneGain{}$ & $\hbBOneDiff{}\,\hbBOneCI{}$ & $\hbBOneFvd{}$ \\
Backbone, our loss & $\hbBTwoGain{}$ & $\hbBTwoDiff{}\,\hbBTwoCI{}$ & $\hbBTwoFvd{}$ \\
CureWM-style & $\hbCwGain{}$ & $\hbCwDiff{}\,\hbCwCI{}$ & $\hbCwFvd{}$ \\
Action gain $\times1.5$ & $\hbGainGain{}$ & $\hbGainDiff{}\,\hbGainCI{}$ & $\hbGainFvd{}$ \\
Contrastive guidance & $\hbContrGain{}$ & $\hbContrDiff{}\,\hbContrCI{}$ & -- \\
FreeAction & $\hbFaGain{}$ & $\hbFaDiff{}\,\hbFaCI{}$ & -- \\
\bottomrule
\end{tabular}
\end{table}

\paragraph{WorldSync-style baseline.}
It trains the flow-matching loss plus a pairwise intervention-effect loss (squared error between $v_a-v_b$ and $x_{0,b}-x_{0,a}$ over the ten branch pairs) on the full-rank action embedders at ground-truth-noised states; settings not in the published description are set as in the full-rank arms.
Two deviations remain: the backbone is frozen, because full fine-tuning does not fit a 40 GB card, and the action-forcing expert is absent, because its weights are not published; data and budget are those of the recipe. We therefore call it a WorldSync-style baseline and not a reproduction.
On the development contexts it lowers the margin ($\wsMargin\wsMarginCI$) and does not raise PairAcc ($\wsPairAcc\wsPairAccCI$) at the selected rate $10^{-6}$; two smaller rates added on pool validation score at most zero as well (best $\wsxScoreMax{}$).

\paragraph{Backbone adapters.}
The two backbone arms use rank-16 adapters on 280 linear projections across the 28 DiT blocks of Prophet ($\hbParamBig{}$ parameters against $\hbParamSmall{}$ for the recipe adapter, $\hbParamRatio\times$ more).
Backbone (GT fit) is trained with ground-truth flow-matching targets and Backbone (our loss) with the recipe objective on on-policy states, both for 1024 updates; the learning rate $3\cdot10^{-4}$ was selected on pool validation for both. Deviations from full-backbone fine-tuning with a longer budget and an action-forcing expert: rank-16 LoRA, 1024 updates, and no action-forcing expert.
On the fresh split Backbone (our loss) reaches a margin gain of $\hbBTwoGain{}$ against $\hbOpieGain{}$ for ours, at an FVD ratio of $\hbBTwoFvd{}$ against $\hbOpieFvd{}$ (Table~\ref{tab:heldout}).
In a matched confirmation (two training seeds, 236 held-out contexts, shared native reference, inherited gate cutoff) the registered contrasts, backbone adapter minus recipe on the identity margin, are positive under full ($\bbXAuditEValue{}\,\bbXAuditEValueCI{}$) and gated ($\bbXAuditMValue{}\,\bbXAuditMValueCI{}$) deployment, and its distributional cost is larger: its FVD ratio to native is $\bbXBTwoFVD{}$ under our gate (interval $\bbXFAuditOValueCI{}$) against $\bbXFAuditRValue{}$ for the recipe.

\paragraph{CureWM-style data-repair baseline.}
Backbone (CureWM data) keeps the adapter and objective of Backbone (GT fit) and changes only its data: 20-step action windows perturbed by the six families of the original perturbation engine (severities $0.4$ and $0.8$); Recipe-CureWM applies the compact recipe to the same data.
Deviations from the original method: 20-step blocks with proxy labels replace full episodes, rank-16 LoRA with 1024 updates replaces full-model tuning, visual measures replace the original value-head measures, and closed-loop and real-robot experiments are omitted.
On our templates the scheduled recipe exceeds Backbone (CureWM data); the fresh-split row is in Table~\ref{tab:heldout}.

\paragraph{Closed loop with the backbone adapters.}
The test compares the recipe with Backbone (GT fit) and Backbone (our loss), all with the step schedule, over four training seeds (1701, 3407, 5501, 7919): LIBERO-10 tasks 0, 1 and 4, 50 initial states and two policy seeds per task, best-of-four selection with the frozen value head, $\eFourNEp{}$ episodes per arm and seed, run to success or 700 steps (520-step rows re-score the same trajectories); native, base, random and oracle come from the earlier three-task test, and all arms are paired by task, initial state and policy seed.
The registered readouts are each backbone adapter minus the recipe, pooled over the four seeds and read by the 95\% interval alone (upper end below zero reads ``recipe better''); intervals are task-stratified cluster bootstraps over (task, initial state), and comparisons with native and between adapters are descriptive.

\begin{table}[htbp]
  \centering\scriptsize\setlength\tabcolsep{3pt}
  \caption{Closed loop on three tasks, four training seeds pooled (rates are means over the seeds). Native is $\eFourNative{}$; the recipe against native is $\eFourRecNat{}\eFourRecNatCI{}$. Differences are success-rate differences with 95\% intervals.}
  \label{tab:s9-closed}
  \begin{tabular}{@{} l l c c c c c @{}}
    \toprule
    Adapter & Limit & Rate & Recipe rate & Minus recipe & $p$ & Minus native \\
    \midrule
    Backbone (GT fit) & 700 steps & \eFourBOneRate{} & \eFourRecipeRate{} & \ptci{\eFourBOneRec}{\eFourBOneRecCI} & \eFourBOneRecP{} & \ptci{\eFourBOneNat}{\eFourBOneNatCI} \\
    Backbone (our loss) & 700 steps & \eFourBTwoRate{} & \eFourRecipeRate{} & \ptci{\eFourBTwoRec}{\eFourBTwoRecCI} & \eFourBTwoRecP{} & \ptci{\eFourBTwoNat}{\eFourBTwoNatCI} \\
    \bottomrule
  \end{tabular}
\end{table}

Both registered readouts read ``recipe better'': Backbone (GT fit) minus recipe is $\eFourBOneRec{}\eFourBOneRecCI{}$ and Backbone (our loss) minus recipe is $\eFourBTwoRec{}\eFourBTwoRecCI{}$, with the same direction and size at 520 steps ($\eFourHBOneRec{}$ and $\eFourHBTwoRec{}$).
Every per-seed estimate of the gap is negative.
Neither adapter shows a gain over native, and the two adapters are not separated ($\eFourBOneBTwo{}\eFourBOneBTwoCI{}$, descriptive).
The adapters differ from the recipe in module, parameter count and, for Backbone (GT fit), loss.

\section{Qualitative examples}
\label{supp:n10}
\begin{figure}[!ht]
  \centering
  \includegraphics{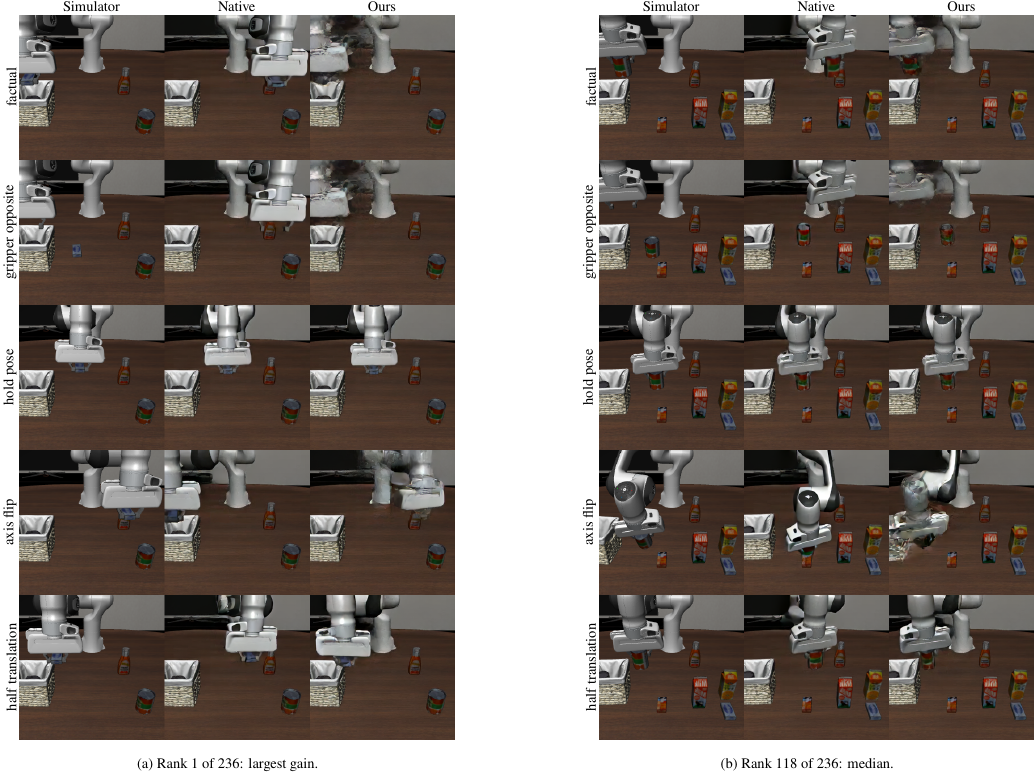}
  \caption{Step-scheduled repair on two held-out contexts. Rows: factual, gripper opposite, hold pose, translation axis flip, translation half. Columns: simulator, native, ours.}
  \label{fig:s10-qual}
\end{figure}

This section supports Fig.~\ref{fig:qual} and the qualitative statements of Sec.~\ref{sec:exp-main}. Fig.~\ref{fig:s10-qual} shows two held-out contexts at the final frame.

\paragraph{Selection.}
The export rule has no visual criterion: the 236 held-out contexts are ranked by the identity-margin gain of the always-on repair (four training seeds, two evaluation seeds, ties by context identifier). We show rank 1 (largest gain) and rank 118 (median, the context of Fig.~\ref{fig:teaser}a and Fig.~\ref{fig:qual}).

\paragraph{What the examples show.}
In the factual, gripper-opposite and translation rows, native renders a sharp arm at a position that differs from the simulator's, while ours moves the arm towards it; the hold-pose rows are close to native. Ours renders the endpoint frames slightly softer, the trade-off for the identity gain quantified in S5.

In Fig.~\ref{fig:qual}, the cosine of the difference from the factual branch with the simulator's difference is $\qlCosNatHi{}$ for native and $\qlCosOurLo{}$ for ours on the axis flip of the median context, and $0.00$ and $\qlCosOurHi{}$ on the half translation of the largest-gain context, computed on the displayed $256\times256$ frames.

\section{Design choices, human participants and licenses}
\label{supp:n11}

This section explains the design choices behind the evaluation (Sec.~\ref{sec:exp}), the human-participant details and the asset licenses.

\paragraph{Splits and selection.}
Splits are made by demonstration within the ten LIBERO-10 tasks, so that every gain is measured on new states of the tasks on which the closed-loop policy is evaluated; four further suites read new scenes (S8). Fresh and sealed were read by separately registered readouts, each with its endpoints fixed beforehand.
The recipe and per-model cut-offs were selected on demonstration 35 (S2). Confidence intervals cluster contexts or initial states, the units over which the repair acts.

\paragraph{Scheduling and closed loop.}
The FVD bound of $1.10$ is a tolerance relative to native, fixed before any repaired video was scored.
Scheduling adds no denoiser calls, parameters or training steps; we claim no wall-clock speedup.
Closed-loop results use a fixed policy, a learned value head and best-of-four selection, which isolate the effect of the WAM on the choice.
Training seeds within a closed-loop batch share one native reference (S7).
All experiments run in the simulator, which supplies the counterfactual ground truth on which the repair is trained and scored.

\paragraph{Transfer and baselines.}
Under the tested budgets the low-rank adapter gives an identity gain where full-rank updates of the two embedders do not (S9), and smaller data pools reduce it little.
The CureWM-style baseline follows the original data-repair idea under our budget (S9).

\paragraph{Human participants.}
Raters gave informed consent before starting, were paid at a rate that follows the norms for such studies, and responses were collected anonymously, in line with the ethical requirements for research with human participants that apply to us.

\paragraph{Assets and licenses.}
Licenses were checked in each public repository or model card.
Prophet checkpoints are derivatives of NVIDIA Cosmos-Predict2 and are governed by the NVIDIA Open Model License (``Built on NVIDIA Cosmos''); A2World code is Apache-2.0 and its checkpoints carry the same NVIDIA license.
The Cosmos Policy checkpoint is released under NVIDIA's one-way non-commercial license, LIBERO is MIT, and the DINOv2 code and weights are Apache-2.0.
For UWM and the Wan-based LIBERO-Goal WAM we found no license file and use them for research evaluation only.
We train adapters on these models for research only and do not redistribute third-party code, weights or data; the adapters are not released.

\paragraph{Code and protocol release.}
We will release the five-action diagnostic, the evaluation code and a reference implementation of \ours{} for Prophet, and then the full code after publication, so that any WAM can be checked for identity before it is used for planning or policy learning.

\section{Real-robot validation}
\label{supp:n12}

This section supports Sec.~\ref{sec:exp-real}.

\paragraph{Hardware and data.}
A PiPER six-degree-of-freedom arm is observed by two Intel RealSense D405 cameras, one fixed third-person view and one on the wrist.
Demonstrations are collected by teleoperation, with a second PiPER arm as leader, and recorded at 30\,Hz: both camera streams at $640\times480$ and seven-dimensional actions (six joint positions and the gripper).
Each demonstration performs one task under one fixed instruction.

\paragraph{World model and policies.}
The world model is Prophet fine-tuned on these demonstrations.
The original arm uses this checkpoint as it is; the repaired arm adds the \ours{} adapter under the step gate of Sec.~\ref{sec:method}.
In each world model a policy is post-trained with the same procedure as in simulation.
Policy initialization, reward, training budget and all other settings are identical for the two arms, so the two policies differ only in the world model they were trained in.

\paragraph{Tasks and success.}
Table~\ref{tab:s12-tasks} lists the scenes. A trial fails on time-out, human contact or an emergency stop.

\begin{table}[!htb]
  \centering\scriptsize\setlength\tabcolsep{4pt}
  \caption{Real-robot scenes and success criteria.}
  \label{tab:s12-tasks}
  \begin{tabular}{@{}l p{11.5cm}@{}}
    \toprule
    Scene & Success \\
    \midrule
    \rwSceneA{} & The lemon rests inside the bowl and the gripper is open. \\
    \rwSceneB{} & The cup rests inside the bowl and the banana lies on the table to the right of the bowl; both steps must be completed. \\
    \rwSceneC{} & The pot rests upright on the stove and the gripper is open. \\
    \rwSceneD{} & The spoon rests on the towel and the gripper is open. \\
    \bottomrule
  \end{tabular}
\end{table}

\paragraph{Blind evaluation.}
For each scene, \rwNEval{} object placements were drawn at random before testing.
On each placement the two policies run one after the other in random order, and the operator, who does not know which policy is running, records success or failure.
Each policy thus has \rwNEval{} trials per scene, \rwNAll{} in total.
The difference pooled over the scenes is tested with a Cochran--Mantel--Haenszel test stratified by scene ($p$~\rwStat{\rwPvCmh{}}); per-scene rates with 95\% Wilson intervals are in Fig.~\ref{fig:real}.

\end{document}